\documentclass[10pt]{article} %
\usepackage[preprint]{tmlr}

\usepackage{amsmath,amsfonts,bm}

\def\eqref#1{equation~\ref{#1}}

\def\1{\bm{1}}

\DeclareMathAlphabet{\mathsfit}{\encodingdefault}{\sfdefault}{m}{sl}
\SetMathAlphabet{\mathsfit}{bold}{\encodingdefault}{\sfdefault}{bx}{n}

\usepackage{hyperref}
\usepackage{url}
\usepackage{makecell}
\usepackage{amsmath}
\usepackage{amssymb}
\usepackage{amsthm}
\usepackage{booktabs}
\usepackage{graphicx}
\graphicspath{{./}{Images/}}
\usepackage{cleveref}
\usepackage{longtable}
\usepackage{booktabs}
\usepackage{array}
\usepackage{subcaption}
\usepackage{booktabs}
\usepackage{wrapfig}
\usepackage{enumitem}
\usepackage{float}

\usepackage{xcolor}
\usepackage{tikz}
\usetikzlibrary{arrows.meta,positioning,calc}

\definecolor{coordblue}{RGB}{222,235,247}
\definecolor{splinegreen}{RGB}{226,239,218}
\definecolor{fnorange}{RGB}{252,228,214}
\definecolor{condpurple}{RGB}{234,224,244}
\definecolor{outputgray}{RGB}{240,240,240}

\usepackage{xcolor}
\definecolor{codegreen}{rgb}{0,0.6,0}
\definecolor{codegray}{rgb}{0.5,0.5,0.5}
\definecolor{codepurple}{rgb}{0.58,0,0.82}
\definecolor{navyblue}{rgb}{0, 0, 0.52}
\definecolor{rubyred}{rgb}{0.6,0.07,0.12}
\definecolor{backcolour}{rgb}{0.95,0.95,0.92}

\usepackage{hyperref}
\hypersetup{
colorlinks=true,
linkcolor=rubyred,      %
citecolor=navyblue,       %
filecolor=codegray,        %
urlcolor=codepurple,       %
pdfborder={0 0 0},         %
}

\usepackage{listings}
\lstdefinestyle{lsplinepython}{
language=Python,
basicstyle=\ttfamily\footnotesize,
keywordstyle=\color{navyblue},
commentstyle=\color{codegray},
stringstyle=\color{codepurple},
columns=fullflexible,
keepspaces=true,
showstringspaces=false,
breaklines=true,
frame=single,
rulecolor=\color{codegray},
captionpos=b,
aboveskip=\medskipamount,
belowskip=\medskipamount
}

\theoremstyle{plain}

\theoremstyle{remark}

\title{Learnable Splines for Unstructured Neural Operators}

\author{\name Vignesh Gopakumar\thanks{Correspondence to: \texttt{v.gopakumar@ucl.ac.uk}}  \; \& Marc Peter Deisenroth  \\
      \addr Centre for Artificial Intelligence, University College London \\
      London, WC1V 6LJ, UK
      \AND
      \name Ander Gray \\
      \addr LIX, \'Ecole Polytechnique, CNRS, Institut Polytechnique de Paris \\
      Paris, 91120, France
      \AND
      \name Josh Williams \& Nuno Nobre \\
      \addr Hartree Centre, Science and Technology Facilities Council \\
      Daresbury, WA4 4AD, UK
      \AND
      \name William M.E. Ellis, Matthew Field, James Panayis, Alexander I. Blair, \\
      \name Andrew Davis, Stanislas Pamela \& Robert Akers \\
      \addr Computing Division, UK Atomic Energy Authority \\
      Oxford, OX14 3EB, UK
      \AND
      \name Zhuoyuan Wang, Valentin Duruisseaux \& Anima Anandkumar \\
      \addr Computing + Mathematical Sciences, Caltech \\
      Pasadena, 91125, USA}

\def\month{MM}  %
\def\year{YYYY} %
\def\openreview{\url{https://openreview.net/forum?id=XXXX}} %

\begin{document}

\maketitle

\begin{abstract}
Neural operators approximate solution operators of partial differential equations at a fraction of the cost of numerical solvers. Efficient grid-based architectures such as the Fourier neural operator (FNO) assume uniform rectangular grids, whereas engineering simulations use large, irregular, and locally refined meshes. We introduce L-SplineFNO, a memory-efficient neural operator that couples an FNO backbone with a sparse, learnable tensor-product B-spline transfer between unstructured meshes and a latent Cartesian grid. A local normalised scatter avoids the geometry-dependent solve of analytic spline transformation while preserving compact support, partition of unity, and spline smoothness by construction. The neural operator, conditioned on global design and operating parameters, learns the mapping of geometry to the field solution. Across six three-dimensional engineering benchmarks, L-SplineFNO achieves strong predictive accuracy while requiring substantially less memory and compute than baseline operators, scaling to 26.6 million nodes in training and 46 million in inference on a single A100 GPU. It also supports zero-shot super-resolution and efficient closed-form spatial derivatives. Uniquely, the analytic spline decoder yields rigorous prediction bounds over continuous space and uncertain parameters, via Bernstein convex hulls and Taylor models, demonstrated on a 29.2-million-node MAST-U tokamak model.

\end{abstract}

\section{Introduction}
\label{sec:introduction}

Engineering design and qualification rest on repeated numerical simulation, evaluating a system's governing equations across a range of parameters and conditions \citep{winsberg2026simulations, simintelligence}. In steady-state problems such as thermal management, structural loading and aerodynamic pressures, the quantity of interest is a field defined over an unstructured finite-element or finite-volume discretisation of a three-dimensional domain, determined by a small set of design and operating parameters \citep{MOOSE_Permann_2020, OpenFOAM}. Design sweeps, sensitivity studies and uncertainty quantification each require this map to be evaluated thousands of times, while a single high-fidelity solve can occupy a compute cluster for hours \citep{Keyes_Sim_Challenges}. The cost of the solver therefore limits how much of a design space can be explored \citep{gopakumar2025calibrated}. Safety-critical settings such as fusion devices also require guaranteed prediction ranges for mesh nodes  under uncertain operating conditions.

Neural operators provide a route past this bound, learning mappings between function spaces with parameters shared across discretisations and approximating solution operators of partial differential equations (PDEs) at a fraction of the cost of the solver that generated their training data \citep{Kovachki_NeuralOperators_2023,azizzadenesheli2024neuraloperatorsacceleratingscientific,Berner2026}. The Fourier neural operator (FNO) \citep{li2021fourierneuraloperatorparametric} has been deployed as a surrogate across fluid dynamics \citep{Wen2022UFNO}, heat transfer \citep{Yuan2025CHT} and plasma modelling \citep{Carey_2025,neural_parareal_pamela2025}, reporting speed-ups of several orders of magnitude while retaining discretisation convergence and zero-shot super-resolution \citep{Gopakumar_2024}. This efficiency follows from the fast Fourier transform, which assumes the field is sampled on a uniform rectangular grid. However, engineering meshes are often irregular, feature-refined and vary across designs, rarely satisfying this assumption. Existing approaches either process the unstructured mesh directly, which can become memory-intensive at large node counts, or preserve structured processing by transferring the field onto a latent grid. Retaining information from the original mesh while keeping the representation memory-efficient is therefore a key challenge in neural PDEs for large unstructured grids. Existing learned transfers use global coordinate deformations \citep{GeoFNO_Li2024} or kernel integrals over a radius-limited neighbourhood graph \citep{li2023geometryinformedneuraloperatorlargescale}, but neither preserves structure by construction (\Cref{sec:background}).

\begin{figure}[t]
    \centering
    \includegraphics[width=0.9\columnwidth]{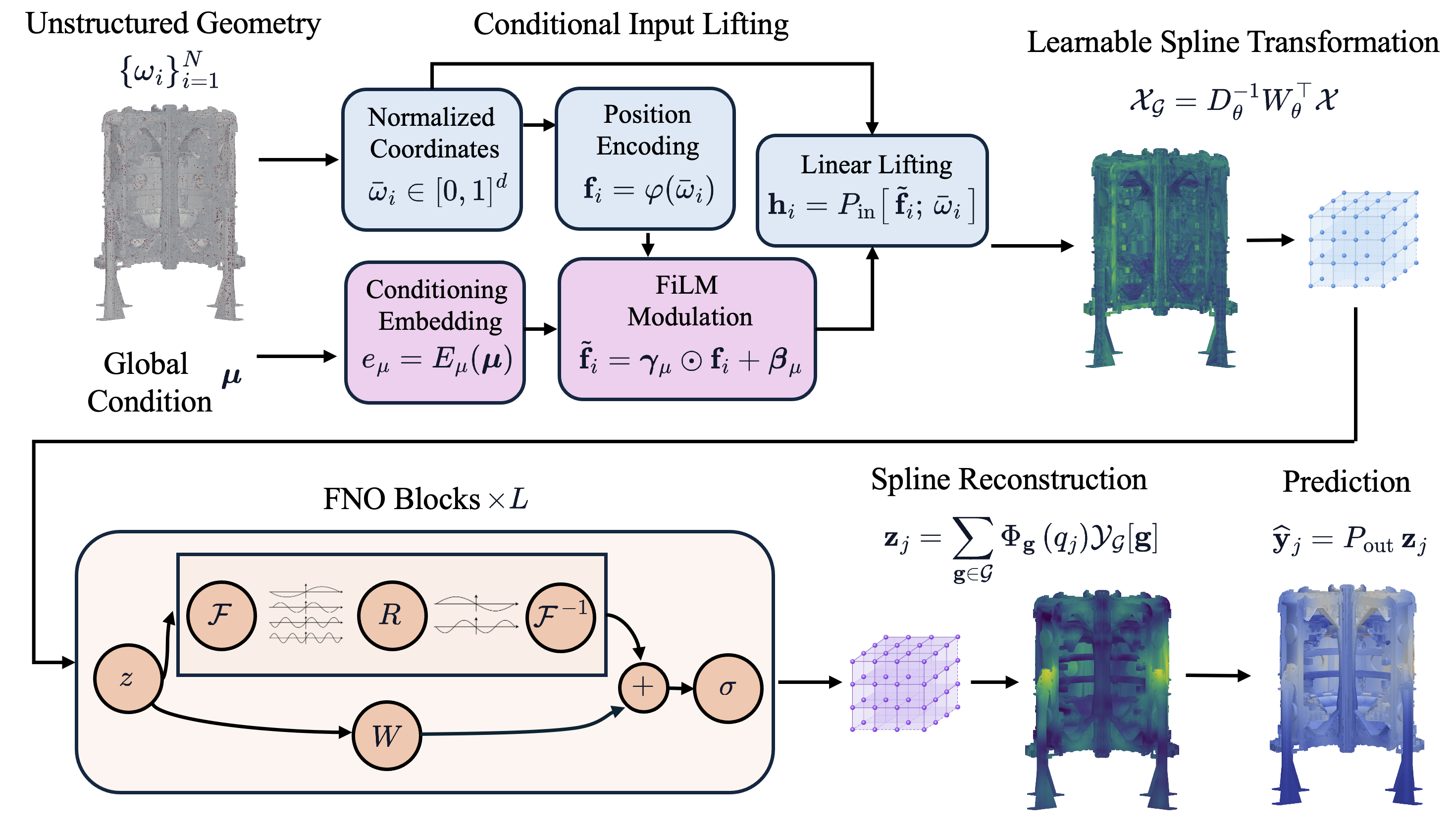}
\caption{\textbf{End-to-end conditional L-SplineFNO architecture.} The unstructured geometry and global conditioning vector are first mapped to conditioned nodal features through coordinate normalisation, positional encoding, and FiLM modulation. A learnable $C^{p-1}$ spline scatter then transforms these features from the unstructured mesh to a structured latent grid, where they are processed by a Fourier neural operator (FNO) backbone. The resulting grid representation is mapped back to the physical mesh using analytic spline reconstruction and a pointwise projection to produce the predicted field. This construction provides a sparse, structure-preserving, differentiable mesh-to-grid-to-mesh mapping with the structured Fourier processing.}
\vspace{-1em}
\label{fig:cond-lsfno-pipeline}
\end{figure}

{\bf Our approach:} We build the transfer as a sparse, differentiable and structurally constrained linear map from tensor-product B-splines \citep{piegl1996nurbs,prautzsch2002bezier}. Compact support gives each mesh point a local stencil whose size is set by the spline degree, not by a search radius. Partition of unity preserves constants. Spline smoothness gives closed-form evaluation and differentiation at any off-grid query point. We call a transfer with all three properties \emph{structure-preserving}. The same structure gives a non-negative Bernstein basis on each decoder cell, so the prediction lies within the convex hull of its coefficients; propagating a Taylor model of the parameters through the network extends this to rigorous bounds over continuous space and uncertain parameters. The resulting sparse transfer preserves physical locality and gives a memory-efficient link between unstructured meshes and structured Fourier processing.

To further address the parametric nature of engineering problems, we incorporate global design and operating parameters into the operator. With no initial condition to propagate in steady-state problems, the geometry $\Omega$ and a conditioning vector $\boldsymbol{\mu}$ of design variables, loads, material properties, and operating conditions carry all sample-specific information. How $\boldsymbol{\mu}$ should enter an operator is unsettled, and its effect is confounded by a redundancy that is rarely controlled for \citep{koupaï2024gepsboostinggeneralizationparametric}. When $\boldsymbol{\mu}$ restates geometric quantities already visible in the point cloud, a model can ignore the conditioning pathway and still appear conditioned. Isolating this pathway therefore requires settings in which $\boldsymbol{\mu}$ carries information that the mesh cannot provide, up to the limiting case of a fixed mesh on which $\boldsymbol{\mu}$ is the only sample-specific signal. Our benchmarks span this range, from the heat sink, where $\boldsymbol{\mu}$ restates the geometry, to HIVE, whose fixed mesh leaves $\boldsymbol{\mu}$ as the only sample-specific signal (\Cref{sec:experiments}). We combine this conditioning with the learnable spline transfer to form a complete conditional L-SplineFNO for parametric fields on complex unstructured meshes. Our contributions are:

\begin{itemize}[leftmargin=1.5em]

\item \textbf{Learnable spline transfer for unstructured neural operators.}
We introduce L-SplineFNO, which couples an FNO backbone with a learnable, structure-preseerving spline transformation from unstructured meshes to a structured latent grid. This replaces the geometry-dependent linear solve of analytic SplineFNO with a \textit{learnable}, structure-preserving B-spline scatter onto a latent FNO grid, with analytic reconstruction at arbitrary query points.

\item \textbf{An end-to-end conditional operator framework.}
We condition the operator on global design and operating parameters through an affine transformation \citep{perez2018film}. The FiLM-style conditioning produces a per-channel scale and shift, applied to the encoded geometry features before the processor backbone. The conditioning method sits in its own module with interchangeable backbones.

\item \textbf{Rigorous spatial and parameter bounds.}
Bernstein convex hulls of the spline decoder and second-order Taylor models of the trained network enclose the prediction over continuous space and uncertain parameter boxes, without reference solutions or retraining.

\item \textbf{Comprehensive evaluation on complex engineering problems.}
Across six three-dimensional thermal, structural, and aerodynamic benchmarks, we compare L-SplineFNO against representative mesh-native, latent-attention, and gridded neural operators under a common conditioning pipeline and at matched parameter counts, demonstrating strong predictive accuracy while retaining the efficiency of structured latent processing. We further demonstrate linear scaling with mesh size, inexpensive analytic spatial derivatives, and deployment on a 29.2-million-node mesh of the MAST-U tokamak.
\end{itemize}

\section{Background and Related Work}
\label{sec:background}

\paragraph{Conditional operator learning.}
Let $\mathcal{U}$ denote a space of solution fields on a domain $\Omega$, and let $\boldsymbol{\mu}\in\mathcal{P}\subset\mathbb{R}^{k}$ collect design variables, material properties, boundary conditions, or operating conditions. For the steady problems considered here, with no input field, the \emph{solution operator} maps the geometry and parameters to the solution,
\begin{equation}
\mathcal{N}:(\boldsymbol{\mu},\Omega)\longmapsto
y\in\mathcal{U},
\qquad
\mathcal{L}_{\boldsymbol{\mu},\Omega}(y)=f,
\label{eq:conditional-solution-operator}
\end{equation}
where $\mathcal{L}_{\boldsymbol{\mu},\Omega}$ is the governing operator and $f$ the forcing, if any. All sample-specific information enters through $\Omega$ and $\boldsymbol{\mu}$, so the learning task is both geometric and conditional.

\paragraph{Neural operators.}
Neural operators approximate maps such as \Cref{eq:conditional-solution-operator} using architectures whose parameters are shared across discretisations, so that a trained model converges across mesh discretisations rather than to a fixed resolution \citep{Kovachki_NeuralOperators_2023,Berner2026}. The FNO parametrises the kernel integral as a global convolution in Fourier space \citep{li2021fourierneuraloperatorparametric,duruisseaux2026fourierneuraloperatorsexplained}: for a lifted feature field $v_{t}$, a layer computes
\begin{equation}
v_{t+1}(x)
=
\sigma\!\left(
    W v_{t}(x)
    +
    \mathcal{F}^{-1}\!\left(
        R_{\phi}\cdot\mathcal{F}(v_{t})
    \right)(x)
\right),
\label{eq:fno-layer}
\end{equation}
where $\mathcal{F}$ is the Fourier transform, $R_{\phi}$ is a learned multiplier on a truncated band of modes, and $W$ is a pointwise linear map. Physics-informed neural operators further constrain the learned solution family by penalising PDE residuals \citep{li2023physicsinformedneuraloperatorlearning,ganeshram2026fcpinohighprecisionphysicsinformed}. Evaluating $\mathcal{F}$ with the fast Fourier transform offers only $\mathcal{O}(N\log N)$ complexity, making the \Cref{eq:fno-layer} efficient and FNOs a popular choice \citep{dehoop2022costaccuracytradeoffoperatorlearning}, but restricts its operation to uniform rectangular grids.

\paragraph{Operator learning on unstructured geometries.}
Extensions to unstructured grids fall into three groups. \emph{Mesh-native} models attend over mesh points or learned groupings of them, as in GNOT \citep{hao2023gnotgeneralneuraloperator}, Transolver \citep{wu2024transolverfasttransformersolver} and its parallel successor Transolver++ \citep{luo2025transolveraccurateneuralsolver}, or compress the point set into latent tokens, as in UPT \citep{alkin2025universalphysicstransformersframework}; they avoid gridding, but cost and fidelity depend on how aggressively the mesh is compressed. \emph{Latent-grid} models keep a structured backbone by transferring the data onto a regular domain, through learned deformations in Geo-FNO \citep{GeoFNO_Li2024}, balanced non-uniform partitions in NUNO \citep{liu2023nunogeneralframeworklearningnuno}, graph-kernel encoders in GINO \citep{li2023geometryinformedneuraloperatorlargescale}, multi-scale attentional encoders with a vision-transformer processor in GAOT \citep{wen2026geometryawareoperatortransformer}, or optimal-transport maps for surfaces \citep{li2025geometricoperatorlearningoptimal}; here the transfer becomes part of the approximation, since coarse grids lose thin features and interfaces, while dense grids and large neighbourhood graphs cost memory. \emph{Precomputed} representations such as PGD-NO build geometry tokens ahead of time, saving memory at very large mesh sizes but removes the representation from end-to-end training \citep{zhong2026pgdnoneuraloperatorprecomputed}. Industry-scale comparisons find no representation uniformly best across accuracy, cost, memory and deployment \citep{zhong2025comprehensive}.

\paragraph{B-splines in scientific machine learning.}
Tensor-product B-splines are compactly supported, smooth, and form a partition of unity \citep{piegl1996nurbs}. Their sparse Jacobians have been used as differentiable layers for geometry reconstruction and finite-element learning \citep{cho2021differentiablesplineapproximations}; physics-informed B-spline networks predict control coefficients directly, so that derivatives and boundary conditions can be imposed analytically \citep{wang2026physicsinformeddeepbsplinenetworks}; and Neural Spline Operators extend this coefficient-learning view from finite-dimensional PDE parameters to functional inputs \citep{wang2025neuralsplineoperatorsrisk}. In each, B-splines serve as the \emph{output} representation, rather than as a parameter-efficient, learnable transfer across an unstructured mesh and a latent structured grid.

\paragraph{Conditioning on physical parameters.}
The parameters $\boldsymbol{\mu}$ either restate information already visible in the geometry or encode quantities the geometry cannot reveal, such as loads, materials and operating conditions. Broadcasting $\boldsymbol{\mu}$ as an extra input channel is simple, but asks the backbone to wastefully preserve and propagate a signal that is spatially constant across the domain \citep{takamoto2023learningneuralpdesolvers}. Alternatives inject $\boldsymbol{\mu}$ into the network directly: feature-wise linear modulation (FiLM) applies parameter-dependent, channel-wise affine transformations \citep{perez2018film}; adaptive normalisation and cross-attention modulate intermediate features in a similar spirit \citep{peebles2023scalablediffusionmodelstransformers,rebain2022attentionbeatsconcatenationconditioning}; and hypernetworks generate the operator's weights from $\boldsymbol{\mu}$ \citep{jing2026extendingfourierneuraloperators}.

\section{Spline-Gridded Fourier Neural Operators}
\label{sec:spline-fno}
\Cref{fig:cond-lsfno-pipeline} summarises the method: sparse spline operators carry mesh features to a uniform latent grid for a standard FNO, and reconstruct the output at arbitrary query points. We first introduce conditional input lifting, then formulate analytic SplineFNO, and finally develop L-SplineFNO by replacing its geometry-dependent spline transform with a local learnable scatter.

\paragraph{Notation.}
Let $\{\omega_i\}_{i=1}^{N}\subset\Omega$ denote the nodes of an unstructured input mesh in a $d-$dimensional domain $\Omega\subset\mathbb{R}^{d}$. We introduce a uniform tensor-product control grid with multi-index set $\mathcal{G} =\prod_{\ell=1}^{d}\{0,\ldots,S_\ell-1\}$; a multi-index is written $\mathbf{g}=(g_1,\ldots,g_d)\in\mathcal{G}$, where $g_\ell$ is the grid index along coordinate axis $\ell$. The grid contains $G := |\mathcal{G}| = \prod_{\ell=1}^{d}S_\ell$ control points, with coordinates $\xi_{\mathbf{g}}^{(\ell)} = \xi_0^{(\ell)} + g_\ell\Delta_\ell$, $\ell=1,\ldots,d$, where $\xi_0^{(\ell)}$ and $\Delta_\ell>0$ are the origin and spacing along axis $\ell$. The spacing may differ between coordinate axes, but is constant along each axis.

\subsection{Conditional Input Lifting}
\label{sec:lsfno-pipeline}

Many computational physics problems depend on both geometry and global parameters such as material properties, boundary conditions, source terms, or operating conditions. We introduce conditioning in our framework at the input lifting stage to incorporate these problem-specific parameters into the nodal features supplied to the operator. The model receives an unstructured set of geometry coordinates $\{\omega_i\}_{i=1}^{N}$, a global conditioning vector $\boldsymbol{\mu}\in\mathbb{R}^{k}$, and a set of output query points $\{q_j\}_{j=1}^{N_q}$. No input field is supplied, and the sample-specific signal is carried by the geometry and $\boldsymbol{\mu}$.

In the feature path, each normalised coordinate $\bar{\omega}_i\in[0,1]^d$ is encoded by integrated sinusoidal features $\varphi(\bar{\omega}_i)\in\mathbb{R}^{2dn_f}$ at $n_f$ frequencies. The global condition is embedded as $e_\mu=E_\mu(\boldsymbol{\mu})$ and applied through FiLM, the most efficient of the conditioning methods we compared (\Cref{app:conditioning_methods}),
\begin{equation}
\tilde{\mathbf{f}}_i
=
\boldsymbol{\gamma}_\mu\odot\varphi(\bar{\omega}_i)
+
\boldsymbol{\beta}_\mu,
\qquad
\mathbf{h}_i
=
P_{\mathrm{in}}
\bigl[\,
    \tilde{\mathbf{f}}_i;\,
    \bar{\omega}_i
\,\bigr],
\label{eq:pipeline-lift}
\end{equation}
where $\boldsymbol{\gamma}_\mu=H_\gamma(e_\mu)$ and $\boldsymbol{\beta}_\mu=H_\beta(e_\mu)$ are broadcast across all mesh nodes. The modulation acts only on the encoded features $\varphi$; the normalised coordinates bypass it and are concatenated back before the pointwise lift. We collect the lifted nodal features as $\mathcal{X}=[\mathbf{h}_1,\ldots,\mathbf{h}_N]^\top\in\mathbb{R}^{N\times c}$. In the geometry path, the same normalised coordinates $\bar{\omega}_i$, unmodulated, determine the compact mesh-to-grid spline stencils (\Cref{fig:cond-lsfno-pipeline}).

\begin{figure}[H]
\centering
\includegraphics[width=0.78\linewidth]{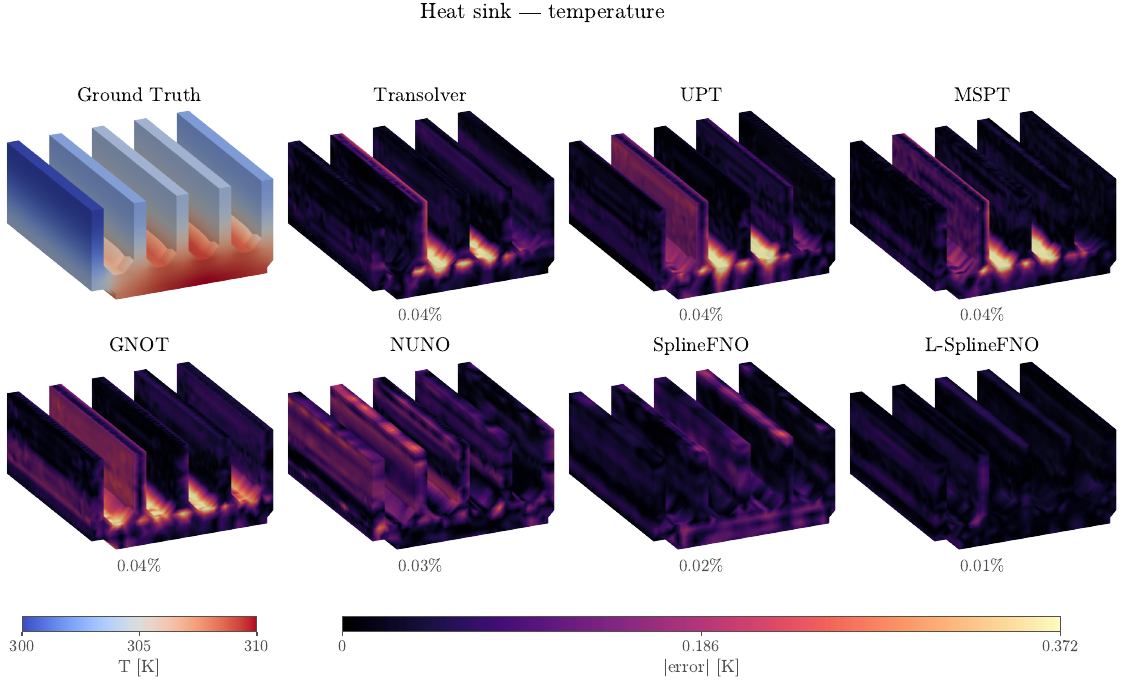}
\caption{Error maps produced by baseline architectures in modelling heat distribution across the heat sink. L-SplineFNO produces the most accurate predictions, with minimal errors concentrated at the edges of the heat sink where the temperature gradient is highest.}
\label{fig:heat_sink_error_maps_main}
\end{figure}

\subsection{SplineFNO}
\label{sec:splinefno}

After constructing the conditioned nodal features, we first introduce SplineFNO, which uses fixed analytic spline operators to transform from the unstructured mesh to a latent uniform grid and to reconstruct at the output coordinates, with a standard FNO operating on the latent grid. Writing $t_i^{(\ell)} = (\omega_i^{(\ell)}-\xi_0^{(\ell)})/\Delta_\ell$ for the continuous grid coordinate of mesh node $\omega_i$ along axis $\ell$, and letting $B_p$ denote the centred cardinal B-spline of degree $p$ introduced in \Cref{sec:tensor-product}, we use $p=3$ by default and write $B:=B_p$ for conciseness. The unnormalised tensor product basis matrix $\widetilde{\Phi}\in\mathbb{R}^{N\times G}$ is defined as
\begin{equation}
\widetilde{\Phi}_{i,\mathbf{g}}
=
\prod_{\ell=1}^{d}
B\!\left(t_i^{(\ell)}-g_\ell\right),
\qquad
i=1,\ldots,N,\quad \mathbf{g}\in\mathcal{G}.
\label{eq:spline-basis}
\end{equation}
Because the degree $p$ spline has compact support over $p+1$ grid points, each row of $\widetilde{\Phi}$ contains at most $(p+1)^d$ nonzero entries. To preserve constants near the boundary, contributions outside $\mathcal{G}$ are discarded and the remaining basis values are normalised row wise as $\Phi_{i,\mathbf{g}}=\widetilde{\Phi}_{i,\mathbf{g}}/\sum_{\mathbf{k}\in\mathcal{G}}\widetilde{\Phi}_{i,\mathbf{k}}$, so that partition of unity is maintained: $\sum_{\mathbf{g}\in\mathcal{G}}\Phi_{i,\mathbf{g}}=1$, whenever the spline support intersects the control grid.

The forward pass consists of three stages. \textbf{(i) Spline transformation: unstructured to structured.} We seek grid coefficients $\mathcal{X}_{\mathcal{G}}\in\mathbb{R}^{G\times c}$ whose spline reconstruction approximates the nodal features,
\begin{equation}
\mathcal{X}_{\mathcal{G}}
=
\left(\Phi^\top\Phi+\lambda I_G\right)^{-1}
\Phi^\top\mathcal{X},
\label{eq:spline-encode}
\end{equation}
where $\lambda>0$ ensures that $\Phi^\top\Phi+\lambda I_G$ is positive definite. \textbf{(ii) Grid-space neural operator.} The coefficients are reshaped according to the tensor-product grid and processed by a standard FNO, $\mathcal{Y}_{\mathcal{G}}=\mathcal{F}_{\theta}(\mathcal{X}_{\mathcal{G}})$. \textbf{(iii) Spline reconstruction: structured to unstructured.} The grid features are evaluated using the same analytic basis,
\begin{equation}
\mathcal{Y}
=
\Phi\,
\mathcal{F}_{\theta}
\!\left(
    \left(\Phi^\top\Phi+\lambda I_G\right)^{-1}
    \Phi^\top\mathcal{X}
\right).
\label{eq:spline-decode}
\end{equation}
The spline matrices are sparse, analytic functions of the sampling coordinates and contain no learned parameters. Learning occurs in $\mathcal{F}_{\theta}$, while gradients propagate through both spline transfers.

\subsection{Learnable SplineFNO (L-SplineFNO)}
\label{sec:l-splinefno}

The spline transformation in \Cref{eq:spline-encode} requires a geometry-dependent linear solve, whose inverse is generally dense. L-SplineFNO replaces this solve with a local, learnable scatter that directly transfers nodal features to the latent grid.
The learned quantities are the transfer kernel shapes, while the grid nodes and analytic spline basis remain fixed. Along each coordinate axis $\ell$, we use
\begin{equation}
\kappa_{\theta}^{(\ell)}(t)
=
B(t)
+
\sum_{j=1}^{J}
\zeta_j^{(\ell)}
B\!\left(\frac{t-\tau_j}{h}\right),
\qquad
\zeta^{(\ell)}=V_0\alpha_{\theta}^{(\ell)},
\label{eq:learnable-spline-kernel}
\end{equation}
that is, the analytic spline $B$ plus $J$ narrower B-spline corrections with fixed centres $\tau_j$ and relative width $h\in(0,1)$. The columns of $V_0$ span the null space of the partition-of-unity constraints, so only the unconstrained coordinates $\alpha_{\theta}^{(\ell)}$ are trained. The constrained correction preserves compact support, partition of unity to numerical precision, and $C^{p-1}$ regularity. Details are given in \Cref{app:learnable-spline-construction}.

The transfer weights are tensor products of the per-axis kernels, normalised over the valid stencil of node $i$: $\mathcal{S}_\ell(i)$ holds the grid indices on axis $\ell$ within the kernel support of $t_i^{(\ell)}$, and $\mathcal{S}(i)=\prod_{\ell}\mathcal{S}_\ell(i)$ the at most $(p+1)^d$ grid points of its tensor-product stencil (\Cref{eq:valid-degree-p-stencil}),
\begin{equation}
(W_\theta)_{i,\mathbf{g}}
=
\prod_{\ell=1}^{d}
\frac{
\kappa_\theta^{(\ell)}\!\left(t_i^{(\ell)}-g_\ell\right)
}{
\displaystyle
\sum_{g'\in\mathcal{S}_\ell(i)}
\kappa_\theta^{(\ell)}\!\left(t_i^{(\ell)}-g'\right)
},
\qquad
\mathbf{g}\in\mathcal{S}(i),
\label{eq:learnable-transfer-weights}
\end{equation}
and zero otherwise. Each grid node then forms a normalised weighted accumulation, with
$n_{\mathbf{g}}=\sum_{i=1}^{N}(W_\theta)_{i,\mathbf{g}}\mathcal{X}_i$,
$d_{\mathbf{g}}=\sum_{i=1}^{N}(W_\theta)_{i,\mathbf{g}}$, and
$\widetilde{\mathcal{X}}_{\mathcal{G}}[\mathbf{g}]
=n_{\mathbf{g}}/\widetilde d_{\mathbf{g}}$,
where $\widetilde d_{\mathbf{g}}$ is the sign preserving safeguarded version of
$d_{\mathbf{g}}$ defined in \Cref{app:stable-spline-scatter}. Defining
$D_\theta=\operatorname{diag}(\widetilde d_{\mathbf{g}})$,
the learnable spline transformation can then be written compactly as
\begin{equation}
\widetilde{\mathcal X}_{\mathcal G}
=
D_\theta^{-1}W_\theta^\top\mathcal X,
\label{eq:learnable-normalised-scatter}
\end{equation}
which provides a local approximation to the spline transformation
$(\Phi^\top\Phi+\lambda I_G)^{-1}\Phi^\top\mathcal X$
without requiring a geometry-dependent linear solve.

To retain mesh coverage information removed by normalisation, we augment the scattered features with an analytic density feature $\rho$ computed from the fixed spline basis and form the FNO input
\begin{equation}
\mathcal{X}_{\mathcal{G}}
=
P_{\rho}
\left[
    \widetilde{\mathcal{X}}_{\mathcal{G}};
    \log(1+\rho)
\right],
\qquad
\mathcal{Y}_{\mathcal{G}}
=
\mathcal{F}_{\theta}(\mathcal{X}_{\mathcal{G}}),
\label{eq:pipeline-encode}
\end{equation}
where $P_\rho$ is a learned pointwise projection. The construction of $\rho$ is detailed in \Cref{app:stable-spline-scatter}.
Finally, the output is reconstructed analytically at each query point and projected to the output channels
\begin{equation}
\mathbf{z}_j
=
\sum_{\mathbf{g}\in\mathcal{G}}
\Phi_{\mathbf{g}}(q_j)
\mathcal{Y}_{\mathcal{G}}[\mathbf{g}],
\qquad
\widehat{\mathbf{y}}_j
=
P_{\mathrm{out}}\mathbf{z}_j,
\label{eq:pipeline-output}
\end{equation}
where $P_{\mathrm{out}}$ is a learned pointwise output projection.

\paragraph{Region-specific transfers.} On meshes with several materials or regions, each node carries a region label and can be routed through its own lift, scatter kernels and output projection, around one shared latent grid and FNO. The per-region scatter sums are added before the division in \Cref{eq:learnable-normalised-scatter}, so constants are still reproduced. A per-region readout lets each region take its own combination of the shared latent features, so the field can remain continuous across an interface while its normal derivative jumps. Only the lift, projection and kernel weights are duplicated; the FNO is shared. We use this whenever the mesh is naturally partitioned by regions or materials.

\section{Experiments}
\label{sec:experiments}

\newcommand{\tm}[1]{{\color{black!65}\tiny #1}}   %
\begin{table}[t]
\centering
\captionsetup{font=small}
\scriptsize
\setlength{\tabcolsep}{4pt}
\renewcommand{\arraystretch}{1.1}

\caption{Relative $L_2$ error (\%, mean $\pm$ std over five seeds; top line of each entry) and training time (hh:mm on a single NVIDIA A100; grey line beneath) for each architecture under FiLM conditioning. The best mean error for each case is shown in bold. $N$ is the number of spatial training nodes. Heat Sink, Heated Duct and HIVE predict absolute temperature, so their relative errors are small by construction. L-SplineFNO outperforms baselines on most experiments, while remaining significantly compute-efficient.}
\label{tab:film_all_cases_metrics}

\resizebox{\linewidth}{!}{%
\begin{tabular}{l*{6}{c}}
\toprule
\textsc{Model}
& \textsc{Heat Sink} & \textsc{Bracket} & \textsc{Aircraft}
& \textsc{Building Fire} & \textsc{Heated Duct} & \textsc{HIVE Monoblock} \\
& \tm{$N=16{,}384$} & \tm{$N=8{,}192$} & \tm{$N=131{,}072$}
& \tm{$N=98{,}496$} & \tm{$N=124{,}800$} & \tm{$N=194{,}691$} \\
\midrule
\multicolumn{7}{l}{\textit{Baselines}} \\
\addlinespace[2pt]
\textsc{Transolver}
& $0.040{\pm}0.002$ & $0.591{\pm}0.025$ & $8.32{\pm}0.88$ & $7.341{\pm}0.220$ & $0.473{\pm}0.050$ & $0.301{\pm}0.061$ \\
& \tm{00:41} & \tm{01:48} & \tm{01:26} & \tm{01:17} & \tm{01:15} & \tm{02:28} \\
\addlinespace[3pt]
\textsc{UPT}
& $0.039{\pm}0.001$ & $0.633{\pm}0.014$ & $13.87{\pm}0.43$ & $7.162{\pm}0.158$ & $0.523{\pm}0.053$ & $0.299{\pm}0.109$ \\
& \tm{00:33} & \tm{01:30} & \tm{00:58} & \tm{01:00} & \tm{0:50} & \tm{01:25} \\
\addlinespace[3pt]
\textsc{MSPT}
& $0.038{\pm}0.002$ & $0.544{\pm}0.049$ & $7.20{\pm}0.51$ & $7.217{\pm}0.209$ & $0.500{\pm}0.057$ & $0.315{\pm}0.090$ \\
& \tm{01:26} & \tm{03:29} & \tm{06:14} & \tm{06:18} & \tm{05:25} & \tm{07:40} \\
\addlinespace[3pt]
\textsc{GNOT}
& $0.042{\pm}0.001$ & $\boldsymbol{0.534{\pm}0.040}$ & $\boldsymbol{6.86{\pm}0.48}$ & $7.456{\pm}0.394$ & $0.652{\pm}0.032$ & $0.243{\pm}0.030$ \\
& \tm{01:33} & \tm{04:10} & \tm{03:25} & \tm{03:16} & \tm{02:52} & \tm{08:50} \\
\addlinespace[3pt]
\textsc{NUNO}
& $0.068{\pm}0.001$ & $1.972{\pm}0.095$ & $11.72{\pm}1.41$ & $6.805{\pm}0.087$ & $0.989{\pm}0.144$ & $0.471{\pm}0.120$ \\
& \tm{00:14} & \tm{00:40} & \tm{00:14} & \tm{00:34} & \tm{00:10} & \tm{00:15} \\
\addlinespace[3pt]
\textsc{GINO}
& $0.052{\pm}0.002$ & $1.385{\pm}0.010$ & $11.35{\pm}0.88$ & $7.201{\pm}0.171$ & $1.095{\pm}0.067$ & $0.370{\pm}0.084$ \\
& \tm{01:06} & \tm{09:18} & \tm{08:44} & \tm{03:04} & \tm{01:17} & \tm{06:36} \\
\midrule
\multicolumn{7}{l}{\textit{Spline-based (Ours)}} \\
\addlinespace[2pt]
\textsc{SplineFNO}
& $0.035{\pm}0.001$ & $5.490{\pm}0.025$ & $18.27{\pm}1.11$ & $7.611{\pm}0.092$ & $0.542{\pm}0.063$ & $0.216{\pm}0.013$ \\
& \tm{01:22} & \tm{03:46} & \tm{00:48} & \tm{01:57} & \tm{00:17} & \tm{00:18} \\
\addlinespace[3pt]
\textsc{L-SplineFNO}
& $\boldsymbol{0.027{\pm}0.001}$ & $1.978{\pm}0.006$ & $8.55{\pm}0.86$ & $\boldsymbol{6.443{\pm}0.131}$ & $\boldsymbol{0.454{\pm}0.044}$ & $\boldsymbol{0.211{\pm}0.045}$ \\
& \tm{00:42} & \tm{00:54} & \tm{01:04} & \tm{00:38} & \tm{00:25} & \tm{01:06} \\
\bottomrule
\end{tabular}%
}
\end{table}

\paragraph{Data:}
We evaluate on six steady-state, three-dimensional engineering benchmarks, including a CPU-board \textbf{heat sink} \citep{zhong2026pgdnoneuraloperatorprecomputed}, a structurally loaded \textbf{bracket} \citep{zhong2025comprehensive}, \textbf{aircraft} flow fields \citep{luo2025transolveraccurateneuralsolver}, a \textbf{building fire} simulated with FDS \citep{FDS_Mcgrattan2013}, conjugate heat transfer through a multi-region \textbf{heated duct} \citep{OpenFOAM}, and the \textbf{HIVE} induction-heating benchmark \citep{farscape_hive,hornejones2026virtualqualificationnuclearfusion}. Together they span thermal, structural, and aerodynamic physics, sample-dependent and shared meshes, and single- and multi-region domains. The suite also varies how much information the conditioning vector $\boldsymbol{\mu}$ provides beyond the observed geometry. It is largely redundant for the heat sink; partly non-redundant for the bracket and for the building fire and essential for Aircraft, heated dict and HIVE, where operating conditions cannot be recovered from the geometry alone. Further details on the physics, data, and preprocessing are provided in \Cref{app:physics_data}.

\paragraph{Models:}
We compare L-SplineFNO with six representative neural operators for unstructured domains, including mesh-based attention models Transolver \citep{wu2024transolverfasttransformersolver} and GNOT \citep{hao2023gnotgeneralneuraloperator}, latent attention models UPT \citep{alkin2025universalphysicstransformersframework} and MSPT \citep{curvo2026msptefficientlargescalephysical}, and gridded FNO-based models NUNO \citep{liu2023nunogeneralframeworklearningnuno} and GINO \citep{li2023geometryinformedneuraloperatorlargescale}. We also include the analytic SplineFNO to isolate the effect of learning the spline transfer. All models use the same outer conditional pipeline in \Cref{fig:cond-lsfno-pipeline} and match in parameter count ($2.3$ to $2.5$ million parameters), so we replace only the backbone and its associated spatial representation. Full architectural and implementation details are provided in \Cref{app:models}.

\paragraph{Training Objective:}
Given a dataset $\mathcal{D} = \{(\boldsymbol{\mu}^{(i)}, \{\omega_n^{(i)}\}_{n=1}^{N_i},\allowbreak \{y^{(i)}(\omega_n^{(i)})\}_{n=1}^{N_i})\}_{i=1}^{M}$ of conditioning vectors, mesh nodes and target fields, with no input field supplied, we minimise the mean unweighted discrete relative $L_2$ error over the sampled mesh nodes (\Cref{eq:training-objective}). 

\begin{wrapfigure}{r}{0.38\textwidth}
  \begin{equation}
    \mathcal{J}(\theta)
    = \frac{1}{M}\sum_{i=1}^{M}
      \frac{\bigl\|\widehat{\mathbf{y}}^{(i)}-\mathbf{y}^{(i)}\bigr\|_2}
           {\bigl\|\mathbf{y}^{(i)}\bigr\|_2}
    \label{eq:training-objective}
  \end{equation}
  \vspace{-1.0em}
\end{wrapfigure}

Here $\mathbf{y}^{(i)}\in\mathbb{R}^{N_i}$ is the target field at the nodes $\{\omega_n^{(i)}\}_{n=1}^{N_i}$ of sample $i$, and $\|\cdot\|_2$ is the unweighted discrete Euclidean norm over those nodes. We use the same metric for training and its percentage formulation for the final error evaluation in \Cref{tab:film_all_cases_metrics}. All models use the same training recipe: Adam with a starting learning rate of $10^{-3}$, halved every 50 epochs, for 250 epochs.

\subsection{Benchmarks}

\begin{figure}[t]
\centering
\captionsetup{font=small}
\includegraphics[width=0.85\linewidth]{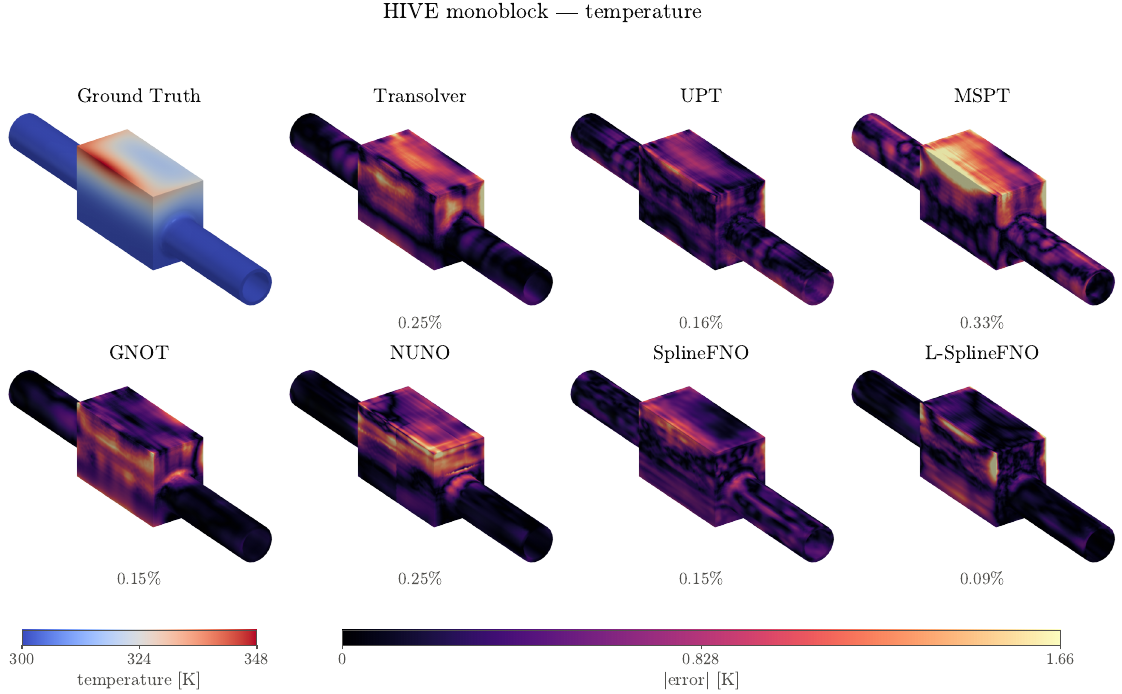}
\caption{Error maps produced by baselines in modelling heat distribution across the HIVE monoblock. L-SplineFNO models the temperature across the structure with $99.9\%$ accuracy (sample-specific error shown).}
\label{fig:hive_error_maps}
\end{figure}

As shown in \Cref{tab:film_all_cases_metrics}, L-SplineFNO occupies a favourable accuracy--cost regime across the benchmark suite: it provides the strongest predictive accuracy on the heat sink, building fire, heated duct, and HIVE cases, and remains close to the leading attention-based models on Aircraft while requiring substantially less training time than MSPT and GNOT. The bracket case is the principal exception, where direct mesh-based attention retains a clear accuracy advantage. The bracket's peak stresses sit in small, sharp concentrations, which the latent grid's averaging and $C^2$ decode flatten, whereas attention acts on the mesh nodes directly (\Cref{fig:bracket_error_maps}). Compared with the analytic SplineFNO, however, learning the spline transfer consistently improves accuracy across all six cases with only a modest change in model size. The faster gridded baselines sacrifice considerably more accuracy, whereas the more accurate attention-based alternatives incur higher training cost; L-SplineFNO therefore combines the efficiency of a structured latent grid with predictive performance competitive with, and often better than, several more computationally demanding unstructured operators. \Cref{fig:heat_sink_error_maps_main} demonstrates how L-SplineFNO achieves near-perfect accuracy in modelling the heat distribution across a heat sink. \Cref{fig:hive_error_maps} shows the error maps produced by each model for the HIVE monoblock case, with L-SplineFNO modelling the heat distribution with $99.92\%$ accuracy. We show further ablations on spline degree and multi-scale scattering in \Cref{app:ablations}.

\begin{figure*}[!t]
    \centering
    \begin{subfigure}[b]{0.49\textwidth}
        \centering
        \includegraphics[width=\linewidth]{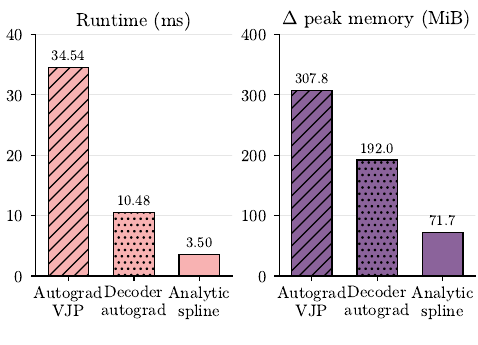}
        \caption{Gradient-evaluation costs for L-SplineFNO.}
        \label{fig:gradient-cost}
    \end{subfigure}\hfill
    \begin{subfigure}[b]{0.49\textwidth}
        \centering
        \includegraphics[width=\linewidth]{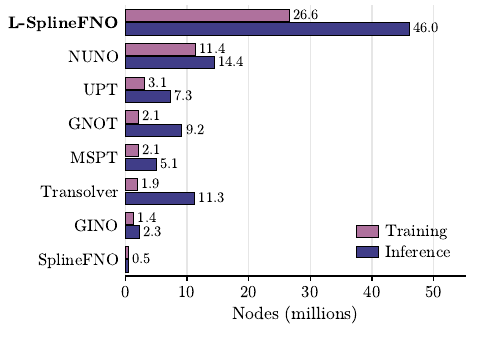}
        \caption{Mesh capacity of models across a single GPU.}
        \label{fig:scalability}
    \end{subfigure}

    \caption{%
        \textbf{Derivative-evaluation cost and scalability.}
        (\subref{fig:gradient-cost}) Mean derivative runtime (left)
        and additional peak GPU allocation (right), averaged
        across the benchmark experiments. Forward preparation is excluded;
        memory is measured above the prepared live state, not total GPU use.
        Full-model autograd VJP measures coordinate sensitivity, whereas
        decoder autograd and analytic spline differentiation evaluate
        pointwise field gradients. Closed-form spline derivatives give exact gradients at a fraction of the cost. 
        (\subref{fig:scalability}) Mean largest successful node count
        across three seeds on an A100 80GB GPU in FP32 with batch size one.
        Training includes a complete Adam step. L-SplineFNO accomodates 26.6M nodes in training and 46.0M in inference, demonstrating superior scalability against the baselines. %
    }
    \label{fig:scalability-gradient-cost}
\end{figure*}

\subsection{Analytic Gradients} %
\label{sec:analytic-gradients}

The analytic decoder of L-SplineFNO yields spatial derivatives without an automatic-differentiation graph over the query coordinates. The decoded field of \Cref{eq:pipeline-output} is
$\widehat{\mathbf{y}}(q)=P_{\mathrm{out}}\sum_{\mathbf{g}\in\mathcal{G}} \mathcal{Y}_{\mathcal{G}}[\mathbf{g}]\,\Phi_{\mathbf{g}}(q)$,
where $\mathcal{Y}_{\mathcal{G}}[\mathbf{g}]$ are the latent-grid coefficients, $\Phi_{\mathbf{g}}$ is the normalised basis (\Cref{eq:spline-basis}), and $P_{\mathrm{out}}\mathbf{z}=W_{\mathrm{out}}\mathbf{z}+\mathbf{b}_{\mathrm{out}}$ is an affine projection.
The bias $\mathbf{b}_{\mathrm{out}}$ vanishes under differentiation, so
\begin{equation}
\partial^{\boldsymbol{n}}\widehat{\mathbf{y}}(q)
=
W_{\mathrm{out}}\sum_{\mathbf{g}\in\mathcal{G}} \mathcal{Y}_{\mathcal{G}}[\mathbf{g}]\,\partial^{\boldsymbol{n}}\Phi_{\mathbf{g}}(q),
\qquad 1\leq|\boldsymbol{n}|\leq 2.
\label{eq:analytic-spline-gradient}
\end{equation}
The derivatives $\partial^{\boldsymbol{n}}\Phi_{\mathbf{g}}$ are evaluated in closed form from the first and second derivatives of the degree-$p$ B-spline, including those of the boundary partition-of-unity normalisation. Each query touches at most $(p+1)^d$ control points, so $N$ queries cost $O(N(p+1)^d)$ with no reverse-mode coordinate tape, and the result stays differentiable in $\mathcal{Y}_{\mathcal{G}}$, keeping derivative-based physics losses trainable.

Full-model \textit{autograd} instead returns the coordinate sensitivity $\partial(\sum_i\widehat{\mathbf{y}}_i)/\partial\omega_j$, which for globally coupled operators differs from $\nabla\widehat{\mathbf{y}}(\omega_j)$ because moving node $j$ can alter predictions elsewhere; we therefore compare against autograd through the same decoder with the latent grid fixed. \Cref{fig:gradient-cost} reports runtime and memory across all benchmark experiments.

\subsection{Scalability}
\label{sec:scalability}

We benchmark all architectures on synthetic three-dimensional point clouds with the HIVE settings, on one NVIDIA A100 80GB GPU in FP32 with batch size one (\Cref{fig:scalability}). Runtime grows near-linearly with node count for every model over its feasible range, so memory decides capacity. L-SplineFNO trains on up to 26.6M nodes and runs inference on 46.0M, well beyond the capacity of existing methods. Details are in \Cref{app:scalability}.

\subsection{Virtual Tokamak of MAST-U}
\label{sec:mastu}

Beyond benchmarks, we apply L-SplineFNO to a whole-device structural modelling of the Mega-Ampere Spherical Tokamak Upgrade (MAST-U), a fusion experiment that informs compact spherical-tokamak power plants \Citep{Milnes2015,Morris2018,Harrison_2024,Chapman2024}. Such simulations, referred to as \emph{virtual tokamaks} load the vessel, centre column, coils and supports together, so each component sees boundary conditions set by the rest of the assembly \Citep{Ellis_2025}. A single static solve of the MAST-U model costs about $500+$ CPU-hours on $800$ cores, challenging the many-query workloads a structural digital twin needs \Citep{neural_parareal_pamela2025}.

\begin{figure}[t]
    \centering
    \includegraphics[width= \linewidth]{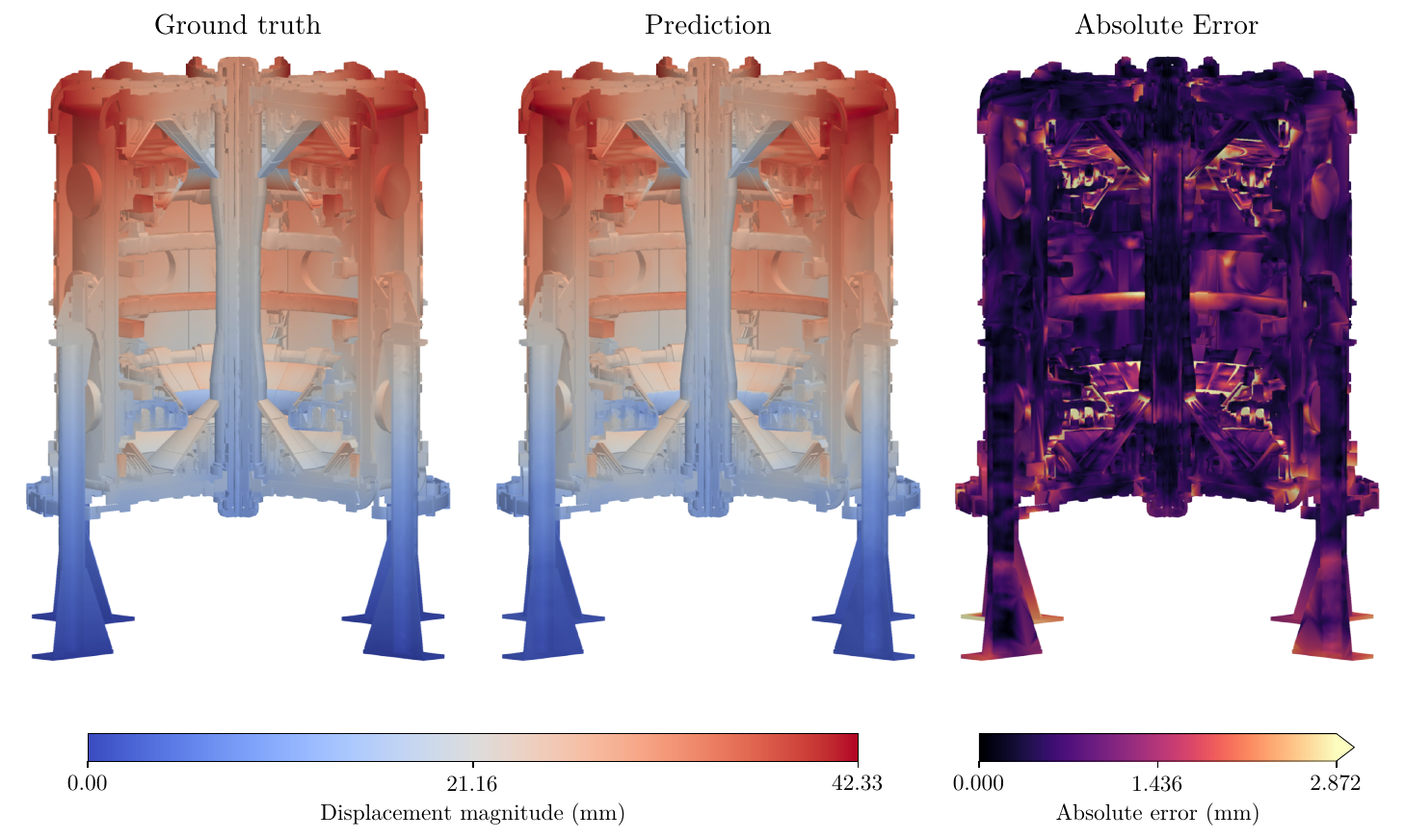}
    \caption{MAST-U virtual tokamak: Displacement magnitude (mm) from the MOOSE simulation (ground truth) and the L-SplineFNO (prediction) and the absolute error across them (mm).}
    \label{fig:mastu-displacement}
\end{figure}

To alleviate this bottleneck, we learn the static displacement field $\mathbf u=(u_x,u_y,u_z)$ of the full MAST-U structural model under thermal expansion, given the thermal expansion coefficient and wall temperature. The task stresses unstructured operators on several axes at once. The mesh has $29.2$M nodes and $134.6$M tetrahedra; merging $3{,}938$ CAD parts leaves element sizes spanning more than four orders of magnitude. Steel, copper and graphite meet at over a million shared interface nodes. L-SplineFNO, with material-specific spline transfers and a shared FNO backbone, is trained on the full mesh with $29.2$M nodes and reaches a relative vector $L_2$ error of $1.8\%$ on a single H100 GPU (\Cref{fig:mastu-displacement}). \Cref{app:mastu} details the device, physics, data, and training.

\paragraph{Spatial bounds and parameter uncertainty.}
The analytic spline decoder enables post-training bounds on the predicted displacement over continuous spatial regions and uncertain operating parameters, without needing additional reference solutions. These bounds provide a rigorous guarantee that the model prediction lies within a specified range, which is critical for safety-critical applications such as nuclear fusion devices.

\emph{Spatial bounds.}
At fixed parameters, each displacement component on a spline cell can be expressed in Bernstein form \citep{Romani2004conversionmatrix} as 

\begin{equation}
\widehat u_j(x)
=\frac{\sum_k Y_{j,k}b_k(x)}{\sum_k D_kb_k(x)},
\qquad
\min_k\frac{Y_{j,k}}{D_k}
\le \widehat u_j(x)\le
\max_k\frac{Y_{j,k}}{D_k}.
\label{eq:spline_bounds}
\end{equation}

\begin{wrapfigure}{l}{0.42\textwidth}
    \centering
    \vspace{-1em}
    \includegraphics[width=\linewidth]{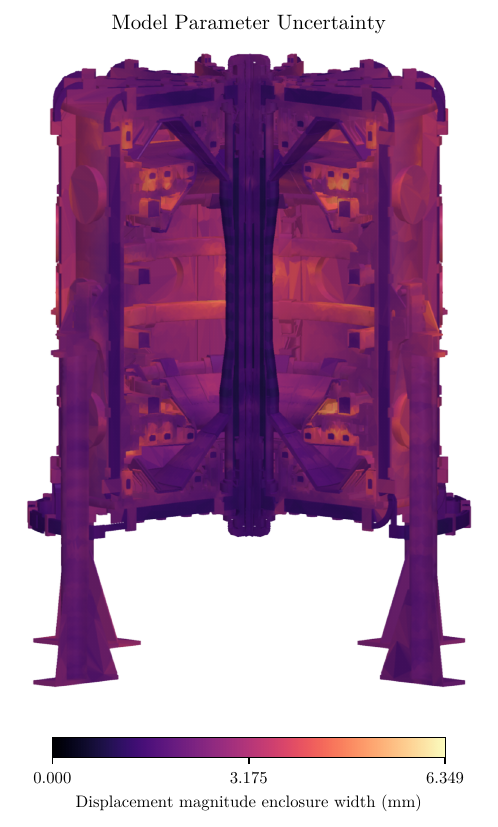}
    \caption{Rigorous enclosure of L-SplineFNO over the uncertain parameter box $\mathcal{B}$, obtained combining a Taylor-remainder bound with the convex-hull property of the spline.}
    \label{fig:mastu-displacement-error}
    \vspace{-2em}
\end{wrapfigure}

Here $Y_{j,k}$ and $D_k$ are the Bernstein coefficients of the decoder numerator and normalisation denominator, respectively, and $b_k$ are the spatial Bernstein basis functions. Since $b_k(x)\ge0$ and $D_k>0$, the prediction is a convex combination of the coefficient ratios $Y_{j,k}/D_k$ \citep{piegl1996nurbs}. Their extrema therefore enclose every point within the cell, including points between mesh nodes. The bound is loose when a cell is large compared with the variation of the field, since the Bernstein coefficients then lie well away from the function values. We tighten it by spatial subdivision. The de Casteljau algorithm splits a cell exactly in half along its longest side and gives each half its own Bernstein coefficients, describing the same spline on a smaller box \citep{Boehm1999deCasteljau}. After $n$ such splits, each cell is covered by $2^n$ sub-boxes. \Cref{eq:spline_bounds} is applied to each sub-box, and as the sub-boxes shrink their coefficients converge to the function values (see \Cref{app:spatial-parameter-bounds} for more details).

\emph{Parameter uncertainty.}
The same enclosure extends to an uncertain parameter box $\mathcal{B}$, where $\boldsymbol{\mu}=(\alpha,T_w)\in\mathcal{B}$. Only the spline coefficients change: at fixed parameters each is a number, whereas over $\mathcal{B}$ each is a function of $\boldsymbol{\mu}$. We obtain these functions by propagating a second-order Taylor model through the fixed trained network \citep{Makino1999Taylor}, so that each physical spline coefficient takes the form

\begin{equation}
\begin{aligned}
&c_i(\boldsymbol{\mu})
=c_i^0+\mathbf a_i^\top\boldsymbol{\delta}
+\tfrac12\boldsymbol{\delta}^\top H_i\boldsymbol{\delta}
+\varepsilon_i, \\[2pt]
&\qquad |\varepsilon_i|\le r_i,
\quad \boldsymbol{\delta}=\boldsymbol{\mu}-\boldsymbol{\mu}_0.
\end{aligned}
\end{equation}

Here $c_i^0$, $\mathbf a_i$, and $H_i$ are the coefficient value, parameter gradient, and parameter Hessian at the expansion centre $\boldsymbol{\mu}_0$, while $r_i$ bounds the remainder uniformly over $\mathcal{B}$. The parameters enter only through the input modulation of \Cref{eq:pipeline-lift}; the affine operations that follow (lifting, scatter, Fourier layers and readout) carry the signed polynomial terms exactly, and only the nonlinear activations add to $r_i$.

The Bernstein transformation in \Cref{eq:spline_bounds} is linear in the spline coefficients, so applying it to these polynomials gives each numerator coefficient as a quadratic in $\boldsymbol{\delta}$ plus a bounded remainder,
\begin{equation}
Y_{j,k}(\boldsymbol{\mu})=Y_{j,k}^0+\pi_{j,k}(\boldsymbol{\delta})+\eta_{j,k},
\qquad |\eta_{j,k}|\le r_{j,k}.
\end{equation}
The transformation acts on the polynomial coefficients \emph{before} any extremum is taken, so all spline coefficients share the same $\boldsymbol{\delta}$ and their parameter contributions can cancel. Minimising and maximising $\pi_{j,k}$ over $\mathcal{B}$ and subtracting or adding $r_{j,k}$ gives $Y_{j,k}^-$ and $Y_{j,k}^+$. The denominator coefficients $D_k$ depend only on geometry, so the convex-hull argument of \Cref{eq:spline_bounds} applies unchanged:

\begin{equation}
\begin{gathered}
\min_k\frac{Y_{j,k}^-}{D_k}
\le \widehat u_j(x,\boldsymbol{\mu})\le
\max_k\frac{Y_{j,k}^+}{D_k} \qquad
\text{for all } x \text{ in the cell and all } \boldsymbol{\mu}\in\mathcal{B}.
\end{gathered}
\end{equation}

Spatial subdivision tightens this enclosure exactly as before. Parameter subdivision is the complementary refinement. It partitions $\mathcal{B}$ and re-centres the Taylor model on each part, which shrinks the nonlinear remainders, and the final enclosure is the union over all parts. In \Cref{fig:mastu-displacement-error} we show the displacement bounds over a parameter box $\mathcal{B}$ such that $\alpha\in[1.4228,\,1.4428]\times10^{-5}\,\mathrm{K}^{-1}$ and $T_w\in[859,869]\,\mathrm{K}$.

These intervals describe the range of the surrogate prediction, including spatial variation, parameter variation and enclosure conservatism. They do not quantify prediction error against a reference numerical solution. Here we present the validated binary64 (FP64) results, which carry an end-to-end rounding certificate: outward rounding accounts for arithmetic error, so the enclosure contains the real-arithmetic surrogate defined by the stored weights. The same enclosure formulas can also be evaluated in ordinary FP32 or FP64, which is cheaper but gives no rounding certificate. All three are compared in \Cref{tab:app_mastu_parameter_arithmetic}, and \Cref{app:spatial-parameter-bounds} gives the construction and arithmetic assumptions.

\section{Discussion}
\label{sec:discussion}

\paragraph{Strengths.}
L-SplineFNO constructs a structure-preserving learnable transfer that maintains compact support, partition of unity and $C^{p-1}$ smoothness. This allows the model to scale linearly in both memory and compute for mesh size. L-SplineFNO achieves better performance on most benchmarks while remaining training-compute efficient and maintaining accuracy on super-resolution tasks. Because the transfer is parameter-efficient, it can be specialised per material or region at negligible cost. The explicit spline reconstruction provides inexpensive analytic gradients, useful for sensitivity analysis and physics-based objectives. The same reconstruction also yields rigorous bounds on the prediction over whole cells, not just at mesh nodes, and over boxes of uncertain parameters. Mesh-native and attention-based operators have no continuous reconstruction, so they provide no comparable guarantee.

\paragraph{Limitations.}
The spline transfer can smooth features in the target, such as kinks at material interfaces and stress concentrations around small geometric features, where mesh-native attention remains more accurate. Per-region output projections recover interface kinks, but the scatter still mixes regions on a shared latent grid; interface-aware or deformed transfers and local graph corrections near thin features could address this \citep{GeoFNO_Li2024,li2025geometricoperatorlearningoptimal,li2023geometryinformedneuraloperatorlargescale}. The fixed linear decoder also limits each output to the span of the spline basis, which sets an accuracy floor that no encoder or backbone can overcome, most visibly at sharp interfaces and on surfaces embedded in a volumetric grid. Finally, stencil cost grows rapidly with spline degree in three dimensions, so higher degrees exhaust memory on large meshes without relieving the smoothness bias (see \cref{app:ablations}). Adaptive refinement of latent cells only where geometry or error demands it, as in NUNO \citep{liu2023nunogeneralframeworklearningnuno}, could ease both the accuracy floor and the memory cost while keeping cubic stencils. The enclosures bound the surrogate, not its error, and are conservative on account of the rigour. They loosen with cell size, parameter-box width and network depth; subdivision tightens them at proportional cost for a fixed geometry.

\paragraph{Future work.}

Several extensions follow directly from the spline structure.
First, interface kinks could be represented exactly. Our decoder is a cubic B-spline basis, which is $C^{p-1}$ at every knot; repeating a knot lowers the continuity at that location alone, so three coincident knots permit a jump in gradient and four a jump in value \citep{deBoor1978,piegl1996nurbs}. Isogeometric analysis, which solves PDEs directly in a spline basis, uses exactly this to resolve material interfaces \citep{Hughes2005isogeometric}. Second, the closed-form derivatives make physics-informed training inexpensive \citep{raissi2019physicsinformed,li2023physicsinformedneuraloperatorlearning}; on multi-region problems, PDE-residual losses could be paired with penalties on flux continuity across interfaces, as in domain-decomposed physics-informed networks \citep{jagtap2020conservative}. Third, the parameter enclosures supply a natural acquisition function for active learning, placing new simulations where the enclosure is widest \citep{musekamp2025activelearningneuralpde,polanska2026dataefficientneuraloperatortraining}. Fourth, because FiLM conditions on a global vector, time can enter as one more component of $\boldsymbol{\mu}$, as in time-conditioned PDE foundation models \citep{herde2024poseidonefficientfoundationmodels,takamoto2023learningneuralpdesolvers}; this would extend the framework from steady states to transient fields, complementing an operator-split formulation \citep{gopakumar2026physicaloperators}. Finally, the enclosures currently bound the surrogate rather than the numerical simulation it approximates. Widening them by a calibrated bound on the surrogate-to-solver error, for instance through conformal prediction on held-out simulations \citep{gopakumar2025calibrated}, would give intervals that contain the simulated field over the whole parameter box with a stated probability.

\subsubsection*{Acknowledgments}
The Fusion Futures Programme partially funds Vignesh Gopakumar and Ander Gray. As announced by the UK Government in October 2023, Fusion Futures aims to provide holistic support for the development of the fusion sector. Ander Gray is partially funded by the Academic and Research Chaire “Architecture des Systèmes Complexes” Dassault Aviation, Naval Group, Dassault Systèmes, KNDS France,
Agence de l’Innovation de Défense, Institut Polytechnique de Paris. Anima Anandkumar is supported in part by Bren endowed chair, ONR (MURI grant N00014-23-1-2654), and the AI2050 senior fellow program at Schmidt Sciences. The authors acknowledge funding from the Fusion Computing Lab collaboration between the Hartree Centre and UKAEA. This work was performed using resources provided by the Cambridge Service for Data Driven Discovery (CSD3) operated by the University of Cambridge Research Computing Service (www.csd3.cam.ac.uk), provided by Dell EMC and Intel using Tier-2 funding from the Engineering and Physical Sciences Research Council (capital grant EP/T022159/1), and DiRAC funding from the Science and Technology Facilities Council (www.dirac.ac.uk). 

\newpage
\bibliography{../references}
\bibliographystyle{tmlr}

\newpage
\appendix
\section{Notation}
\label{app:notation}

\Cref{tab:notation} lists the symbols used across the paper, grouped by topic,
with a pointer to where each is defined. Bold lowercase denotes a vector at one
node, query point or sample ($\mathbf{h}_i$, $\boldsymbol{\mu}$); calligraphic
letters denote collections over nodes or grid points, operators and sets
($\mathcal{X}$, $\mathcal{G}$, $\mathcal{N}$); a subscript $\mathcal{G}$ marks a
latent-grid quantity, and a hat marks a prediction ($\widehat{y}$). A
parenthesised superscript indexes a coordinate axis on coordinates
($\omega_i^{(\ell)}$) and a sample on dataset quantities
($\mathbf{y}^{(i)}$); it never denotes a power. Symbols local to one
derivation, such as the intermediate Taylor-model terms in
\Cref{app:spatial-parameter-bounds} and the internals of individual baselines
in \Cref{app:models}, are defined where they appear. Symbols with more than one
meaning are collected in \Cref{tab:notation-overloads}.

\begin{longtable}{
>{\raggedright\arraybackslash}p{0.27\linewidth}
>{\raggedright\arraybackslash}p{0.66\linewidth}
}
\caption{Notation.}\label{tab:notation}\\
\toprule
Symbol & Meaning \\
\midrule
\endfirsthead
\toprule
Symbol & Meaning \\
\midrule
\endhead
\bottomrule
\endfoot

\multicolumn{2}{@{}l}{\textit{Problem and data}} \\
\addlinespace[2pt]
$\Omega\subset\mathbb{R}^{d}$, $d$ & Spatial domain and its dimension ($d=3$ throughout). \\
$\mathbf{x}\in\Omega$ & Generic continuous point; written $x$ in the enclosure formulas. \\
$\boldsymbol{\mu}\in\mathcal{P}\subset\mathbb{R}^{k}$ & Conditioning vector of design, load, material and operating parameters. \\
$\mathcal{U}$, $u_0$ & Solution space (\Cref{eq:conditional-solution-operator}); optional per-node input field, absent in every benchmark ($c_{\mathrm{in}}=0$). \\
$\mathcal{L}_{\boldsymbol{\mu},\Omega}$ & Governing PDE operator. \\
$\mathcal{N}$, $\mathcal{N}_\theta$ & Exact solution operator and learned surrogate, $(\boldsymbol{\mu},\Omega)\mapsto y$ and $\widehat{y}$. \\
$y$, $\widehat{y}$ & Reference and predicted steady-state field. \\
$\mathbf{y}^{(i)}$, $\widehat{\mathbf{y}}^{(i)}\in\mathbb{R}^{N_i}$ & The same fields at the $N_i$ nodes of sample $i$. \\
$\mathcal{D}$, $M$, $M_{\mathrm{test}}$ & Dataset of $M$ samples and the size of its test split. \\
$\mathcal{J}(\theta)$, $\mathcal{E}$ & Training objective (\Cref{eq:training-objective}) and reported test error in percent, in physical units (\Cref{eq:app_relL2}). \\
$\|\cdot\|_2$, $\|\cdot\|_\infty$ & Unweighted Euclidean norm over one sample's nodes; largest absolute entry. \\

\addlinespace[4pt]
\multicolumn{2}{@{}l}{\textit{Mesh and latent grid}} \\
\addlinespace[2pt]
$N$, $\omega_i$, $\bar{\omega}_i$ & Number of mesh nodes, coordinate of node $i$, and its normalisation to $[0,1]^d$. \\
$i$, $\ell$ & Mesh-node index $i\in\{1,\ldots,N\}$; coordinate-axis index $\ell\in\{1,\ldots,d\}$. \\
$q_j$, $N_q$ & Output query points and their number; by default the input nodes. \\
$\mathcal{G}$, $\mathbf{g}$ & Latent grid $\prod_{\ell}\{0,\ldots,S_\ell-1\}$ and a multi-index $\mathbf{g}=(g_1,\ldots,g_d)$ in it. \\
$S_\ell$, $G$ & Control points along axis $\ell$, and in total, $G=|\mathcal{G}|=\prod_{\ell}S_\ell$. \\
$\xi_0^{(\ell)}$, $\Delta_\ell$ & Grid origin and spacing along axis $\ell$; $\xi_{\mathbf{g}}^{(\ell)}=\xi_0^{(\ell)}+g_\ell\Delta_\ell$. \\
$t_i^{(\ell)}$ & Continuous grid coordinate of node $i$, $(\omega_i^{(\ell)}-\xi_0^{(\ell)})/\Delta_\ell$. \\

\addlinespace[4pt]
\multicolumn{2}{@{}l}{\textit{Features and conditioning} (\Cref{sec:lsfno-pipeline,app:conditioning_methods})} \\
\addlinespace[2pt]
$c$, $c_{\mathrm{in}}$, $c_{\mathrm{out}}$ & Model width; input-field and output channel counts. \\
$L$, $B_{\mathrm{batch}}$ & Number of backbone layers; batch size. \\
$\varphi$, $n_f$, $c_f$ & Integrated sinusoidal encoding of $\bar{\omega}$ at $n_f=16$ frequencies, of width $c_f=2dn_f=96$; with $\bar{\omega}$ appended, the lift sees 99 channels. \\
$\mathbf{f}_i$, $\tilde{\mathbf{f}}_i$ & Encoded features $\varphi(\bar{\omega}_i)$ and their FiLM-modulated form (\Cref{eq:pipeline-lift}). \\
$E_\mu$, $e_\mu$ & Conditioning embedding network and its output $E_\mu(\boldsymbol{\mu})$. \\
$H_\gamma$, $H_\beta$, $H_\mu$ & FiLM scale and shift heads; $H_\mu$ denotes the full map $\boldsymbol{\mu}\mapsto(\boldsymbol{\gamma}_\mu,\boldsymbol{\beta}_\mu)$. \\
$\boldsymbol{\gamma}_\mu$, $\boldsymbol{\beta}_\mu\in\mathbb{R}^{c_f}$ & FiLM scale and shift, broadcast over nodes; exactly $\mathbf{1}$ and $\mathbf{0}$ at initialisation. \\
$\odot$, $\mathbf{1}_N$ & Channel-wise product broadcast over nodes; length-$N$ vector of ones. \\
$P_{\mathrm{in}}$, $P_\rho$, $P_{\mathrm{out}}$ & Pointwise lift, density-augmented grid projection (\Cref{eq:pipeline-encode}) and output projection. \\
$\mathbf{h}_i$, $\mathcal{X}\in\mathbb{R}^{N\times c}$ & Lifted, conditioned features at node $i$, stacked as the rows of $\mathcal{X}$. \\
$\mathcal{X}_{\mathcal{G}}$, $\mathcal{Y}_{\mathcal{G}}$ & FNO input and output on the latent grid (\Cref{eq:spline-encode,eq:pipeline-encode}). \\
$\mathcal{Y}$, $\mathbf{z}_j$, $\widehat{\mathbf{y}}_j$ & Decoded features at the mesh nodes; decoded latent at $q_j$ and prediction $P_{\mathrm{out}}\mathbf{z}_j$ (\Cref{eq:pipeline-output}). \\
$\mathcal{F}$, $\mathcal{F}_\theta$ & Fourier transform (\Cref{eq:fno-layer}); grid-space FNO. \\
$\theta$ & All trainable parameters; $\kappa_\theta^{(\ell)}$ and $\mathcal{F}_\theta$ use the scatter and FNO parts. \\

\addlinespace[4pt]
\multicolumn{2}{@{}l}{\textit{Spline basis and learnable scatter} (\Cref{sec:splinefno,sec:l-splinefno,app:spline-construction})} \\
\addlinespace[2pt]
$B_p$, $B$, $p$ & Centred cardinal B-spline of degree $p$ (\Cref{eq:cardinal-cubic-definition}); $B:=B_p$ at fixed degree, $p=3$ by default. \\
$R_p$, $t$ & Support radius $(p+1)/2$; signed distance along one axis in grid cells. \\
$\Phi$, $\widetilde{\Phi}\in\mathbb{R}^{N\times G}$ & Row-normalised and raw spline basis matrices (\Cref{eq:spline-basis}); $\Phi_{\mathbf{g}}(q)$ is basis function $\mathbf{g}$ at a query point. \\
$\varphi^{(\ell)}_{g_\ell}$ & One-dimensional factor of $\Phi_{\mathbf{g}}$ (\Cref{eq:tensor-product-basis}). \\
$\lambda$ & Tikhonov coefficient of the SplineFNO solve (\Cref{eq:spline-encode}). \\
$\kappa_\theta^{(\ell)}$ & Learned one-dimensional scatter kernel on axis $\ell$ (\Cref{eq:learnable-spline-kernel}). \\
$J$, $\tau_j$, $h$, $\zeta^{(\ell)}_j$ & Number, fixed centres, relative width and coefficients of the compact corrections in $\kappa_\theta^{(\ell)}$. \\
$A$, $Q$, $u_q$ & Partition-of-unity constraint matrix $A\in\mathbb{R}^{Q\times J}$ at collocation points $u_q\in[0,1)$ (\Cref{eq:partition-unity-matrix}). \\
$V_0$, $\alpha_\theta^{(\ell)}$, $r$ & Null-space basis $V_0\in\mathbb{R}^{J\times r}$ of $A$ and free coordinates $\alpha_\theta^{(\ell)}\in\mathbb{R}^{r}$, $\zeta^{(\ell)}=V_0\alpha_\theta^{(\ell)}$ (\Cref{eq:null-space-parameterisation}). \\
$s$ & Number of scatter scales in the multiscale encoder. \\
$\mathcal{O}_p(u)$ & Active stencil offsets at fractional cell position $u$ (\Cref{eq:degree-p-offset-set}); $\mathcal{O}_3=\{-1,0,1,2\}$. \\
$\mathcal{S}_\ell(i)$, $\mathcal{S}(i)$ & Valid grid indices of node $i$ on axis $\ell$ (\Cref{eq:valid-degree-p-stencil}) and their product, at most $(p+1)^d$ points. \\
$W_\theta\in\mathbb{R}^{N\times G}$ & Sparse learned transfer (\Cref{eq:learnable-transfer-weights}); applied by indexed scatter-add, never materialised. \\
$n_{\mathbf{g}}$, $d_{\mathbf{g}}$, $a_{\mathbf{g}}$ & Weighted feature sum, signed weight sum and absolute weight sum at grid point $\mathbf{g}$. \\
$\widetilde{d}_{\mathbf{g}}$, $D_\theta$ & Safeguarded denominator (\Cref{eq:stable-scatter-denominator}) and $D_\theta=\operatorname{diag}(\widetilde{d}_{\mathbf{g}})$. \\
$\operatorname{sgn}_{+}$, $\eta$, $\varepsilon$ & Sign function with $\operatorname{sgn}_{+}(0)=+1$; relative and absolute denominator floors. \\
$\widetilde{\mathcal{X}}_{\mathcal{G}}$ & Normalised scatter $D_\theta^{-1}W_\theta^\top\mathcal{X}$ (\Cref{eq:learnable-normalised-scatter}). \\
$\rho_{\mathbf{g}}$ & Analytic sampling density $\sum_i\Phi_{i,\mathbf{g}}$ (\Cref{eq:spline-density-channel}). \\

\addlinespace[4pt]
\multicolumn{2}{@{}l}{\textit{Spatial and parameter enclosures} (\Cref{sec:mastu,app:spatial-parameter-bounds})} \\
\addlinespace[2pt]
$\widehat{\mathbf{u}}(x,\boldsymbol{\mu})$, $\widehat{u}_j$ & Predicted displacement and its component $j\in\{1,2,3\}$; $\boldsymbol{\mu}$ is dropped at fixed parameters. \\
$Y_j(x)$, $D(x)$ & Numerator and geometry-only denominator of the rational decoder, $\widehat{u}_j=Y_j/D$; $D=1$ wherever spline support is untruncated. \\
$k$, $b_k(x)$ & Bernstein multi-index in $\{0,\ldots,p\}^d$ and tensor-product Bernstein polynomial on the current box; $b_k\ge0$, $\sum_k b_k=1$. \\
$Y_{j,k}$, $D_k$ & Bernstein coefficients of $Y_j$ and $D$ (\Cref{eq:spline_bounds}); the enclosure requires $D_k>0$. \\
$w_k(x)$ & Convex weight $D_kb_k(x)/D(x)$, so $\widehat{u}_j=\sum_k w_k Y_{j,k}/D_k$. \\
$[L_j,U_j]$ & Component enclosure; at fixed $\boldsymbol{\mu}$, the $\min_k$ and $\max_k$ of $Y_{j,k}/D_k$. \\
$n$ & de Casteljau subdivision depth, giving $2^n$ sub-boxes per cell. \\
$m^-$, $m^+$ & Lower and upper bounds on $\|\widehat{\mathbf{u}}\|_2$ derived from the component intervals. \\
$\mathcal{B}\subset\mathcal{P}$ & Axis-aligned parameter box, partitioned into leaves with their own expansion centres. \\
$\alpha$, $T_w$ & MAST-U parameters: thermal expansion coefficient ($\mathrm{K}^{-1}$) and wall temperature ($\mathrm{K}$). \\
$\boldsymbol{\mu}_0$, $\boldsymbol{\delta}$ & Taylor expansion centre and offset $\boldsymbol{\mu}-\boldsymbol{\mu}_0$. \\
$c_i(\boldsymbol{\mu})$; $c_i^0$, $\mathbf{a}_i$, $H_i$ & Scalar physical spline coefficient; its value, gradient and Hessian at $\boldsymbol{\mu}_0$. \\
$\varepsilon_i$, $r_i$ & Taylor remainder of $c_i$ and its uniform bound over the box. \\
$Y_{j,k}^0$, $\pi_{j,k}(\boldsymbol{\delta})$ & Nominal Bernstein numerator coefficient and its quadratic in $\boldsymbol{\delta}$. \\
$\eta_{j,k}$, $r_{j,k}$ & Remainder after the Bernstein transform and its uniform bound. \\
$\pi_{j,k}^{\pm}$, $Y_{j,k}^{\pm}$ & Extrema of $\pi_{j,k}$ over the box, and numerator bounds $Y_{j,k}^0+\pi_{j,k}^{\pm}\pm r_{j,k}$. \\

\addlinespace[4pt]
\multicolumn{2}{@{}l}{\textit{Baselines} (\Cref{app:models})} \\
\addlinespace[2pt]
$Q$, $K$, $V$, $c_h$ & Query, key and value matrices of an attention block; head dimension. \\
$\varsigma$ & Feature-wise softmax in GNOT's linear attention. \\
$n_s$ & Transolver slices or UPT latent tokens, $n_s\ll N$. \\
$n_b$, $n_e$ & GNOT input branches ($n_b=1$ here) and mixture experts. \\
$n_p$, $n_q$ & MSPT patches and supernodes pooled per patch. \\
$n_c$ & NUNO K-D tree subdomains. \\
$r$, $\mathcal{N}_r$ & GINO graph radius in normalised coordinates and the neighbourhood it induces. \\
$\kappa_{\mathrm{G}}$, $\gamma_{\mathrm{G}}$, $\beta_{\mathrm{G}}$ & GINO graph-kernel MLP and adaptive-instance-norm scale and shift. \\
$F$ & Number of sinusoidal frequencies in the UPT and GINO embeddings. \\
$n_\tau$, $c_\tau$ & Cross-attention parameter tokens and their width. \\
\end{longtable}

\begin{table}[H]
\centering
\small
\caption{Symbols with more than one meaning. The context, given in the last
column, always fixes which is meant.}
\label{tab:notation-overloads}
\begin{tabular}{@{}l>{\raggedright\arraybackslash}p{0.27\linewidth}>{\raggedright\arraybackslash}p{0.52\linewidth}@{}}
\toprule
Symbol & Main meaning & Other meaning (context) \\
\midrule
$i$ & mesh node & dataset sample (training objective, \Cref{app:physics_data}); scalar coefficient $c_i$ (enclosures) \\
$j$ & query point $q_j$ & correction index in $\kappa_\theta^{(\ell)}$; displacement component (enclosures) \\
$k$ & $\dim\boldsymbol{\mu}$ & Bernstein multi-index (enclosures) \\
$q$ & query point $q_j$ & collocation index $u_q$ (\Cref{app:learnable-spline-construction}) \\
$p$ & spline degree & static pressure (aircraft) \\
$c$ & model width & spline coefficient $c_i(\boldsymbol{\mu})$ (enclosures) \\
$B$ & cardinal B-spline & parameter box $\mathcal{B}$ (enclosures) \\
$\Phi$ & spline basis & normal CDF in GELU (enclosures) \\
$\varphi$, $\phi$ & positional encoding & 1D spline factor $\varphi^{(\ell)}_{g_\ell}$; normal density $\phi$ (enclosures) \\
$D$ & scatter normaliser $D_\theta$ & decoder denominator $D(x)$, $D_k$ (enclosures) \\
$H$ & FiLM heads $H_\gamma$, $H_\beta$ & Hessian $H_i$ (enclosures) \\
$\alpha$ & null-space coordinates $\alpha_\theta^{(\ell)}$ & thermal expansion coefficient (MAST-U) \\
$L$ & backbone depth & lower endpoint $L_j$; linear Taylor term (enclosures) \\
$Q$ & collocation count & attention query (baselines); quadratic Taylor term (enclosures) \\
$r$ & null-space dimension & GINO radius; remainder bounds $r_i$, $r_{j,k}$ \\
$s$, $t$ & scatter scales; grid distance & Taylor-term bounds (enclosures) \\
$\eta$, $\varepsilon$ & scatter floors & remainders $\eta_{j,k}$, $\varepsilon_i$ (enclosures) \\
$g$ & grid index $g_\ell$ & scalar activation (enclosures) \\
$\sigma$ & activation (\Cref{eq:fno-layer}) & von Mises stress (bracket) \\
$\rho$ & sampling density & fluid density (aircraft) \\
$\kappa$ & scatter kernel $\kappa_\theta^{(\ell)}$ & thermal conductivity (\Cref{app:physics_data}); GINO kernel $\kappa_{\mathrm{G}}$ \\
$\mathcal{N}$ & solution operator & GINO neighbourhood $\mathcal{N}_r$ \\
$\mathcal{O}$ & stencil offsets $\mathcal{O}_p$ & complexity $O(\cdot)$ \\
\bottomrule
\end{tabular}
\end{table}

\newpage
\section{\textit{Learnable-}Spline Construction}
\label{app:spline-construction}

The spline transfers used in this work are built from one-dimensional cardinal
B-spline kernels and lifted to $d$ dimensions by tensor products. SplineFNO
uses the fixed analytic spline $B_p$ of degree $p$, while L-SplineFNO augments
this kernel with a constrained learnable correction. For clarity, the main
text presents the default cubic case $p=3$, while the construction below is
stated for a general spline degree $p$. The higher-degree variants considered
in the spline-degree ablation follow the same construction with the
corresponding support and smoothness.

This appendix proceeds in three parts. \Cref{sec:tensor-product} establishes
the tensor-product properties of the analytic degree-$p$ basis.
\Cref{app:learnable-spline-construction} introduces the constrained learned
kernel and shows how compact support, partition of unity, and spline
smoothness are retained. Non-negativity is not imposed on the learned
correction, and \Cref{app:stable-spline-scatter} introduces the safeguarded
normalised scatter used when signed weights cancel.

\subsection{From 1D to \texorpdfstring{$d$}{d}-D: Tensor-Product Construction}
\label{sec:tensor-product}

Let $B_p:\mathbb{R}\to\mathbb{R}$ denote the centred cardinal B-spline of
degree $p$. We write $B:=B_p$ when the degree is fixed. The one-dimensional
basis satisfies
\begin{enumerate}
\item[(P1)] compact support,
$\mathrm{supp}\,B\subseteq[-R_p,R_p]$ with
$R_p=(p+1)/2$;
\item[(P2)] partition of unity,
$\sum_{n\in\mathbb{Z}}B(t-n)=1$ for all $t$;
\item[(P3)] smoothness,
$B\in C^{p-1}(\mathbb{R})$;
\item[(P4)] non-negativity,
$B(t)\geq0$.
\end{enumerate}
For the default cubic case $p=3$, $R_p=2$ and $B\in C^2$, with
\begin{equation}
B_3(t)
=
\begin{cases}
\tfrac{1}{6}\left(4-6t^2+3|t|^3\right), & 0\le|t|<1,\\[1mm]
\tfrac{1}{6}(2-|t|)^3, & 1\le|t|<2,\\[1mm]
0, & |t|\ge2.
\end{cases}
\label{eq:cardinal-cubic-definition}
\end{equation}

Writing $\{\xi^{(\ell)}_{g_\ell}\}_{g_\ell=0}^{S_\ell-1}$ for the control
points along axis $\ell$, define
\begin{equation}
\varphi^{(\ell)}_{g_\ell}(x^{(\ell)})
=
B\!\left(
\frac{x^{(\ell)}-\xi^{(\ell)}_{g_\ell}}{\Delta_\ell}
\right),
\qquad
\Phi_{\mathbf{g}}(\mathbf{x})
=
\prod_{\ell=1}^{d}
\varphi^{(\ell)}_{g_\ell}(x^{(\ell)}).
\label{eq:tensor-product-basis}
\end{equation}
Row $i$ of $\Phi$ stores $\Phi_{\mathbf{g}}(\omega_i)$ over the active
multi-indices.

\paragraph{Compact support and sparsity.}
A product vanishes whenever any factor does, so
$\Phi_{\mathbf{g}}(\mathbf{x})=0$ unless
\[
\left|
\frac{x^{(\ell)}-\xi^{(\ell)}_{g_\ell}}{\Delta_\ell}
\right|
<
R_p
\]
along every coordinate axis. Since a degree-$p$ cardinal spline has at most
$p+1$ active shifts along each axis,
\begin{equation}
\#\{
\mathbf{g}
:
\Phi_{\mathbf{g}}(\mathbf{x})\neq0
\}
\leq
(p+1)^d.
\label{eq:sparsity-bound}
\end{equation}
The default cubic case therefore gives a $4^d$ stencil.

\paragraph{Partition of unity.}
Because each axis sum is independent of the remaining indices,
\begin{equation}
\sum_{\mathbf{g}}
\Phi_{\mathbf{g}}(\mathbf{x})
=
\prod_{\ell=1}^{d}
\left(
\sum_{g_\ell}
\varphi^{(\ell)}_{g_\ell}(x^{(\ell)})
\right)
=
1.
\label{eq:nd-partition-of-unity}
\end{equation}
The reconstruction therefore reproduces constant control-grid fields exactly
in the interior. Near a domain boundary, part of the infinite-grid stencil
falls outside $\mathcal{G}$. Renormalising the surviving basis values restores
partition of unity over the valid stencil.

\paragraph{Smoothness.}
Since the tensor-product factors depend on disjoint coordinates,
\begin{equation}
\partial^{\boldsymbol{n}}
\Phi_{\mathbf{g}}(\mathbf{x})
=
\prod_{\ell=1}^{d}
\frac{
d^{n_\ell}
\varphi^{(\ell)}_{g_\ell}
}{
d(x^{(\ell)})^{n_\ell}
}
(x^{(\ell)}).
\end{equation}
Each factor is continuous whenever $n_\ell\leq p-1$, so the tensor-product
basis inherits the $C^{p-1}$ smoothness of the one-dimensional spline.

\paragraph{Non-negativity.}
A product of non-negative factors is non-negative. Together with
\Cref{eq:nd-partition-of-unity}, the active basis values form a convex
combination, so analytic spline reconstruction cannot overshoot the range of
the active control-point values.

\paragraph{Implication for the transformation.}
Partition of unity makes a constant field exactly representable,
$\Phi(c\mathbf{1}_G)=c\mathbf{1}_N$. Provided $\Phi$ has full column rank,
the regularised transformation in \Cref{eq:spline-encode} recovers this
constant in the limit $\lambda\to0^+$. The compact support in
\Cref{eq:sparsity-bound} also bounds the number of basis evaluations per
mesh point by $(p+1)^d$, independently of $G$.

\subsection{From Analytic to Learned Kernels}
\label{app:learnable-spline-construction}

L-SplineFNO augments the fixed degree-$p$ spline with compact spline
corrections. Along coordinate axis $\ell$,
\begin{equation}
\kappa_{\theta}^{(\ell)}(t)
=
B(t)
+
\sum_{j=1}^{J}
\zeta_j^{(\ell)}
B\!\left(
\frac{t-\tau_j}{h}
\right),
\qquad
\zeta^{(\ell)}
=
V_0\alpha_{\theta}^{(\ell)},
\label{eq:learnable-spline-kernel-app}
\end{equation}
where the correction centres $\tau_j$ and relative width $h$ are fixed and
only $\alpha_{\theta}^{(\ell)}$ is trained. The cubic construction shown in
the main text corresponds to $p=3$.

\paragraph{Compact support and smoothness (P1, P3).}
Let $R_p=(p+1)/2$. We choose the correction centres and width to satisfy
\begin{equation}
|\tau_j|
\leq
R_p(1-h),
\qquad
0<h<1.
\label{eq:correction-support-condition}
\end{equation}
The support of each correction is then contained within
$[-R_p,R_p]$, so $\kappa_\theta^{(\ell)}$ retains the same bounded stencil
as the analytic spline. Since both the base spline and every correction are
degree-$p$ splines, the learned kernel remains $C^{p-1}$. These two
properties hold independently of the learned coefficients.

\paragraph{Partition of unity in the interior (P2).}
For a fractional cell position $u\in[0,1)$, let
\begin{equation}
\mathcal{O}_p(u)
=
\left\{
o\in\mathbb{Z}
:
B(u-o)\neq0
\right\}
\label{eq:degree-p-offset-set}
\end{equation}
denote the active analytic stencil. Its cardinality is at most $p+1$, and
\[
\sum_{o\in\mathcal{O}_p(u)}
B(u-o)
=
1.
\]
For $p=3$, this reduces to the familiar four-point stencil
$\mathcal{O}_3=\{-1,0,1,2\}$ for $u\in[0,1)$.

The learned kernel retains partition of unity when its correction has zero
stencil sum,
\begin{equation}
\sum_{j=1}^{J}
\zeta_j^{(\ell)}
\sum_{o\in\mathcal{O}_p(u)}
B\!\left(
\frac{u-o-\tau_j}{h}
\right)
=
0.
\label{eq:learnable-correction-zero-sum}
\end{equation}
Given collocation points $\{u_q\}_{q=1}^{Q}\subset[0,1)$, define
\begin{equation}
A_{qj}
=
\sum_{o\in\mathcal{O}_p(u_q)}
B\!\left(
\frac{u_q-o-\tau_j}{h}
\right),
\qquad
A\in\mathbb{R}^{Q\times J}.
\label{eq:partition-unity-matrix}
\end{equation}
Let $V_0\in\mathbb{R}^{J\times r}$ contain the right singular vectors
spanning the numerical null space of $A$. We parameterise
\begin{equation}
\zeta^{(\ell)}
=
V_0\alpha_{\theta}^{(\ell)},
\qquad
\alpha_{\theta}^{(\ell)}
\in
\mathbb{R}^{r},
\label{eq:null-space-parameterisation}
\end{equation}
which gives $A\zeta^{(\ell)}\simeq0$ to numerical precision for every value of
the learned parameters. The matrix $V_0$ is computed once and remains fixed
throughout training. We additionally verify
\Cref{eq:learnable-correction-zero-sum} on a dense set of fractional
coordinates.

\paragraph{Partition of unity at the boundary (P2).}
For mesh point $\omega_i$, define the valid one-dimensional stencil
\begin{equation}
\mathcal{S}_\ell(i)
=
\left\{
g_\ell\in\{0,\ldots,S_\ell-1\}
:
\left|
t_i^{(\ell)}-g_\ell
\right|
<
R_p
\right\}.
\label{eq:valid-degree-p-stencil}
\end{equation}
The tensor-product transfer weights are
\begin{equation}
(W_\theta)_{i,\mathbf{g}}
=
\prod_{\ell=1}^{d}
\frac{
\kappa_\theta^{(\ell)}
\!\left(
t_i^{(\ell)}-g_\ell
\right)
}{
\displaystyle
\sum_{q\in\mathcal{S}_\ell(i)}
\kappa_\theta^{(\ell)}
\!\left(
t_i^{(\ell)}-q
\right)
},
\qquad
\mathbf{g}\in\mathcal{S}(i),
\label{eq:learnable-transfer-weights-app}
\end{equation}
and zero otherwise, where
$\mathcal{S}(i)=\prod_{\ell=1}^{d}\mathcal{S}_\ell(i)$.
The explicit normalisation restores partition of unity over the surviving
entries whenever the denominator is nonzero. Each mesh point therefore
interacts with at most $(p+1)^d$ latent-grid points.

\paragraph{Non-negativity is surrendered (P4).}
The null-space parameterisation does not constrain the sign of the learned
correction. Consequently, $\kappa_\theta^{(\ell)}$ may contain negative
lobes. The learned transfer retains compact support, $C^{p-1}$ smoothness,
and partition of unity to numerical precision, but no longer has the
convex-combination property of the analytic spline. Signed contributions may
therefore cancel when they are accumulated at a latent-grid point.

\subsection{Stable Normalised Scatter and Density Feature}
\label{app:stable-spline-scatter}

The main text defines
\[
n_{\mathbf{g}}
=
\sum_{i=1}^{N}
(W_\theta)_{i,\mathbf{g}}\mathcal{X}_i,
\qquad
d_{\mathbf{g}}
=
\sum_{i=1}^{N}
(W_\theta)_{i,\mathbf{g}}.
\]
Because the learned weights may be signed, $d_{\mathbf{g}}$ can approach zero
through cancellation even when the corresponding grid point is well covered.
We therefore also accumulate the total unsigned weight
\begin{equation}
a_{\mathbf{g}}
=
\sum_{i=1}^{N}
\left|
(W_\theta)_{i,\mathbf{g}}
\right|
\end{equation}
and use the sign preserving safeguarded denominator
\begin{equation}
\widetilde{d}_{\mathbf{g}}
=
\operatorname{sgn}_{+}(d_{\mathbf{g}})
\max
\left\{
|d_{\mathbf{g}}|,
\eta a_{\mathbf{g}},
\varepsilon
\right\},
\label{eq:stable-scatter-denominator}
\end{equation}
where $\operatorname{sgn}_{+}(z)=-1$ if $z<0$ and $+1$ otherwise,
$\eta\in(0,1]$, and $\varepsilon>0$. The normalised feature is then
\[
\widetilde{\mathcal{X}}_{\mathcal{G}}[\mathbf{g}]
=
\frac{
n_{\mathbf{g}}
}{
\widetilde d_{\mathbf{g}}
}.
\]
Equivalently, defining
$D_\theta=\operatorname{diag}(\widetilde d_{\mathbf{g}})$ gives the compact
form in \Cref{eq:learnable-normalised-scatter}.

The relative floor bounds the amplification caused by cancellation,
\begin{equation}
\left\|
\widetilde{\mathcal{X}}_{\mathcal{G}}[\mathbf{g}]
\right\|_\infty
\leq
\frac{1}{\eta}
\max_{i:(W_\theta)_{i,\mathbf{g}}\neq0}
\|\mathcal{X}_i\|_\infty
\end{equation}
whenever $a_{\mathbf{g}}>0$. The absolute floor $\varepsilon$ handles
uncovered or numerically degenerate grid points. For non-negative weights,
$d_{\mathbf{g}}=a_{\mathbf{g}}$, so the relative safeguard is inactive for
$\eta\leq1$.

\paragraph{Sampling-density feature.}
Normalised scatter reduces variations caused solely by local point density,
but it also removes information about mesh coverage. We retain this
information through the fixed analytic spline basis,
\begin{equation}
\rho_{\mathbf{g}}
=
\sum_{i=1}^{N}
\Phi_{i,\mathbf{g}},
\qquad
\mathcal{X}_{\mathcal{G}}
=
P_\rho
\left[
\widetilde{\mathcal{X}}_{\mathcal{G}};
\log(1+\rho)
\right].
\label{eq:spline-density-channel}
\end{equation}
Because $\Phi$ is non-negative, $\rho$ provides a non-negative analytic
measure of local mesh coverage that is unaffected by cancellation in the
learned scatter. The pointwise projection $P_\rho$ is initialised to preserve
the scattered feature channels and assign zero weight to the density feature,
so adding $\rho$ does not change the initial model output.

\paragraph{Implementation.}
All transfers remain local. For degree $p$, each mesh point contributes to at
most $(p+1)^d$ latent-grid points, and the products in
\Cref{eq:learnable-normalised-scatter} are implemented as indexed scatter-add
operations over these active mesh-to-grid edges, as detailed in
\Cref{app:lspline-implementation}.

\subsection{Continuous spatial bounds and parameter uncertainty}
\label{app:spatial-parameter-bounds}

\paragraph{Setup.}
All bounds are computed post-training, using the checking across the 29.2M nodal points. Geometry, model weights, preprocessing,
and material assignments remain fixed. We consider three local
material regions around held-out sample, one each in steel,
copper, and graphite. No reference displacement values enter
the enclosure construction.

\paragraph{Continuous spatial bounds.}
For fixed parameters, each physical displacement component
on a decoder cell is a rational tensor-product cubic spline,
\[
\widehat u_j(x)=\frac{Y_j(x)}{D(x)}.
\]
The numerator contains the physical output coefficients,
including the affine readout, while the denominator accounts
for the normalisation of spline support truncated at the
decoder boundary. In the interior, this denominator is one.

We cover the requested material region with restricted
decoder-cell boxes. On each box, a linear basis transformation
expresses the numerator and denominator in tensor-product
Bernstein form \citep{piegl1996nurbs}:
\[
Y_j(x)=\sum_k Y_{j,k}b_k(x),
\qquad
D(x)=\sum_k D_kb_k(x).
\]
Here $k$ is a multi-index over the spatial Bernstein
coefficients. Since the basis functions are nonnegative
and sum to one, positive denominator coefficients give
\[
\widehat u_j(x)
=\sum_k w_k(x)\frac{Y_{j,k}}{D_k},
\qquad
w_k(x)=\frac{D_kb_k(x)}{\sum_\ell D_\ell B_\ell(x)},
\qquad
w_k(x)\ge0,\quad \sum_k w_k(x)=1.
\]
Thus the prediction is a convex combination of the
coefficient ratios, giving the continuous enclosure
\[
L_j=\min_k\frac{Y_{j,k}}{D_k},
\qquad
U_j=\max_k\frac{Y_{j,k}}{D_k}.
\]
These are the spatial bounds in \cref{eq:spline_bounds}.
Their validity follows from the basis representation,
rather than evaluations at sampled points.

We tighten the enclosure using de Casteljau subdivision,
which restricts the numerator and denominator together
to smaller spatial boxes. Each step splits a box at the
midpoint of its longest side, so depth $n$ covers every
original decoder cell with $2^n$ sub-boxes. The split is
an exact change of basis: each sub-box carries Bernstein
coefficients for the same rational spline, and these
approach the function values as the box shrinks.
Taking the smallest lower endpoint and largest upper
endpoint across all covering boxes gives a regional
enclosure. Each box uses the decoder head associated
with its material. \Cref{tab:app_mastu_spatial_partition_ablation}
reports the resulting trade-off: the median width
halves between depths zero and three, while the cost
roughly doubles with every further level.

\begin{table}[htbp]
\centering
\footnotesize
\setlength{\tabcolsep}{6pt}
\begin{tabular}{rrrrr}
\toprule
\textbf{Depth $n$} &
\textbf{Partitions $2^n$} &
\textbf{Median width (mm)} &
\textbf{Max width (mm)} &
\textbf{Time (s)} \\
\midrule
0 & 1 & 8.70 & 148.12 & 5.71 \\
1 & 2 & 7.65 & 142.32 & 10.64 \\
2 & 4 & 6.62 & 130.71 & 20.34 \\
3 & 8 & 4.24 & 93.26 & 39.61 \\
4 & 16 & 3.89 & 79.88 & 81.99 \\
6 & 64 & 2.71 & 57.98 & 322.43 \\
\bottomrule
\end{tabular}
\caption{MAST-U spatial subdivision ablation at nominal parameters.
Depth zero uses no subdivision within each original spline cell;
each subsequent depth doubles the number of sub-boxes per cell.
Widths are displacement-magnitude enclosure widths over the same
$4{,}077{,}312$ displayed triangles, with equal weight per triangle.
Times are medians of three warm, synchronised bound calculations on
one NVIDIA A100 GPU, excluding model loading, source encoding and
reporting. Depth three is the current subdivision setting.}
\label{tab:app_mastu_spatial_partition_ablation}
\end{table}

\paragraph{Taylor propagation through the network.}
For the prescribed parameter box, we propagate a
second-order polynomial plus a bounded interval remainder
through the unchanged network \citep{Makino1999Taylor}.
Writing $\boldsymbol{\delta}
=\boldsymbol{\mu}-\boldsymbol{\mu}_0$, an intermediate
scalar quantity is represented as
\[
z(\boldsymbol{\mu})
=z_0+\mathbf a^\top\boldsymbol{\delta}
+\tfrac12\boldsymbol{\delta}^\top H\boldsymbol{\delta}
+e,
\qquad |e|\le R.
\]
The parameter endpoints are converted using the model's
fixed normalisation before propagation. With two parameters,
the retained polynomial consists of six coefficient fields:
a constant, two linear terms, and three quadratic terms.

FiLM conditioning introduces the parameter dependence.
Affine lifting, fixed-geometry scatter, Fourier processing,
and physical readout propagate the signed polynomial
coefficient fields. Remainders are enclosed using
absolute-weight or norm bounds; Fourier remainder propagation
uses Parseval-based norm envelopes. The same parameter symbols
are shared across source points, channels, and spline
coefficients, rather than treating their perturbations
as independent intervals.

Nonlinear activations require an additional remainder.
For a scalar activation input $z_0+L+Q+e$, where $L$ and $Q$
are the retained linear and quadratic terms, let
$|L|\le s$, $|Q|\le t$, and $|e|\le R$. The retained
second-order activation polynomial is
\[
g(z_0)+g'(z_0)(L+Q)+\tfrac12 g''(z_0)L^2.
\]
A valid bound on the discarded terms is
\[
R_{\mathrm{new}}
\le
L_g R
+|g''(z_0)|\left(st+\tfrac12t^2\right)
+\tfrac16 M_3(z_0,s+t)(s+t)^3,
\]
where $L_g$ bounds $|g'|$ and $M_3(z_0,v)$ bounds $|g'''|$
on $[z_0-v,z_0+v]$. This accounts both for the incoming
remainder and for higher-order products discarded during
composition. For exact GELU, $g(z)=z\Phi(z)$,
\[
g'''(z)=(z^3-4z)\phi(z),
\]
so the third-derivative envelope is obtained analytically
from interval endpoints and contained stationary points.
The implementation also evaluates a direct composition
remainder bound and takes the smaller of the two valid
envelopes.

\paragraph{Combining parameter and spatial bounds.}
The joint bound reuses the spatial construction above
unchanged; only the coefficients it is applied to differ.
After physical readout, every spline coefficient has a
quadratic parameter polynomial and a bounded remainder.
We transform the signed polynomial coefficient fields
into spatial Bernstein form before computing their
parameter extrema. This preserves cancellations that
would be lost by first replacing each spline coefficient
with an independent interval.

On each spatial patch, write
\[
Y_{j,k}(\boldsymbol{\mu})
=Y_{j,k}^0+\pi_{j,k}(\boldsymbol{\delta})+\eta_{j,k},
\qquad
|\eta_{j,k}|\le r_{j,k},
\]
where $\pi_{j,k}$ contains the linear and quadratic parameter
terms. We obtain its extrema $\pi_{j,k}^{\pm}$ over the
parameter-offset box by checking corners, admissible
stationary points on its edges, and any admissible interior
stationary point. For a singular quadratic Hessian,
boundary candidates suffice; degenerate parameter boxes
are handled in their remaining dimensions.

The lower and upper numerator-coefficient bounds are
\[
Y_{j,k}^-=Y_{j,k}^0+\pi_{j,k}^- -r_{j,k},
\qquad
Y_{j,k}^+=Y_{j,k}^0+\pi_{j,k}^+ +r_{j,k}.
\]
Since the denominator is parameter-independent and its
Bernstein coefficients are positive, the spatial
convex-hull argument gives
\[
\min_k\frac{Y_{j,k}^-}{D_k}
\le \widehat u_j(x,\boldsymbol{\mu})
\le
\max_k\frac{Y_{j,k}^+}{D_k}
\]
for every position and parameter setting in the covered
product domain.

\paragraph{Refinement and magnitude ranges.}
Spatial subdivision tightens the Bernstein enclosure.
Parameter subdivision reduces the local nonlinear
remainders: we partition the complete original parameter
box, recenter the Taylor model on every leaf, and take
the union of all leaf enclosures. Every leaf must be
included to retain coverage of the original uncertainty
range.

For component intervals $[L_j,U_j]$, a displacement-magnitude
enclosure follows from the nearest and furthest points
of the component rectangle from the origin:
\[
m^-=
\left(\sum_j \operatorname{dist}
       (0,[L_j,U_j])^2\right)^{1/2},
\qquad
m^+=
\left(\sum_j\max\{L_j^2,U_j^2\}\right)^{1/2}.
\]
Magnitude enclosures are computed before combining parameter
leaves, avoiding the unnecessary combination of component
extrema from different leaves.

At spatial subdivision depth three, with physical
parameter half-widths $10^{-7}~\mathrm{K}^{-1}$ in $\alpha$
and $5$~K in $T_w$, the reported second-order results are

\[
\begin{array}{lcc}
\hline
\text{Material region}
& \text{Nominal spatial range (mm)}
& \text{Joint range, }8\times8\text{ partition (mm)}\\
\hline
\text{Steel}    & [41.47,42.37] & [41.06,42.80]\\
\text{Copper}   & [39.86,41.79] & [39.47,42.22]\\
\text{Graphite} & [35.62,38.80] & [35.26,39.21]\\
\hline
\end{array}
\]
All parameter leaves cover the same original uncertainty
box; these results do not arise from shrinking that box.

\paragraph{Interval arithmetic and rounding certificates.}
We compare arithmetic implementations on the held-out case
as shown in
\Cref{fig:mastu-displacement,fig:mastu-displacement-error}, using the
same frozen checkpoint and the $1\times$ parameter box with physical
half-widths $10^{-7}~\mathrm{K}^{-1}$ in $\alpha$ and $5$~K in $T_w$.
Every evaluation encodes all $29{,}183{,}754$ native source nodes and
computes displacement-magnitude intervals at the same $4{,}077{,}312$
display face centres. These are parameter bounds at fixed positions;
they differ from the continuous spatial-subdivision experiment in
\Cref{tab:app_mastu_spatial_partition_ablation} and the partitioned
local-region ranges above. No parameter subdivision is used here.

\begin{figure}[H]
\centering
\includegraphics[width=\linewidth]{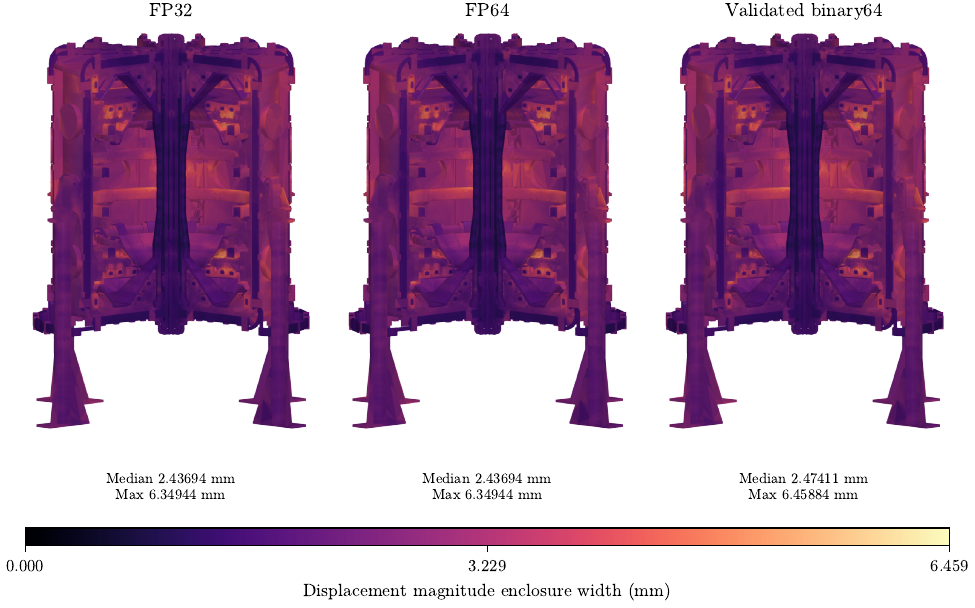}
\caption{Displacement-magnitude enclosure widths at fixed face centres
for \texttt{eval\_02} over the $1\times$ parameter box, using ordinary
FP32, ordinary FP64 and validated binary64. The FP32/FP64 calculations
use verified norms and quadratic extrema with ordinary propagation;
validated binary64 encloses the real stored-weight surrogate.
All panels use the manuscript camera, the same geometry and a shared
linear colour scale without clipping. The distributions correspond to
\Cref{tab:app_mastu_parameter_arithmetic}; panel annotations give the
median and maximum widths over all $4{,}077{,}312$ face centres.}
\label{fig:app_mastu_parameter_arithmetic_surfaces}
\end{figure}

The original formulation implements the real-arithmetic Taylor and
spline inequalities using ordinary FP32 scatter, FFT, nonlinear and
SVD calculations. The updated ordinary FP32/FP64 formulation retains
the same second-order construction but replaces numerical norm
estimates by verified upper bounds: SVD proposals are checked using
outward-enclosed Hermitian LDL pivots, with conservative norm
fallbacks. Quadratic parameter extrema are evaluated with outward
binary64 intervals, including stationary values and conservative
fallbacks when the Hessian determinant cannot be resolved safely.
These checks validate those individual calculations; the remaining
ordinary propagation is not an end-to-end rounding certificate.

Validated binary64 instead propagates outward-enclosed arithmetic
through parameter normalisation, coordinate/positional preprocessing,
FiLM, encoder accumulation, Fourier/Taylor propagation, nonlinear
remainders and the surface decoder. Arithmetic uncertainty is
included alongside the Taylor remainder. This encloses the real
surrogate defined by the exact stored weight values under the checked
arithmetic assumptions. It does not certify every output of the
existing vendor FP32/TF32 inference program.

The validated implementation specializes interval products by exact stored
weights and fuses spline gathers and transcendental series recurrences
into CUDA kernels. These kernels retain explicit directed binary64
rounding, the original accumulation order and analytic remainder bounds.
For the measured box, the saved magnitude endpoints are identical to
those from the earlier validated implementation at every face centre.

\Cref{tab:app_mastu_parameter_arithmetic} reports medians of three
CUDA-synchronised repetitions in one NVIDIA A100-SXM4-80GB allocation. The original and
updated implementations ran in separate allocations of the same GPU
model, with matched geometry, parameter boxes, chunk sizes and thread
counts. Timed calls include source/query preprocessing and transfers,
FiLM conditioning, encoding, FNO propagation and surface queries.
They exclude model restoration, mesh loading, constant preparation,
warmup, statistics, archive writes and plotting. Width statistics
use the first measured repetition, with equal weight per face centre.
Every call rebuilds its Taylor state and output bounds; only loaded
fixed geometry and weight/grid-dependent constants are reused.
Source features and spline stencils are recomputed.

\begin{table}[htbp]
\centering
\footnotesize
\setlength{\tabcolsep}{5pt}
\begin{tabular}{lrrr}
\toprule
\textbf{Interval implementation} & \textbf{Time (s)} &
\textbf{Median width (mm)} & \textbf{Max width (mm)} \\
\midrule
Original formulation, FP32 & 74.63 & 2.43694 & 6.34944 \\
Verified norms/extrema, FP32 & 106.68 & 2.43694 & 6.34944 \\
Verified norms/extrema, FP64 & 133.83 & 2.43694 & 6.34944 \\
Validated binary64 & 356.40 & 2.47411 & 6.45884 \\
\bottomrule
\end{tabular}
\caption{Parameter-box arithmetic comparison for the evaluation case at $1\times$, using the optimized validated binary64 implementation.
Times are median per-box costs over three repetitions in the same A100
allocation with cached constants, excluding setup and warmup. Widths are
displacement-magnitude enclosure widths at fixed face centres.
The ordinary FP32/FP64 rows have no end-to-end rounding certificate;
validated binary64 encloses the real surrogate defined by the stored
weights.}
\label{tab:app_mastu_parameter_arithmetic}
\end{table}

The original and updated FP32 widths agree at the precision of
\Cref{tab:app_mastu_parameter_arithmetic}: their largest pointwise
width difference is $7.97\times10^{-6}$~mm and their largest endpoint
difference is $3.41\times10^{-5}$~mm. Thus this case shows negligible
change in enclosure width from verifying norms and extrema. The
updated FP32 surface-query stage costs about $40.07$~s versus
$8.45$~s for the original, consistent with its more expensive
enclosed extremum calculation. Validated binary64 increases the
median width by approximately $1.53\%$ and the maximum by $1.72\%$.
It also encloses real preprocessing, whereas ordinary FP64 promotes
the captured FP32 processed features, so this increase is not solely
a final rounding pad. Agreement of computed widths and sampled
containment checks are empirical diagnostics; they do not establish
the enclosure guarantee of ordinary arithmetic. Caching reduces
repeated constant-construction cost in both formulations and is
independent of rounding validation.

\paragraph{Effect of parameter-box size in FP32.}

\begin{figure}[t]
\centering
\includegraphics[width=\linewidth]{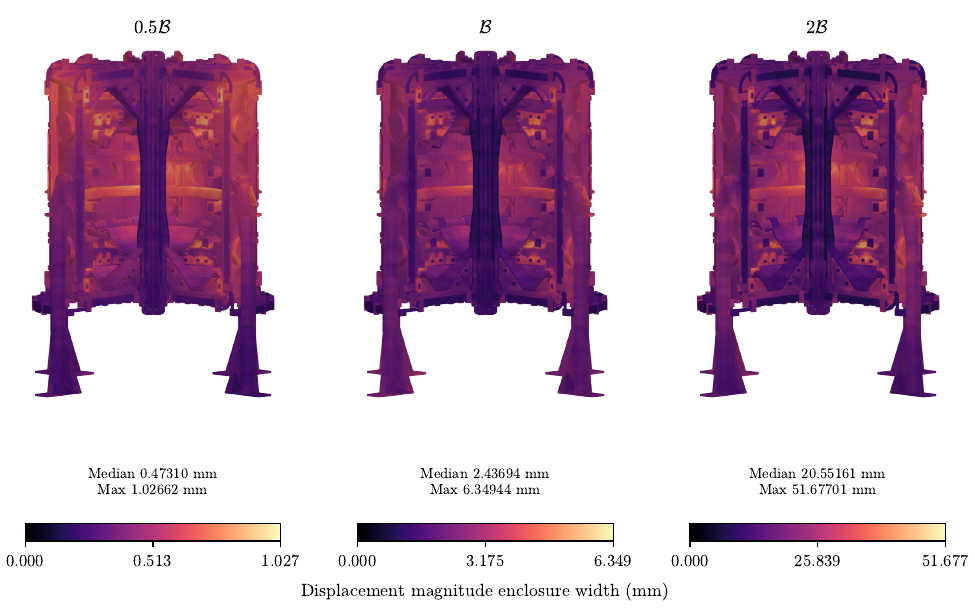}
\caption{FP32 displacement-magnitude enclosure widths for
\texttt{eval\_02} over the $0.5\times$, $1\times$ and $2\times$
parameter boxes in \Cref{tab:app_mastu_fp32_box_widths}. Both parameter
half-widths scale together about the same nominal parameters, with
no parameter subdivision. The geometry and camera are identical to
\Cref{fig:app_mastu_parameter_arithmetic_surfaces}. A shared linear
colour scale spanning the largest width in these three panels shows
the growth of field variation and enclosure conservatism without
clipping or independent panel rescaling. Widths are evaluated at
fixed face centres using the updated ordinary FP32 formulation.}
\label{fig:app_mastu_parameter_box_size_surfaces}
\end{figure}

\Cref{tab:app_mastu_fp32_box_widths} varies both physical parameter
half-widths together for the same evaluation case, using the
updated formulation with verified norms/extrema and ordinary FP32
propagation. The checkpoint, native source nodes and face centres
remain fixed. Width statistics again use the first measured
repetition, with no parameter subdivision. Larger boxes increase
both the allowed field variation and enclosure conservatism.

\begin{table}[htbp]
\centering
\footnotesize
\setlength{\tabcolsep}{5pt}
\begin{tabular}{rrrrr}
\toprule
\textbf{Box factor} &
\makecell{\textbf{$\alpha$ half-width}\\$(\mathrm{K}^{-1})$} &
\makecell{\textbf{$T_w$ half-width}\\\textbf{(K)}} &
\makecell{\textbf{Median width}\\\textbf{(mm)}} &
\makecell{\textbf{Max width}\\\textbf{(mm)}} \\
\midrule
$0.5\times$ & $5\times10^{-8}$ & 2.5 & 0.47310 & 1.02662 \\
$1\times$ & $10^{-7}$ & 5 & 2.43694 & 6.34944 \\
$2\times$ & $2\times10^{-7}$ & 10 & 20.55161 & 51.67701 \\
\bottomrule
\end{tabular}
\caption{FP32 displacement-magnitude enclosure widths for
\texttt{eval\_02} across three parameter-box sizes. Both half-widths
scale by the box factor about the same nominal parameters. Median
and maximum widths are computed over all $4{,}077{,}312$ fixed face
centres. These are ordinary-arithmetic enclosure calculations,
without an end-to-end rounding certificate.}
\label{tab:app_mastu_fp32_box_widths}
\end{table}

\paragraph{Setup, warmup and constant reuse.}
Model restoration and geometry loading cost $10.69$~s for the
original run and $12.25$~s shared across the three updated methods.
\Cref{tab:app_mastu_parameter_setup} gives the additional constant
preparation/check and warmup costs. The original formulation builds
ordinary spectral constants in $2.64$~s. The updated run first builds
verified Fourier and skip norm constants in FP32 mode ($336.39$~s);
FP64 and validated binary64 reuse the same exact promoted weight/grid
keys, with cache checks of $0.46$~s and $0.41$~s, respectively.
The aggregate verified-constant preparation/check cost is $337.25$~s.
An independent cold start of an updated method must build the verified
constants; the later modes' lookup times do not represent cold-start
setup costs. Each method receives one full-workload warmup on
\texttt{eval\_02} at $0.5\times$ before any timed repetitions.

\begin{table}[htbp]
\centering
\footnotesize
\setlength{\tabcolsep}{6pt}
\begin{tabular}{lrr}
\toprule
\textbf{Interval implementation} &
\textbf{Constant setup/check (s)} & \textbf{Warmup (s)} \\
\midrule
Original formulation, FP32 & 2.64 & 88.75 \\
Verified norms/extrema, FP32 & 336.39 & 108.64 \\
Verified norms/extrema, FP64 & 0.46 & 134.67 \\
Validated binary64 & 0.41 & 522.53 \\
\bottomrule
\end{tabular}
\caption{One-time constant preparation/checks and one full warmup per
method, excluding the separately reported restore/geometry cost.
The updated methods ran in order FP32, FP64, validated binary64 in
one process: FP32 builds the shared verified constants, while the
last two rows report cache checks. Warmup uses the $0.5\times$ box;
the measured comparison in \Cref{tab:app_mastu_parameter_arithmetic}
uses $1\times$.}
\label{tab:app_mastu_parameter_setup}
\end{table}

These bounds enclose the surrogate prediction. They do
not quantify prediction error relative to a reference
solution or bound PDE discretisation, solver convergence,
or physical modelling error.

\newpage
\section{L-SplineFNO Implementation}
\label{app:lspline-implementation}

L-SplineFNO scatters point features onto a regular grid with learned spline
weights, processes the grid with an FNO, and gathers the result at query
points with the fixed analytic spline. The listings describe these cubic
transfers in PyTorch-style pseudocode using ordinary autograd. We describe
the single-scale, single-region
path with an affine readout, followed by the extensions and memory optimisations.

\paragraph{Notation.}
We use $B_{\mathrm{batch}}$ samples per batch, $C$ feature channels (the width $c$ of the main
text), $N$ input points and $N_q$ output query points in $d$ spatial
dimensions. The regular grid has $S_\ell$ nodes along axis $\ell$ and
$G=\prod_{\ell=1}^{d}S_\ell$ nodes in total. Each point interacts with at
most $K=4^d$ grid nodes. The listings abbreviate $B_{\mathrm{batch}}$ as \texttt{B};
all tensors and kernels must have compatible devices and floating-point dtypes.

\paragraph{Step 1: neighbouring grid nodes and their weights.}
The \emph{stencil} of a point is the set of grid nodes it contributes to.
Along axis $\ell$, with grid origin $\xi_0^{(\ell)}$ and spacing
$\Delta_\ell$, the coordinate
$t_i^{(\ell)}=(\omega_i^{(\ell)}-\xi_0^{(\ell)})/\Delta_\ell$ gives the
position of point $i$ in grid cells. Its four candidate neighbours are
$g_\ell=\lfloor t_i^{(\ell)}\rfloor+o$ with $o\in\{-1,0,1,2\}$, at
distances $t_i^{(\ell)}-g_\ell\in[-2,2)$. Each neighbour is weighted by the
kernel evaluated at that distance. Neighbours that fall outside the grid
receive zero weight, and the remaining weights are normalised along each axis
before being multiplied across axes:
\begin{equation}
v_{i,g_\ell}^{(\ell)}=
\frac{\mathbf{1}_{0\leq g_\ell<S_\ell}\,
   \kappa_\theta^{(\ell)}(t_i^{(\ell)}-g_\ell)}
{\operatorname{sgn}_{+}(s_{i\ell})\max(|s_{i\ell}|,\delta)},
\qquad
(W_\theta)_{i,\mathbf{g}}=\prod_{\ell=1}^{d}v_{i,g_\ell}^{(\ell)},
\label{eq:implementation-axis-weights}
\end{equation}
where $s_{i\ell}$ is the sum of the valid kernel weights of point $i$ along
axis $\ell$, $\operatorname{sgn}_{+}(s)=-1$ for $s<0$ and $+1$ otherwise,
and $\delta$ prevents division by a near-zero sum. Whenever $\delta$ is
inactive, each axis's weights sum to one, and hence so does each row of
$W_\theta$. We use $\delta=10^{-3}$ for the learned kernel. The analytic
kernel uses $\delta=10^{-12}$; this floor is inactive for points inside
the grid bounds.

The learned kernel follows \Cref{eq:learnable-spline-kernel}, with $J=9$
correction centres uniformly spaced on $[-1,1]$ and width $h=1/2$ by default.
The coefficients $c=V_0\alpha$ preserve the interior partition of unity to
numerical precision. A float64 SVD of the $257\times9$ collocation matrix
in \Cref{app:learnable-spline-construction} gives six free parameters per
axis; $V_0$ is cached per kernel configuration. Initialising $\alpha=0$
recovers the analytic cubic kernel, not the least-squares encoder of SplineFNO.
The implementation evaluates the correction using a piecewise-cubic table
and Horner's rule. Listing~\ref{lst:lspline-stencil} stores the four neighbours
per axis separately; their tensor product is formed only for the point chunk
being processed.

\begin{lstlisting}[style=lsplinepython,caption={Build a stencil: the four neighbouring grid nodes and their normalised weights along each axis, stored per axis.},label={lst:lspline-stencil}]
import torch
from torch import nn

# Assumed helpers: C2SplineKernel1D is the learned kernel;
# cubic_bspline is the analytic kernel; _scatter_axis builds
# the per-axis weights; _FactoredStencil stores their factors.

# One trainable kernel per axis; create these once in the model.
d = 3  # use 2 for planar grids
kernels = nn.ModuleList([C2SplineKernel1D() for _ in range(d)])

def stencil(coords, axes, kernels, floor=1e-3):
    # coords: [B, d, N]; axes[ax]: grid-node coordinates along axis ax.
    # Returns, per axis, [B, N, 4] neighbour indices and weights.
    pairs = [_scatter_axis(coords[:, ax], axes[ax], kernels[ax],
                           pou_floor=floor) for ax in range(len(axes))]
    indices, weights = zip(*pairs)
    return _FactoredStencil(indices, weights, [len(a) for a in axes])

# Learned weights at the source points; fixed analytic weights at the queries.
# src = stencil(source_coords, axes, kernels)
# qry = stencil(query_coords, axes, [cubic_bspline] * d, floor=1e-12)
\end{lstlisting}

\paragraph{Step 2: scatter to the grid and gather back.}
\emph{Scatter} adds each input point's features to its stencil nodes,
weighted by $W_\theta$. For features $H\in\mathbb{R}^{B_{\mathrm{batch}}\times N\times C}$,
each grid node $\mathbf g$ accumulates three sums (batch index omitted):
\begin{equation}
n_{\mathbf g}=\sum_i (W_\theta)_{i,\mathbf g}H_i,
\qquad d_{\mathbf g}=\sum_i (W_\theta)_{i,\mathbf g},
\qquad a_{\mathbf g}=\sum_i |(W_\theta)_{i,\mathbf g}|,
\end{equation}
the weighted feature sum, the signed weight sum and the absolute weight sum.
Without safeguards, the normalised value $n_{\mathbf g}/d_{\mathbf g}$
reproduces constant features wherever $d_{\mathbf g}\neq0$. Signed weights
can, however, cancel. We therefore use \Cref{eq:stable-scatter-denominator}
with $\eta=0.1$ and $\varepsilon=10^{-6}$. This bounds each output channel
by $1/\eta$ times the largest absolute input value contributing to that node.
For non-negative weights, $a_{\mathbf g}=d_{\mathbf g}$, so only the
absolute floor can be active. Uncovered nodes remain zero; constant features
are reproduced wherever neither floor changes the denominator.

After the FNO produces grid features $Z\in\mathbb{R}^{B_{\mathrm{batch}}\times G\times C}$,
\emph{gather} evaluates them at each query point $q_{b,j}$ as a weighted sum
of that point's stencil nodes, using the analytic weights $\Phi$:
\begin{equation}
\widehat H_{b,j}=\sum_{\mathbf g}
   \Phi(q_{b,j})_{\mathbf g}\, Z_{b,\mathbf g}.
\end{equation}
Scatter and gather build separate stencils, so the source and query clouds
may differ in both location and size. The output projection then maps
$\widehat H$ to the physical output variables.

Listing~\ref{lst:lspline-transfer} processes \texttt{chunk} points at a time.
The method \texttt{taps} returns flattened grid indices and weights of shape
$[B_{\mathrm{batch}},\texttt{hi}-\texttt{lo},K]$, or batch dimension one for a
shared mesh. The first grid axis varies slowest. Adding $bG$ to sample $b$'s
indices lets indexed additions and gathers process the batch together.
Gradients propagate to features, grid values and learned kernel coefficients.

\begin{lstlisting}[style=lsplinepython,caption={Scatter features from points to the grid with guarded normalisation, and gather grid features at query points. \texttt{chunk} is the number of points processed at once.},label={lst:lspline-transfer}]
def scatter(H, src, G, chunk, eta=0.1, eps=1e-6):
    # H: [B, N, C] point features -> [B, G, C] grid features.
    B, N, C = H.shape
    offset = (torch.arange(B, device=H.device) * G)[:, None, None]
    num = H.new_zeros(B * G, C)    # n_g
    mass = H.new_zeros(B * G, 2)   # (d_g, a_g)
    for lo in range(0, N, chunk):
        hi = min(lo + chunk, N)
        col, w = src.taps(node_start=lo, node_stop=hi)
        w = w.expand(B, -1, -1)  # a shared-mesh stencil has batch 1
        idx = (col + offset).reshape(-1)
        values = H[:, lo:hi, None, :] * w[..., None]
        num.index_add_(0, idx, values.reshape(-1, C))
        mass.index_add_(0, idx,
            torch.stack((w, w.abs()), -1).reshape(-1, 2))
    # Normalise only after every point has contributed.
    den, absolute = mass.unbind(-1)
    sign = torch.where(den < 0, -1., 1.)
    safe = sign * torch.maximum(den.abs(), (eta * absolute).clamp_min(eps))
    return (num / safe[:, None]).view(B, G, C)

def gather(Z, qry, chunk):
    # Z: [B, G, C] grid features -> [B, N_q, C] features at the queries.
    B, G, C = Z.shape
    offset = (torch.arange(B, device=Z.device) * G)[:, None, None]
    flat = Z.reshape(B * G, C)
    out = Z.new_empty(B, qry.n_points, C)
    for lo in range(0, qry.n_points, chunk):
        hi = min(lo + chunk, qry.n_points)
        col, w = qry.taps(node_start=lo, node_stop=hi)
        idx = (col + offset).reshape(-1)
        values = flat.index_select(0, idx).view(B, hi - lo, -1, C)
        out[:, lo:hi] = (values * w[..., None]).sum(2)
    return out
\end{lstlisting}

\paragraph{Step 3: the full model.}
The pointwise lift forms $H_i=P_{\mathrm{in}}[x_i;\bar\omega_i]$, as in
\Cref{eq:pipeline-lift}. For the FiLM-conditioned benchmarks,
$x_i=\boldsymbol\gamma_\mu\odot\varphi(\bar\omega_i)+\boldsymbol\beta_\mu$
contains 96 sinusoidal features ($d=3$, $n_f=16$), modulated in the outer
wrapper. The three normalised coordinates bypass modulation, so
$P_{\mathrm{in}}:\mathbb{R}^{99}\rightarrow\mathbb{R}^{C}$.

Normalisation removes the overall weight magnitude at each node. We retain
coverage information through $\rho_{\mathbf g}=\sum_i\Phi(\bar\omega_i)_{\mathbf g}$,
computed on the fine grid from the source coordinates. This non-negative
feature depends on the sampled geometry but not on the learned kernels.
The channel $\log(1+\rho)$ is appended to the fused features and mixed back to
$C$ channels by a $1\times1$ convolution $P_\rho$, initialised as the identity
on the features and zero on the density, so that the density has no effect at
initialisation. The full forward pass is
\begin{equation}
\begin{aligned}
X_s &= D_{\theta,s}^{-1}W_{\theta,s}^{\top}H,
& X_{\mathrm{fine}} &= \operatorname{Fuse}\!\big(\{\operatorname{Up}_s X_s\}_{s\in\mathcal{S}}\big),\\
Z &= \mathcal F_\theta\!\left(
   P_\rho[X_{\mathrm{fine}};\log(1+\rho)]\right),
& \widehat{\mathcal Y}(q) &= P_{\mathrm{out}}\!\left(\Phi(q)Z\right),
\end{aligned}
\label{eq:implementation-full-forward}
\end{equation}
where $D_{\theta,s}$ contains the guarded denominators of Step~2,
$\mathcal{S}$ is the set of scales, $\operatorname{Up}_s$ interpolates scale $s$ to the fine grid
($\operatorname{Up}_1$ is the identity) and $\operatorname{Fuse}$ is the
$1\times1$ fusion; both are omitted with a single scale. When spectral
modulation is selected, a condition embedding is also passed to every FNO
layer. Region-specific variants may use separate lifts, output projections or
kernel banks. With separate kernels, the three scatter sums are added across
regions before division; constant reproduction has the same coverage and
safeguard conditions as in Step~2. Region routing uses node labels supplied
alongside the coordinates.

The two halves are exposed separately. \texttt{encode} runs the lift, scatter
and FNO and returns the grid state $Z$, and \texttt{decode} gathers $Z$ at a
query cloud and applies $P_{\mathrm{out}}$. Evaluating at new query points
therefore needs only a new \texttt{decode} in the path described here.
Region-specific readouts additionally require query-region labels, and the
optional node skip connection requires lifted features at the query nodes.

\paragraph{Cost and memory.}
At fixed width and stencil size, both transfers are linear in the number of
points: writing $B=B_{\mathrm{batch}}$, scatter costs $O(BNKC)$ and gather
$O(BN_qKC)$, plus $O(BGC)$ to
initialise the grid buffers. The encoder needs no $G\times G$ Gram matrix and
no factorisation. Each FNO layer costs $O(BCG\log G)$ for its real FFTs, and
dense spectral channel mixing costs $O(BC^2M_f)$ for $M_f$ retained modes,
with a further $O(BC^2G)$ for pointwise channel mixing.
The main transfer optimisations are:
\begin{itemize}[leftmargin=*,itemsep=2pt,topsep=3pt]
\item \textbf{Factored stencils.}
Storing four index--weight pairs per axis takes $4dBN$ pairs, against
$4^dBN$ for the full tensor product ($5.3\times$ fewer for $d=3$). The product
is formed only for the chunk being processed.
\item \textbf{Fused accumulation.}
On both CPU and GPU, each point chunk uses one indexed addition for features
and one for the signed and absolute weight sums together:
$2\lceil N/M\rceil$ additions for chunks of $M$ points per scale and region.
\item \textbf{Recomputed backward passes.}
Custom autograd functions save the per-axis stencil factors and the source
features or grid state. They reconstruct tensor-product indices and weights
in chunks during backward, avoiding retention of every chunk's expanded
intermediates as ordinary autograd would do for the reference listings.
\item \textbf{Shared geometry.}
When every sample in a batch has the same coordinates and these do not
require gradients, one stencil is built and broadcast across the batch, and
the weight sums $d_{\mathbf g}$ and $a_{\mathbf g}$, which depend only on the
geometry, are accumulated once. Samples with different meshes keep their own
stencils.
\item \textbf{Matrix-free decoding.}
The gather computes $\Phi Z$ directly, so the sparse reconstruction matrix is never
assembled in the forward path.
\end{itemize}
The parameter \texttt{scatter\_max\_temp\_mb} (default 128 MiB) sets an
approximate temporary-buffer budget for both transfers on every device.
It controls chunk size, not total model or training memory. Gather reduces each query's $K$ contributions within a
single chunk; scatter reductions may differ in floating-point rounding
with chunk size or device. For source clouds too large to lift in one pass,
streamed single-sample inference reads and lifts the inputs in chunks,
accumulates the scatter and density sums over all chunks before normalisation,
and runs the FNO once. This matches the complete-cloud computation up to
floating-point rounding.

\newpage
\section{Physics and Data}
\label{app:physics_data}

All six benchmarks are steady-state, conditioned operator-learning problems
posed on unstructured meshes. They were chosen to span two axes that matter for
a conditional neural operator. The first is \emph{geometric}: shared-mesh
problems, where a single node ordering is valid for every case (Aircraft, HIVE,
Heated duct), against per-sample-mesh problems, where node index $i$ refers to
unrelated physical locations in different cases (Heat sink, Bracket, Building fire). The second
is the \emph{informativeness of $\boldsymbol{\mu}$}: how much of what
$\boldsymbol{\mu}$ says is already recoverable from the per-node input.

\begin{itemize}
\item \textbf{Redundant.} Heat sink --- every parameter describes a
shape the model already sees, so any gain from conditioning measures how hard
that shape information is to extract from a sampled point cloud rather than
how much new information $\boldsymbol{\mu}$ adds.
\item \textbf{Partly non-redundant.} Bracket, Building fire --- on the bracket the three geometry parameters restate the mesh, but the applied pressure load is
invisible in it, so two designs can share a shape and differ only in
$\boldsymbol{\mu}$. The fire splits the same way: the refinement box that
follows the burner places the two fire coordinates in the point cloud exactly,
while the five ventilation parameters move no node at all. These are the only
cases here where $\boldsymbol{\mu}$ splits into an inferable half and a hidden
one, and both loaders expose each half separately so the two can be ablated apart. 
\item \textbf{Non-redundant.} Aircraft, HIVE, Heated duct --- the five flight conditions of one airframe have identical coordinates, so $\boldsymbol{\mu}$ is the only
signal separating Mach 2 from Mach 7. HIVE and the heated duct are the same
in kind: one shared mesh across every run, so nothing in the geometry
distinguishes them. 
\end{itemize}

A third property is carried by two cases. The heated duct is
\emph{multi-region}: its domain is three separate meshes that share no cells and
couple only across two conjugate interfaces, so temperature is continuous there
while its gradient is not. All other benchmarks here are single structures, and the
gridded operators reduce to a single latent grid without loss; on the heated duct and building fire the architecture deploys three different branches for mesh encoding and decoding with a shared latent backbone.

\begin{table*}[t]
\centering
\small
\renewcommand{\arraystretch}{1.08}
\resizebox{\textwidth}{!}{%
\begin{tabular}{@{}llrrrrl@{}}
\toprule
\textbf{Dataset}
& \textbf{Physics / target}
& \makecell{\textbf{Cases}\\\textbf{train/test}}
& \makecell{\textbf{Native}\\\textbf{nodes}}
& \makecell{\textbf{Used}\\\textbf{nodes}}
& $\dim\boldsymbol{\mu}$
& \textbf{Mesh structure} \\
\midrule
Heat sink
& Conduction / $T$
& $625$ ($438/187$)
& $4{,}671$--$32{,}645$
& $16{,}384$
& $4$
& Sample-dependent \\

Bracket
& Elasto-plasticity / $\sigma$
& $3{,}000$ ($2{,}100/900$)
& $5{,}054$--$15{,}210$
& $8{,}192$
& $4$
& Sample-dependent \\

Aircraft
& Compressible flow / $(\rho,u,v,w,p)$
& $150$ ($105/45$)
& $331{,}971$
& $131{,}072$
& $3$
& Shared correspondence \\

Building fire
& Fire LES (time-avg.) / $(\Delta T,\rho_{\rm air},\rho_{\rm prod})$
& $547$ ($383/164$)
& $116{,}928$
& $98{,}496$
& $7$
& Sample-dependent, three-storey \\

Heated duct
& Conjugate heat transfer / $T$
& $122$ ($103/19$)
& $124{,}800$
& $124{,}800$
& $5$
& Shared, three-region \\
HIVE
& Induction heating / $T$
& $186$ ($149/37$)
& $194{,}691$
& $194{,}691$
& $6$
& Shared \\

\bottomrule
\end{tabular}%
}
\caption{Overview of the six benchmarks. ``Native nodes'' denotes the original
mesh size, whereas ``used nodes'' is the spatial budget presented to the model.
Variable-size meshes are sampled or duplicate-padded to this budget.}
\label{tab:data_overview}
\end{table*}

\subsection{Conventions common to all datasets}
\label{app:data_conventions}

Each data loader returns a single tensor
$\mathbf{F}\in\mathbb{R}^{N\times(c_{\mathrm{in}}+c_{\mathrm{out}})\times
N}$ together with $\boldsymbol{\mu}\in\mathbb{R}^{N\times k}$ and
$\Delta t = 0$, the latter flagging the steady-state path. The first
$c_{\mathrm{in}}$ channels of $\mathbf{F}$ are inputs and the remainder are
targets.

Where the node count varies case to case (Heat sink, Bracket), every case is
drawn down to one fixed node budget $N$ so the cases stack into
a dense tensor. The budget is a free parameter of the sweep rather than a
property of the corpus; we set it to the power of two nearest the median mesh, so
a majority of cases are drawn without replacement and the remainder are
duplicate-padded with their own real nodes. Padding with repeats rather than with
zeros or a sentinel matters for the gridded backbones: a repeated node adds
weight to a knot that a real node already occupies, which the scatter and its
density normalisation absorb exactly, whereas an artificial coordinate would
invent geometry. The cost is a reduced effective sample --- at the budgets used
here, 432 of the 625 heat-sink cases and 878 of the 3{,}000 bracket designs are
padded, leaving on average 76\% and 97\% of their slots occupied by distinct
nodes (\Cref{tab:data_overview}). The per-sample geometry path then reads
each case's own coordinates out of its input channels rather than broadcasting a
cross-case mean mesh, which for independently drawn meshes would collapse towards
the domain centroid. The three shared-mesh cases need none of
this: Aircraft draws all 150 cases at the same node indices, and HIVE and the
heated duct use every node of their one mesh.

All six benchmarks are \emph{geometry-only}: no case carries a field input
channel, so the per-node input is the processed coordinates and nothing else.
Where node positions differ between cases, the three leading channels of
$\mathbf{F}$ hold that case's own $(x, y, z)$ and $c_{\mathrm{in}} = 3$. This
happens in two ways: either each case has its own mesh (Heat sink, Bracket,
Building fire), or every case is a deformation of one shared template, so
node $k$ is the same template node in every case, but its position changes with
the airframe (Aircraft). The per-sample geometry path consumes these channels
as geometry rather than as features, so they are not passed to the backbone as
channels. Where the mesh is genuinely fixed (HIVE, Heated duct), the
coordinates live in the mesh metadata instead and $c_{\mathrm{in}} = 0$. Either
way, the coordinates reach the backbone only through the coordinate processor,
where an integrated sinusoidal positional encoding at 16 frequencies gives them
bandwidth, mapping
$\mathbb{R}^3 \to \mathbb{R}^{3 + 2\cdot 3\cdot 16} = \mathbb{R}^{99}$. The
network therefore consumes
$\mathcal{X}\in\mathbb{R}^{B_{\mathrm{batch}}\times N\times 99}$ and emits
$\mathbb{R}^{B_{\mathrm{batch}}\times N\times c_{\mathrm{out}}}$ throughout.
The first 3 of these 99 channels are the normalised coordinates and the other
96 are the sinusoidal features $\varphi(\bar\omega)$. Under the field-space
conditioning methods (FiLM, AdaLN-Zero, cross-attention), $\boldsymbol{\mu}$
modulates only the 96 sinusoidal channels, and the 3 coordinates pass through
unmodulated (\Cref{eq:pipeline-lift}).

\begin{figure}[tbp]
\centering
\includegraphics[width=0.65\textwidth]{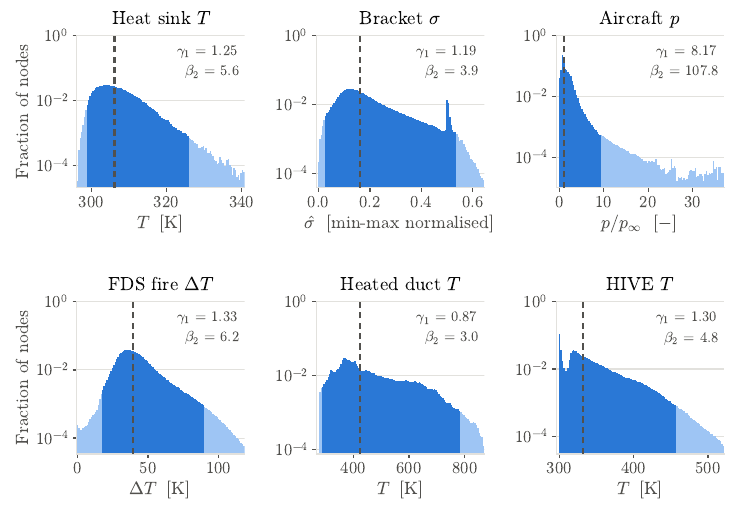}
\caption{Distribution of the target field over all nodes of all cases, one panel
per benchmark (pressure shown for aircraft and temperature change for building fire). Counts are normalised to a fraction of nodes and plotted on a log ordinate; the horizontal axis is
trimmed to $[p_{0.05}, p_{99.95}]$ and the lighter shading marks the tails
outside $[p_1, p_{99}]$. The dashed line is the median. Sample skewness
$(\gamma_1)$ and kurtosis $(\beta_2)$ are annotated (a Gaussian has
$\mathrm{Skew} = 0$, $\mathrm{Kurt} = 3$).}
\label{fig:app_target_dists}
\end{figure}

Field normalisation is per-variable and standardised throughout; conditioning
parameters are min--max normalised per variable.
Train/test splits are random permutations under a fixed seed of 42, except on the
heated duct, whose designed split is honoured instead (\Cref{app:duct}).
The field normaliser, the parameter normaliser and the coordinate processor are all
fitted on the pooled corpus before the train--test split. The Building fire departs 
from these conventions in one respect: a cell-size-aware positional encoding,
described in \Cref{app:fds}.

Reported errors are the percentage relative $L_2$ error, computed per sample in
physical (denormalised) units and averaged over the test set,
\begin{equation}
\mathcal{E} = \frac{100}{M_{\mathrm{test}}}\sum_{i=1}^{M_{\mathrm{test}}}
\frac{\lVert \widehat{\mathbf{y}}^{(i)} - \mathbf{y}^{(i)} \rVert_2}{\lVert \mathbf{y}^{(i)} \rVert_2},
\label{eq:app_relL2}
\end{equation}
which equals $100\times$NRMSE under the definition used in
\Cref{sec:experiments}. Because \Cref{eq:app_relL2} normalises by the
magnitude of the target rather than by its variation, it is generous to fields
carrying a large constant offset; all three thermal cases sit on a pedestal ---
$\sim\!300$~K for the heat sink and HIVE, $280$~K for the heated duct --- and the
per-dataset notes below give the scale of that effect.

\subsection{Heat sink}
\label{app:heat_sink}

\paragraph{Governing equation.}
Steady conduction through a finned aluminium heat sink cooling a CPU,
\begin{equation}
-\nabla\!\cdot\!\left(\kappa \nabla T\right) = Q,
\qquad
-\kappa \frac{\partial T}{\partial n} = h\,(T - T_\infty)
\ \ \text{on}\ \Gamma_{\mathrm{lat}} \cup \Gamma_{\mathrm{top}},
\label{eq:app_heatsink}
\end{equation}
with forced convection on the lateral surfaces, free convection on the top
surface and the chip heat load applied on the baseplate underside. The material
conductivity $\kappa$, the load $Q$, the film coefficient $h$ and the ambient
$T_\infty$ are held \emph{fixed} across the whole dataset. Only the geometry
varies, so $\boldsymbol{\mu}$ fully determines the solution and the benchmark is a
direct test of whether a conditional operator can read a 3D shape.

\paragraph{Parameter scan.}
A complete $5^4 = 625$ factorial over four design variables at five levels each
(\Cref{tab:app_heatsink_params}). The baseplate is a fixed 26~mm along $x$
(widened only when $W > 26$~mm), so the fin pitch is $26\,\mathrm{mm}/n_{\rm fin}
\in [3.25, 6.5]$~mm and is not an independent variable. This is the source of the
benchmark's difficulty for grid-based methods: fins 1.5--2.5~mm thick at a
3.25--6.5~mm pitch alias badly on any uniform latent grid, and at least 16 cells
along $x$ are needed for eight fins to separate at all.

\begin{table}[htbp]
\centering
\footnotesize
\begin{tabular}{llll}
\toprule
\textbf{Parameter} & \textbf{Symbol} & \textbf{Levels} & \textbf{Meaning} \\
\midrule
Sink width      & $W$          & $20, 22.5, 25, 27.5, 30$~mm   & $y$ extent; fins span it fully \\
Fin height      & $H$          & $15, 16.25, 17.5, 18.75, 20$~mm & $z$ extent above the baseplate \\
Thickness       & $t$          & $1.5, 1.75, 2, 2.25, 2.5$~mm  & fins and baseplate; $z$ spans $[-t, H]$ \\
Fin count       & $n_{\rm fin}$& $4, 5, 6, 7, 8$               & plates stacked along the 26~mm base \\
\bottomrule
\end{tabular}
\caption{Heat sink: the $5^4$ factorial design. All 625 combinations are
present exactly once.}
\label{tab:app_heatsink_params}
\end{table}

\paragraph{Mesh and provenance.}
The dataset is the heat-sink benchmark of
\citet{zhong2026pgdnoneuraloperatorprecomputed}, distributed as a MATLAB struct
of nodal coordinates and temperatures with no connectivity. Every case carries
its own tetrahedral mesh, from 4{,}671 to 32{,}645 nodes (median 12{,}673, with
613 distinct node counts among 625 cases), so node index $i$ is an unrelated
physical location in different cases. At the node budget used here,
$N = 16384$, the 193 cases with larger meshes are drawn without
replacement and the remaining 432 are duplicate-padded; the padding factor
reaches $3.5\times$ on the smallest mesh, and averaged over the corpus 76\% of
the budget is distinct nodes. The pooled bounding box is $x \in [0, 30]$,
$y \in [-30, 0]$, $z \in [-2.5, 18.5]$~mm; each design fills its own sub-box of
that domain, and that occupancy \emph{is} the geometry signal, which is why
coordinates are min--max scaled per axis over the pooled cloud rather than per
sample.

\paragraph{Learning setup.}
$\mathbf{F} \in \mathbb{R}^{625\times 4\times 16384}$ with channels $(x, y, z, T)$
and $c_{\mathrm{in}} = 3$, so inputs are $\mathbb{R}^{N\times 3\times 16384}$ and
targets $\mathbb{R}^{N\times 1\times 16384}$; $\boldsymbol{\mu} \in
\mathbb{R}^{625\times 4}$. Coordinates are dropped as features and re-enter only
through the 99-channel integrated encoding, so the network sees geometry and
$\boldsymbol{\mu}$ and nothing else. The split is 70/30 (438 train, 187 test) at
batch size 8, matching the test share used by the source paper
\citep{zhong2026pgdnoneuraloperatorprecomputed}.

\paragraph{Data statistics.}
$T \in [296.46, 343.76]$~K with mean 307.25~K and standard deviation 5.79~K
($\mathrm{Skew} = 1.25$, $\mathrm{Kurt} = 5.60$). Case-mean temperatures span
$[300.07, 322.86]$~K. The measurements are taken on the temperature rise above
ambient, $T - T_{\min}$ with $T_{\min} = 296.46$~K. 

\begin{figure}[tbp]
\centering
\includegraphics[width=0.78\linewidth]{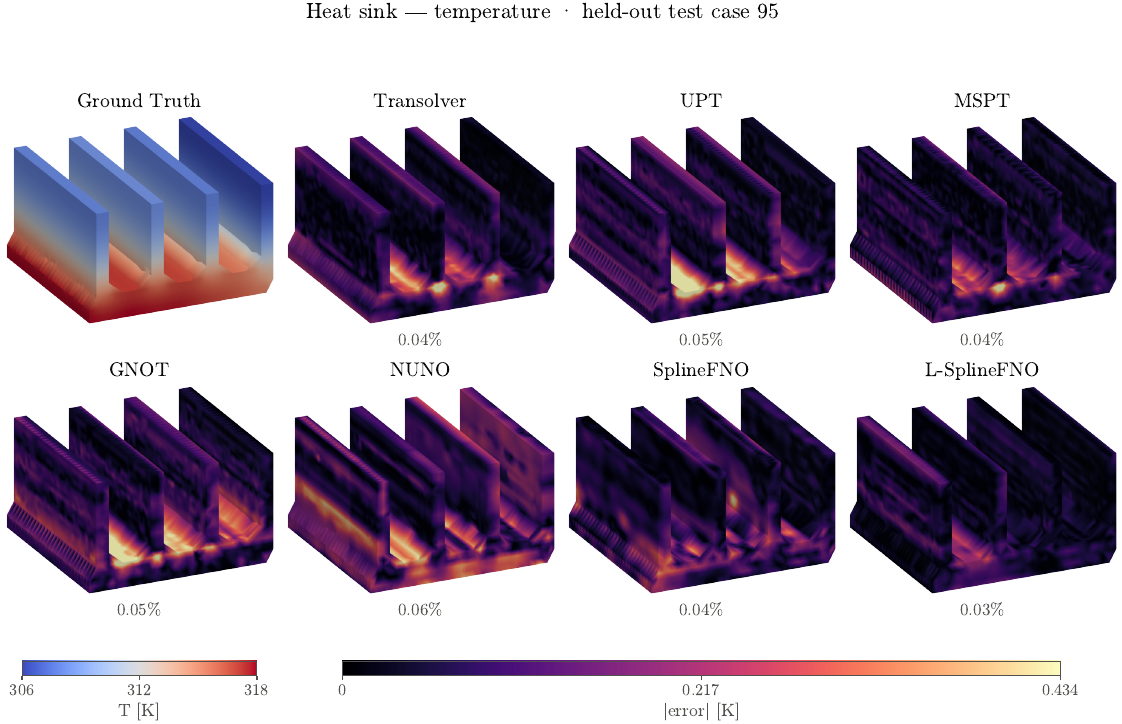}
\caption{Error maps produced by various architectures in modelling heat distribution across the heat sink.}
\label{fig:heat_sink_error_maps}
\end{figure}

\subsection{Bracket}
\label{app:bracket}

\paragraph{Governing equations.}
Steady elasto-plastic deformation of a lug, i.e. a hollow beam with a bolt hole
which is fixed at one end and loaded by a pressure over a marked area:
\begin{equation}
\nabla\!\cdot\!\boldsymbol\sigma = \mathbf{0},
\qquad
\boldsymbol\sigma = \mathbb{C}\left(\boldsymbol\varepsilon -
\boldsymbol\varepsilon^{\mathrm{p}}\right),
\qquad
\boldsymbol\varepsilon = \tfrac{1}{2}\!\left(\nabla \mathbf{u} +
\nabla \mathbf{u}^{\!\top}\right),
\label{eq:app_bracket}
\end{equation}
with an elastic--plastic constitutive law under isotropic hardening and the
small-deformation assumption, solved with quadratic hexahedral elements. The
prediction target is the nodal stress field. This is the only non-thermal,
non-fluid problem in the set.

\paragraph{Why this dataset earns its place.}
$\boldsymbol{\mu}$ splits cleanly into a redundant half and a non-redundant one:
the three geometry parameters restate what the node coordinates already show,
while the applied pressure is \emph{invisible} in the mesh --- two designs can
share a geometry and differ only in load. It is the only benchmark here where
half of $\boldsymbol{\mu}$ is inferable from the input and half is not, and the
loader exposes each half separately so the two contributions can be ablated
independently.

\paragraph{Parameter scan and normalisation.}
$N = 3000$ designs over four continuous parameters: length, thickness and hole
radius (geometry) plus the applied pressure (loading), each realised at 3000
distinct values. The published release \citep{zhong2025comprehensive} is
\emph{already} min--max normalised by its authors: coordinates, design parameters
and stress all arrive on $[0,1]$, and the scaling factors for coordinates and
parameters are not distributed, so physical units cannot be recovered for those
channels. The normalisation is global rather than per sample, so each design
occupies its own sub-box of the unit cube --- mean per-design spans and relative scale between designs is preserved. The latent grids used for this dataset are consequently isotropic: the pooled cloud fills the unit cube exactly, and the true physical aspect ratio is unknowable.

\paragraph{Mesh and learning setup.}
Every design has its own hexahedral mesh, 5{,}054 to 15{,}210 nodes (median
9{,}405), with no node correspondence between designs. At the node budget used
here, $N = 8192$, the 2{,}122 designs with larger meshes are
drawn without replacement and the remaining 878 are duplicate-padded, at worst
$1.6\times$; averaged over the corpus 97\% of the budget is distinct nodes, so
padding is a marginal effect on this dataset in a way it is not on the heat sink.
$\mathbf{F} \in \mathbb{R}^{3000\times 4\times 8192}$ with channels
$(x, y, z, \hat\sigma)$ and $c_{\mathrm{in}} = 3$; $\boldsymbol{\mu} \in
\mathbb{R}^{3000\times 4}$. Split 70/30 (2100 train, 900 test) at batch size 8.
The random split is sound here: 3000 independent designs with no grouping
structure to leak across, unlike the aircraft case.

\paragraph{Data statistics.}
$\hat\sigma$ fills $[0, 1]$ by construction of the authors' normalisation, with
mean 0.197 and standard deviation 0.121 ($\mathrm{Skew} = 1.19$, $\mathrm{Kurt} = 3.92$);
the 99th percentile is 0.534, so the upper half of the normalised range is
populated only by stress concentrations at the hole and the fixed end. Case-mean
stresses span $[0.086, 0.383]$. The metric is not degenerate here, so unlike the two temperature datasets, the stress field carries spatial structure that no constant can fake. See \Cref{fig:app_target_dists} for the distribution.

\begin{figure}[tbp]
\centering
\includegraphics[width=0.68\linewidth]{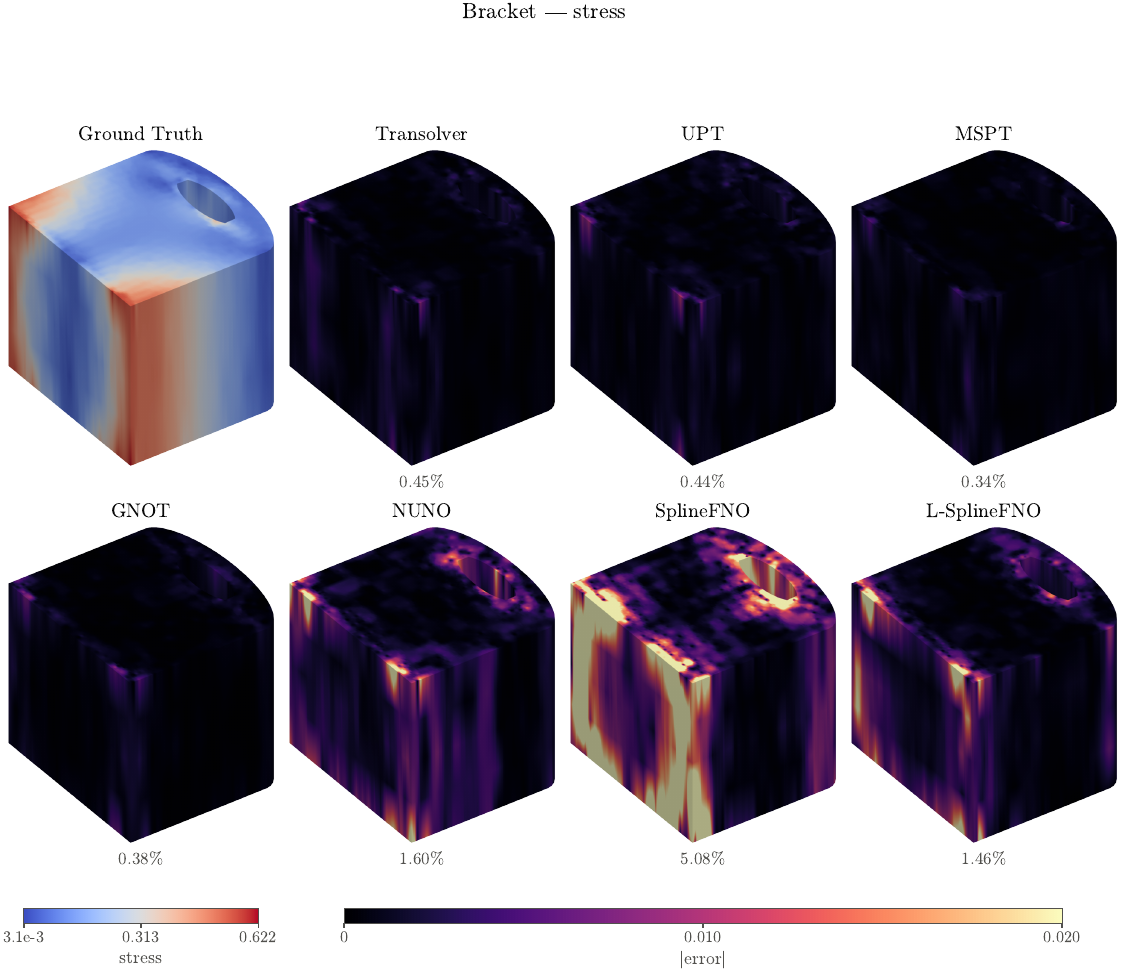}
\caption{Error maps produced by various architectures in modelling stress across the bracket.}
\label{fig:bracket_error_maps}
\end{figure}

\subsection{Aircraft}
\label{app:aircraft}

\paragraph{Governing equations and flight envelope.}
Steady compressible external flow over a parameterised airframe, with the
solution sampled on the surface triangulation. The released fields are the
freestream-normalised density, the three velocity components and the pressure.
The envelope is supersonic to hypersonic, with shocks present on the surface:
30 airframes are each solved at five flight conditions,
\begin{equation*}
(\mathrm{Ma}, \alpha, \beta) \in
\{(2,0,0),\ (7,0,0),\ (7,0,2),\ (7,7,0),\ (7,7,2)\},
\end{equation*}
with $\alpha$ the angle of attack and $\beta$ the sideslip angle, both in
degrees, giving $N = 150$ cases.

\paragraph{Why this dataset earns its place.}

All five flight conditions of one airframe share bit-identical coordinates, so the
point cloud carries \emph{no} information about the flight condition: only
$\boldsymbol{\mu}$ distinguishes them. This is the opposite extreme from the heat
sink, where $\boldsymbol{\mu}$ merely restates the shape. Unlike HIVE and the heated
duct, where the mesh is fixed outright, the geometry here still varies across
airframes as we subsample them, so the model must use both: the shape to tell airframes apart, and
$\boldsymbol{\mu}$ to tell flight conditions apart.

\paragraph{Mesh and provenance.}
The data ships with Transolver++ \citep{luo2025transolveraccurateneuralsolver} as Tecplot ASCII
\texttt{FEPOINT} blocks and is used as one of the six industrial benchmarks of
\citet{zhong2026pgdnoneuraloperatorprecomputed}. Every case has the same
331{,}971 surface nodes and 663{,}938 triangles; the airframes are deformations
of a shared template (mean node displacement between the two extreme geometries
is 1.3\% of the bounding-box diagonal, maximum 13\%), so node index $i$ is the
same semantic vertex everywhere. We subsample all 150 cases at the \emph{same}
131{,}072 node indices --- 39\% of the surface. The pooled bounding box over the
sampled nodes is $x \in [0, 5000]$, $y \in [-298, 774]$, $z \in [-2320, 2320]$: a
thin sheet in $y$, which is why coordinates are min--max scaled per axis; a
global scaling would leave the $y$ axis of a latent grid nearly empty. Per-case
force coefficients $(C_A, C_N, C_Z, C_l, C_n, C_m)$ are carried through, with the
lift coefficient $C_L = C_N\cos\alpha - C_A\sin\alpha$ precomputed, so the
aggregate quantity of interest needs no surface integration.

\paragraph{Learning setup.}
$\mathbf{F} \in \mathbb{R}^{150\times 8\times 131072}$ with channels
$(x, y, z, \rho, u, v, w, p)$ and $c_{\mathrm{in}} = 3$, giving inputs
$\mathbb{R}^{N\times 3\times 131072}$ and targets
$\mathbb{R}^{N\times 5\times 131072}$; $\boldsymbol{\mu} \in
\mathbb{R}^{150\times 3}$. Split 70/30 (105 train, 45 test) at batch size 4.

\paragraph{Split caveat.}
The split is a random permutation over the 150 cases, so the same airframe
appears in train and test under different flight conditions. With only 30
airframes, almost every test geometry has been seen. The reported numbers
therefore measure flight-condition interpolation on known shapes, not geometry
generalisation.

\paragraph{Data statistics.}
Alongside the histogram in \Cref{fig:app_target_dists}, per-channel statistics are
given in \Cref{tab:app_aircraft_stats}. The channels are strikingly unequal.
The stream-wise velocity $u$ is nearly determined by the flight condition alone whereas the transverse components $v$ and $w$ are largely geometry responses (67.8\% and 61.3\%). The
pressure is the most heavy-tailed target in the whole benchmark set
($\mathrm{Kurt} = 108$, reaching $46\,p_\infty$ at the shock), and it is the channel
that drives the joint score.

\begin{table}[htbp]
\centering
\footnotesize
\begin{tabular}{lrrrrrrr}
\toprule
\textbf{Channel} & \textbf{min} & \textbf{max} & \textbf{mean} & \textbf{std} &
$\mathrm{Skew}$ & $\mathrm{Kurt}$ & \makecell{\textbf{per-condition}\\\textbf{mean} (\%)} \\
\midrule
$\rho$ & $-0.017$ & 18.24 & 1.315 & 0.867 & 2.33 & 16.0  & 35.95 \\
$u$    & $-0.324$ & 7.670 & 5.665 & 1.957 & $-1.42$ & 3.20 & 5.25 \\
$v$    & $-3.044$ & 3.500 & 0.245 & 0.522 & $-0.13$ & 4.30 & 67.77 \\
$w$    & $-3.900$ & 4.070 & 0.063 & 0.714 & $-0.05$ & 5.00 & 61.33 \\
$p$    & $-0.035$ & 46.46 & 1.650 & 2.131 & 8.17 & 107.8 & 46.73 \\
\midrule
joint  & $-3.900$ & 46.46 & 1.788 & 2.471 & 2.08 & 14.3 & 23.30 \\
\bottomrule
\end{tabular}
\caption{Aircraft: per-channel statistics over all 150 cases at the 131{,}072
sampled nodes. All quantities are freestream-normalised and dimensionless. The
last column is the relative $L_2$ error of predicting the mean field of the
matching flight condition, i.e.\ the score of a predictor that uses
$\boldsymbol{\mu}$ perfectly and ignores the geometry; a model must beat 23.3\% on
the joint field before it has learned anything from the airframe shape.}
\label{tab:app_aircraft_stats}
\end{table}

\begin{figure}[tbp]
\centering
\includegraphics[width=0.78\linewidth]{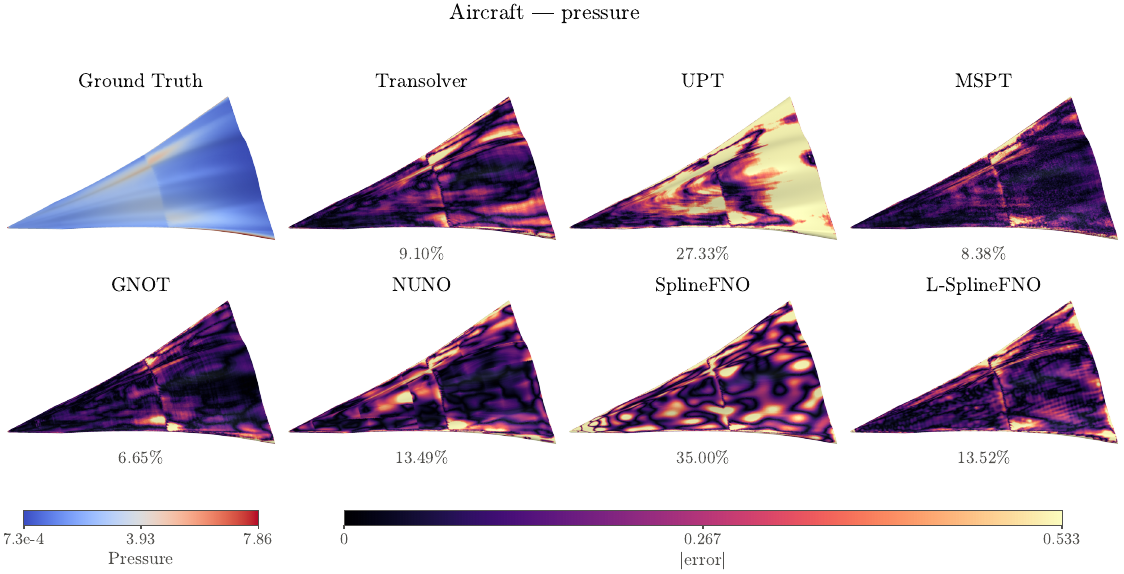}
\caption{Error maps produced by various architectures in modelling the pressure distribution across the aircraft.}
\label{fig:aircraft_error_maps}
\end{figure}

\subsection{Building fire: smoke and heat transport in a building}
\label{app:fds}

\paragraph{Governing equations.}
A 500~kW methane fire in a ventilated three-storey building, computed with
Fire Dynamics Simulator (FDS~6) \citep{FDS_Mcgrattan2013} in large-eddy
simulation mode. FDS solves the low-Mach-number form of the reacting
Navier--Stokes equations for thermally driven flow,
\begin{equation}
\begin{gathered}
\partial_t \rho + \nabla\!\cdot\!(\rho\mathbf{u}) = 0,
\qquad
\partial_t(\rho Z_\alpha) + \nabla\!\cdot\!(\rho Z_\alpha\mathbf{u})
= \nabla\!\cdot\!(\rho D_\alpha \nabla Z_\alpha) + \dot m'''_\alpha, \\
\partial_t(\rho\mathbf{u}) + \nabla\!\cdot\!(\rho\mathbf{u}\mathbf{u})
+ \nabla\tilde p = (\rho - \rho_0)\,\mathbf{g} + \nabla\!\cdot\!\boldsymbol\tau, \\
\partial_t(\rho h_s) + \nabla\!\cdot\!(\rho h_s\mathbf{u})
= \frac{\mathrm{D}\bar p}{\mathrm{D}t} + \dot q'''
- \nabla\!\cdot\!\dot{\mathbf{q}}'',
\qquad
\bar p = \rho R T \textstyle\sum_\alpha Z_\alpha / W_\alpha .
\end{gathered}
\label{eq:app_fds}
\end{equation}
The three lumped species are
$\alpha \in \{\text{air}, \text{fuel}, \text{products}\}$, with mass fractions
$Z_\alpha$. Here, $h_s$ is sensible enthalpy, $\dot q'''$ is the volumetric
heat-release rate, and $\dot{\mathbf{q}}''$ contains conductive, diffusive and
radiative heat fluxes. Combustion is a single-step, mixing-controlled reaction
(soot yield 0.01, CO yield 0.005) and radiation is a grey-gas solve.
FDS uses the sensible-enthalpy balance to constrain
$\nabla\!\cdot\!\mathbf{u}$ rather than advancing it explicitly, and recovers
temperature from the equation of state.

The flow is turbulent and unsteady, so we predict a late-time average rather
than an instantaneous field,
\begin{equation}
\bar\phi(\mathbf{x}) = \frac{1}{120\,\mathrm{s}}\int_{480\,\mathrm{s}}^{600\,\mathrm{s}}
\phi(\mathbf{x}, t)\,\mathrm{d}t ,
\label{eq:app_fds_avg}
\end{equation}
taken over a window beginning 450~s after the burner reaches full output.
The averaged field is treated as a static learning target ($\Delta t = 0$).

\paragraph{Target fields.}
We predict three cell-centred averages: the temperature rise above ambient,
$\Delta T = T - 293.15$~K, and the partial densities
$(\rho_{\rm air},\rho_{\rm prod})$ of the lumped air and combustion-products
species. These partial densities measure the mass of each mixture per unit
volume; $\rho_{\rm prod}$ tracks smoke transport through the building.

\paragraph{Scenario.}
The building measures $12 \times 12 \times 10$~m and comprises three levels:
a car park at $0$--$4$~m, a middle level at $4.25$--$7$~m, and an upper
level at $7.25$--$10$~m. Concrete slabs separate the levels, with four
openings connecting the car park to the middle level and a
$6 \times 6$~m atrium void connecting the middle and upper levels.
The upper floor is partitioned into adjoining rooms surrounding the atrium
void.
A floor-mounted methane burner has a $0.5 \times 0.5$~m burning surface
at $z=0.25$~m above the car-park floor. Its prescribed heat-release rate
increases linearly over the first 30~s to $2000$~kW\,m$^{-2}$ ($500~kW$).
Two $1.5 \times 1.0$~m wall vents on opposite faces provide push--pull
ventilation: for $Q > 0$, air enters through the vent at $x = 0$ and exits
through the vent at $x = 12$~m; $Q < 0$ reverses the direction.
The enclosure is otherwise closed to gas flow.
Ambient conditions are $20\,^\circ$C with no stratification.
$\boldsymbol{\mu}$ contains all prescribed case-to-case variation.

\paragraph{Parameter scan.}
Seven randomly sampled parameters control the burner position,
the signed ventilation rate, and the position of each wall vent
(\Cref{tab:app_fds_params}). The six positional parameters are restricted
to a $0.25$~m lattice to maintain alignment with the computational mesh,
whereas the signed ventilation rate is sampled continuously.
The corpus contains 547 cases.

\begin{table}[htbp]
\centering
\footnotesize
\setlength{\tabcolsep}{4pt}
\begin{tabular}{llllll}
\toprule
\textbf{Parameter} & \textbf{Symbol} & \textbf{Range} & \textbf{Discretisation} & \textbf{Group} & \textbf{Visible in the mesh} \\
\midrule
Burner centre, $x$     & $x_{\rm f}$ & $[2.25, 9.75]$~m              & $0.25$~m   & fire        & yes, exactly \\
Burner centre, $y$     & $y_{\rm f}$ & $[2.25, 9.75]$~m              & $0.25$~m   & fire        & yes, exactly \\
Ventilation rate      & $Q$         & $[-4.0, 4.0]$~m$^3$\,s$^{-1}$ & continuous & ventilation & no \\
Left vent centre, $y$  & $y_{\rm L}$ & $[1.0, 11.0]$~m               & $0.25$~m   & ventilation & no \\
Left vent centre, $z$  & $z_{\rm L}$ & $[4.75, 6.5]$~m               & $0.25$~m   & ventilation & no \\
Right vent centre, $y$ & $y_{\rm R}$ & $[1.0, 11.0]$~m               & $0.25$~m   & ventilation & no \\
Right vent centre, $z$ & $z_{\rm R}$ & $[4.75, 6.5]$~m               & $0.25$~m   & ventilation & no \\

\bottomrule
\end{tabular}
\caption{Building fire: the seven conditioning parameters, their sampling ranges,
and positional discretisation for the 547-case corpus. Left and right denote
the vents at $x=0$ and $x=12$~m, respectively. The last column indicates whether
each parameter is recoverable from the node coordinates; see the redundancy
discussion below.}
\label{tab:app_fds_params}
\end{table}

\paragraph{Mesh.}
FDS discretises the domain into 11 rectilinear blocks containing
116{,}928 Cartesian cells in three cell classes
(\Cref{tab:app_fds_cells}). Of these, 111{,}237 are gas cells and 5{,}691 lie
inside solid obstructions. Solid cells are excluded from the learning point
clouds, targets, field statistics and error metrics. The counts and index
ranges below describe the full Cartesian arrays, before either solid-cell
masking or refinement-box removal.
A $24 \times 24 \times 32$ refinement block, centred horizontally on the
burner and extending from $z=0$ to $4$~m, uses cubic cells of side
$0.125$~m. Outside this block, four vertically refined blocks with
flattened $0.25 \times 0.25 \times 0.125$~m cells form a
layer at $z=3$--$4$~m, immediately below the intermediate floor.
The remaining six blocks use cubic cells of side $0.25$~m. We use cell
centres as nodes; their pooled bounding box is
$x,y \in [0.125,11.875]$~m and $z \in [0.0625,9.875]$~m.
The refinement box follows the burner, resizing the surrounding car-park
blocks and producing 415 distinct point clouds, all with the same node count.
These cases therefore require per-sample coordinates,
as with the heat sink and bracket cases. The middle and upper
meshes are bit-identical across cases, and the concatenated ordering preserves
fixed storey ranges: car park $[0,61632)$, middle $[61632,89280)$ and upper
$[89280,116928)$.

\begin{table}[htbp]
\centering
\footnotesize
\begin{tabular}{lrrl}
\toprule
\textbf{Cell dimensions [m]} & \textbf{Cells} & \textbf{Volume [m$^3$]} & \textbf{Where} \\
\midrule
$0.125^3$                     & $18{,}432$ (15.8\%) & $36$ (2.5\%)    & refinement box around the burner \\
$0.25\times0.25\times0.125$   & $17{,}280$ (14.8\%) & $135$ (9.4\%)   & $z=3$--$4$~m excluding refinement box \\
$0.25^3$                      & $81{,}216$ (69.5\%) & $1269$ (88.1\%) & everywhere else \\
\bottomrule
\end{tabular}
\caption{Building fire: the three cell classes of the 116{,}928-cell mesh
(1440~m$^3$ in total, including solid obstructions), before solid-cell masking
and refinement-box removal.
On this full Cartesian mesh, an unweighted node average over-weights the refinement box
by $\sim\!6.3\times$ relative to its volume, so integral quantities of interest
over the gas domain must exclude solid cells and be volume-weighted; the loader
carries the cell volume per node for this purpose.}
\label{tab:app_fds_cells}
\end{table}

\paragraph{Why this dataset earns its place.}
Its conditioning vector splits in a way no other benchmark here does. The
refinement box is centred on the burner, so the centre of the finest cells
reproduces $(x_{\rm f}, y_{\rm f})$ exactly. The five ventilation parameters,
by contrast, move no cell centre at all. Two dimensions of
$\boldsymbol{\mu}$ are therefore recoverable from the geometry, and five are
invisible in it. The unconditioned arm is consequently not a 
geometry-only model; it knows exactly where the fire is and nothing about the ventilation.
As on the bracket, the loader exposes the fire and ventilation parameter groups of
$\boldsymbol{\mu}$ separately so that they can be ablated apart.

The case is also \emph{multi-region}, in a different sense from the heated duct.
The three storeys are separated by slabs that the flow crosses only through the
openings, so the fields change near-discontinuously in value across a slab
rather than kinking in gradient across a material interface ---
but the difficulty for a single $C^2$
basis on a single latent grid is the same. The storeys serve as the region set for
the per-region encoding and decoding branches described for the heated duct.
The 11 FDS blocks cannot be used for this, because four of their boundaries move
with the fire and block membership is therefore a per-sample property.

\paragraph{Prediction domain: the smoke, not the flame.}
A fire-safety engineer does not run this simulation to learn what happens
inside the flame. The design fire is prescribed: the burner's position and
output are inputs to the study, not results of it. The question is whether the
rest of the building stays tenable, meaning how hot the gas gets, how far the
smoke spreads and how much combustion product reaches the car park and the
upper storeys, where occupants are and where the escape routes run. The error
metric pulls the other way. The $18{,}400$ gas nodes of the refinement box
around the burner are 16.5\% of the gas-cell cloud but carry roughly two
thirds of the error energy, because they
contain the flame itself ($\Delta T$ up to 1002~K there, against 189~K
everywhere else). With them left in, every joint metric on this case would
score the flame, which is the one region the engineer does not need predicted.

We therefore remove the box's gas nodes from the prediction domain, leaving
$92{,}837$ nodes: $43{,}200$ in the car park and $49{,}637$ in the middle and
upper storeys. The task becomes predicting heat and combustion-product
transport into the occupied building, away from the source. Removing the box
does not hide where the fire is, because the surrounding mesh blocks still
abut it and the retained cloud still moves with the burner. It does make that
information far less accessible. The two fire-position dimensions of
$\boldsymbol{\mu}$ can still be recovered from the geometry, but they are no
longer cheap to extract \citep{panayis2025leveragingaimodellingfds}. 

\paragraph{Learning setup.}
$\mathbf{F} \in \mathbb{R}^{547\times 6\times 98496}$ with channels
$(x, y, z, \Delta T, \rho_{\rm air}, \rho_{\rm prod})$ and
$c_{\mathrm{in}} = 3$, giving inputs $\mathbb{R}^{N\times 3\times 98496}$ and
targets $\mathbb{R}^{N\times 3\times 98496}$; $\boldsymbol{\mu} \in
\mathbb{R}^{547\times 7}$. The coordinates are consumed as geometry by the
per-sample path, so the network again sees the 99-channel encoding and
$\boldsymbol{\mu}$ only. One departure from the common convention: because this
mesh has two cell sizes, the integrated positional encoding is given a per-node
footprint, attenuating frequency band $f$ by $\exp(-\tfrac12 f^2\sigma^2)$ with
$\sigma = h/\sqrt{12}$ the standard deviation of a uniform distribution over the
node's cell of side $h$ (in normalised units); on every other benchmark
$\sigma = 0$ and the encoding reduces to the plain sinusoidal one. The storey index
is carried alongside the mesh for the region branches and is not an input.
The split is a random 70/30 permutation (383 train, 164 test) at batch size 4.
It is sound here, unlike on the aircraft: the cases were submitted in batches,
but the within-batch parameter spread matches the corpus spread on all seven
dimensions, so the batches are not design clusters.

\paragraph{Data statistics.}
Per-channel statistics over the retained gas nodes are given in
\Cref{tab:app_fds_stats}. The three targets sit at very different scales, so a
joint relative $L_2$ over the raw channels (\Cref{eq:app_relL2}) comes down to
the error in $\Delta T$ alone. We therefore report the per-variable normalised
error. Unlike the other thermal cases, the temperature target is a rise over
ambient and carries no pedestal. The cases differ substantially from one
another: the case-mean temperature rise and the case-mean product density each
vary by roughly a factor of two across the corpus. Air density varies least,
only a few percent about its mean, and that small spread makes it the hardest
of the three targets to predict in normalised terms.

\begin{table}[htbp]
\centering
\footnotesize
\begin{tabular}{lrrrr}
\toprule
\textbf{Channel} & \textbf{min} & \textbf{max} & \textbf{mean} & \textbf{std} \\
\midrule
$\Delta T$ [K]                   & $0$     & $189.4$  & $39.32$  & $16.62$  \\
$\rho_{\rm air}$ [kg\,m$^{-3}$]  & $0.763$ & $1.630$  & $1.247$  & $0.0845$ \\
$\rho_{\rm prod}$ [kg\,m$^{-3}$] & $0$     & $0.1007$ & $0.0462$ & $0.0164$ \\
\bottomrule
\end{tabular}
\caption{Building fire: per-channel statistics over all 547 cases at the
$92{,}837$ retained gas nodes (solid cells and refinement box removed).}
\label{tab:app_fds_stats}
\end{table}

\begin{figure}[H]
\centering
\includegraphics[width=0.78\linewidth]{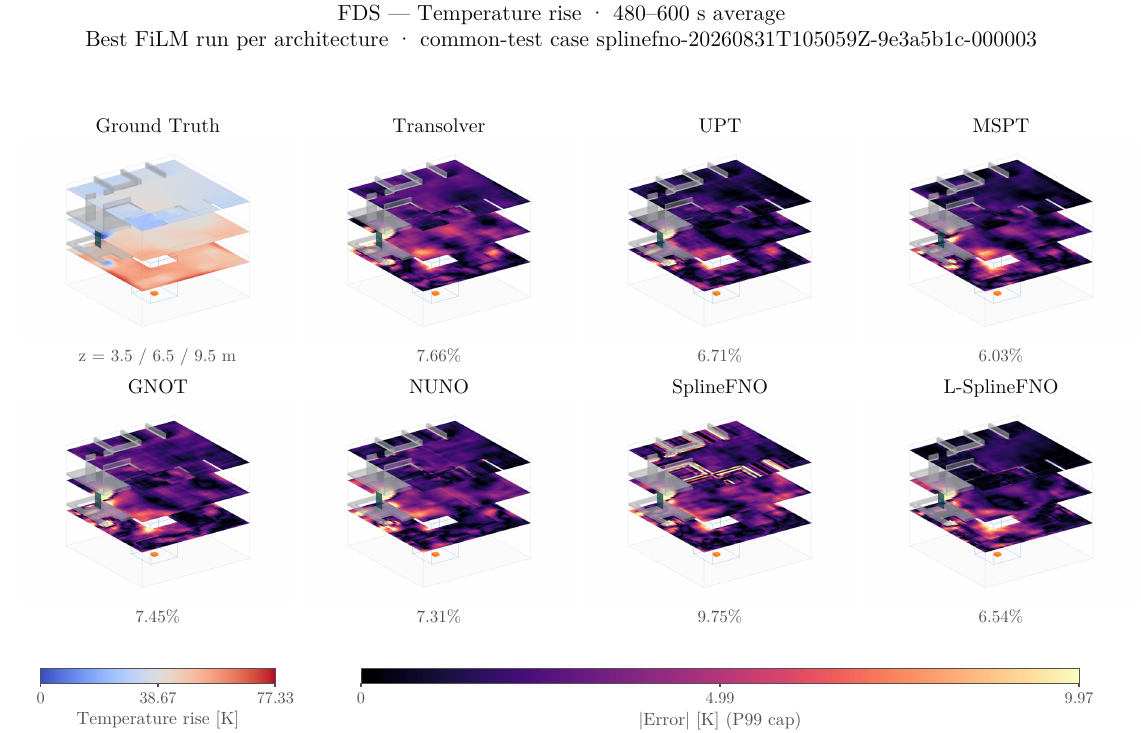}
\caption{Error maps produced by various architectures in modelling temperature change across the building due to a fire in the lower floor.}
\label{fig:fds_error_maps}
\end{figure}

\subsection{Heated duct: conjugate heat transfer}
\label{app:duct}
\paragraph{Governing equations.}
Steady conjugate heat transfer over three materials --- a laminar water duct, the
metal shell around it, and a volumetrically heated block beneath --- coupled only
through two interfaces:
\begin{align}
\text{fluid:}  &&  -\nabla\!\cdot\!\left(k_f \nabla T\right)
                 + \rho c_p\, \mathbf{U}\!\cdot\!\nabla T &= 0, \label{eq:app_duct_f}\\
\text{metal:}  &&  -\nabla\!\cdot\!\left(k_m \nabla T\right) &= 0, \label{eq:app_duct_m}\\
\text{heater:} &&  -\nabla\!\cdot\!\left(k_h \nabla T\right) &= q_h, \label{eq:app_duct_h}
\end{align}
subject at each interface to continuity of temperature and of normal heat flux,
\begin{equation}
T_a = T_b,
\qquad
k_a \frac{\partial T_a}{\partial n} = k_b \frac{\partial T_b}{\partial n} .
\label{eq:app_duct_iface}
\end{equation}
The flow is laminar ($\mathrm{Re} = 21$--$83$) and, with constant properties and
gravity switched off, the momentum equation does not depend on temperature ---
a fact with sharp consequences for the parameter scan, taken up below. Data were
generated with the OpenFOAM Foundation v13 \texttt{multiRegion/CHT/heatedDuct}
tutorial geometry \citep{OpenFOAM}, run to steady state with
\texttt{foamMultiRun}.

\paragraph{Why this dataset earns its place.}
It is the only multi-material case in the set. The three regions are three separate
meshes that share no cells: 40{,}560 fluid, 67{,}440 metal and 16{,}800 heater
cell centres, which the loader concatenates into one 124{,}800-node cloud so a
single operator spans all of them. They occupy disjoint sub-volumes of one
bounding box --- the heater sits below the duct at $y < 0$, the fluid inside it,
the metal in the shell between --- so the concatenated cloud is a partition of
the domain rather than an overlay, and region membership is a function of
position. That is why the region label can be withheld: the default configuration
feeds coordinates alone and the model must read the material off the geometry,
while a one-hot variant hands it over explicitly and measures what naming the
material is worth.

The interfaces are conformal to machine precision --- 15{,}360 matched
metal--fluid face pairs and 3{,}360 heater--metal pairs, with a maximum
pair separation of $8\times10^{-17}$~m --- so \Cref{eq:app_duct_iface} is imposed
on coincident points and the kink in $\nabla T$ is a property of the solution
rather than of the discretisation.

\paragraph{Geometry.}
The pooled bounding box is $x \in [0.5, 49.5]$, $y \in [-9.5, 49.5]$,
$z \in [0.5, 149.5]$~mm. The structure is cross-sectional and the geometry is
prismatic in $z$: the fluid occupies $x, y \in [5.6, 44.4]$~mm, leaving a 5.0~mm
metal shell, and the heater spans $x \in [8.0, 42.0]$, $y \in [-9.5, -0.6]$~mm.
Under the per-axis min--max scaling this puts the fluid--metal wall at $0.103$ and
$0.897$ along $x$, and at $0.255$ and $0.915$ along $y$, so a latent grid must
resolve roughly a tenth of its own extent to see the wall at all. This is why the
latent grids for this case are sized to feature scale in $x$ and $y$ rather than
to the $49 : 59 : 149$ bounding-box aspect, which would put the resolution along
the one axis that carries no structure.

\paragraph{Parameter scan.}
Five parameters, drawn on a Sobol sequence and listed in
\Cref{tab:app_duct_params}. Of 544 proposed designs, 146 converged and were
accepted; the rest were rejected for an estimated peak temperature above 900~K or
for failing the convergence gate. Acceptance is tight: heater power spans
131--1050~W, matched by the advected power to an energy imbalance of at most
$9.6\times10^{-4}$, with a maximum solver residual of $2.3\times10^{-5}$.

\begin{table}[htbp]
\centering
\footnotesize
\setlength{\tabcolsep}{4pt}
\begin{tabular}{lllll}
\toprule
\textbf{Parameter} & \textbf{Symbol} & \textbf{Physical range} & \textbf{Transform} & \textbf{Shape} \\
\midrule
Heater power density & $q_h$        & $2.5\times10^{6}$--$2.0\times10^{7}$~W\,m$^{-3}$ & $\log(q_h/10^{7})$    & no  \\
Inlet velocity       & $U_{\rm in}$ & $5\times10^{-4}$--$2\times10^{-3}$~m\,s$^{-1}$   & $U_{\rm in}/10^{-3}$  & yes \\
Inlet temperature    & $T_{\rm in}$ & $280$--$320$~K                                   & $(T_{\rm in}-300)/78.5$ & no \\
Heater conductivity  & $k_h$        & $40$--$120$~W\,m$^{-1}$K$^{-1}$                  & $k_h/80$              & yes \\
Metal conductivity   & $k_m$        & $100$--$300$~W\,m$^{-1}$K$^{-1}$                 & $k_m/200$             & yes \\
\bottomrule
\end{tabular}
\caption{Heated duct: the five conditioning parameters. $\boldsymbol{\mu}$ is the
dimensionless transform, not the raw value --- only the $\log$ on $q_h$, which
spans a factor of eight, is non-trivial. The last column records whether the
parameter changes the \emph{shape} of the temperature field or only rescales it;
see the degeneracy noted below.}
\label{tab:app_duct_params}
\end{table}

\paragraph{A degeneracy in two of the five dimensions.}
Because the momentum equation decouples from temperature, the coupled energy
problem is linear in $T$ and admits the exact separation
\begin{equation}
T(\mathbf{x};\, q_h, U_{\rm in}, T_{\rm in}, k_h, k_m)
= T_{\rm in} + q_h\,\Theta(\mathbf{x};\, U_{\rm in}, k_h, k_m) .
\label{eq:app_duct_affine}
\end{equation}
Two of the five parameters therefore carry no shape information at all: they are
an exact affine rescaling of the field, and only $U_{\rm in}$, $k_h$ and $k_m$
change $\Theta$. This has to be read into any conditioning ablation on this case,
because a method that captures the $q_h$/$T_{\rm in}$ affinity exactly and gets
$\Theta$ wrong will still post a respectable relative $L_2$.

\paragraph{The interface is where the gridded family is stressed.}
Temperature is continuous across a conjugate interface but its gradient is not,
and the jump is set by the conductivity ratio in \Cref{eq:app_duct_iface} ---
here $k_h \in [40, 120]$ against $k_m \in [100, 300]$~W\,m$^{-1}$K$^{-1}$. Every
gridded backbone in the comparison scatters all three regions onto \emph{one}
latent grid through \emph{one} $C^2$ basis, and no number of retained modes lets a
$C^2$ function reproduce a kink. The effect is measurable rather than notional:
on the $[48, 48, 32]$ latent grid, 24.0\% of covered knots receive mass from more
than one region, and on a coarser $[32, 32, 24]$ grid that rises to 37.8\%. Each
such knot holds a single number serving two materials, at exactly the location
where the kink should be. We treat this as a property of the benchmark to be
measured, not a defect to be hidden.

\paragraph{Split.}
The corpus ships a designed split --- 85 train, 18 validation, 19 interpolation
test and 24 extrapolation corners --- and the loader honours it by permuting the
samples so the pipeline's own seeded split lands exactly on the design. The
default mode trains on train plus validation (103 cases) and tests on the 19
interpolation cases, dropping the corners; that is the $N = 122$ recorded in
\Cref{tab:data_overview}. An extrapolation mode instead trains on all 122
and tests on the 24 corners, answering the harder question of performance outside
the sampled parameter box. A plain random split over all 146 is available but
leaks the corner design into training and is not used.

\paragraph{Learning setup.}
$\mathbf{F} \in \mathbb{R}^{122\times 1\times 124800}$ carrying $T$ alone, with
$c_{\mathrm{in}} = 0$: there are no field channels whatsoever, so inputs are
$\mathbb{R}^{N\times 0\times 124800}$ and the network sees only the 99-channel
encoded coordinates, $\mathbb{R}^{B_{\mathrm{batch}}\times 124800\times 99}$, with targets
$\mathbb{R}^{N\times 1\times 124800}$. $\boldsymbol{\mu} \in
\mathbb{R}^{122\times 5}$; batch size 4. The per-node region index is carried
alongside the mesh for masked losses and per-region metrics, but is not an input
under the default configuration.

\paragraph{Data statistics.}
$T \in [280.00, 880.63]$~K with mean 461.0~K and standard deviation 118.1~K. With
$\mathrm{Skew} = 0.87$ and $\mathrm{Kurt} = 3.01$ this is much the most nearly Gaussian
target in the set (\Cref{fig:app_target_dists}), which is a direct consequence of
\Cref{eq:app_duct_affine}: the field is a smooth $\Theta$ scaled and shifted by
two broadly sampled parameters, with none of the shock or stress-concentration
tails that make the aircraft and bracket targets heavy. Case-mean temperatures
span $[319.5, 732.2]$~K, a factor of 2.3, so the case-to-case variation is far
larger than on any other benchmark here.

\begin{figure}[H]
\centering
\includegraphics[width=0.78\linewidth]{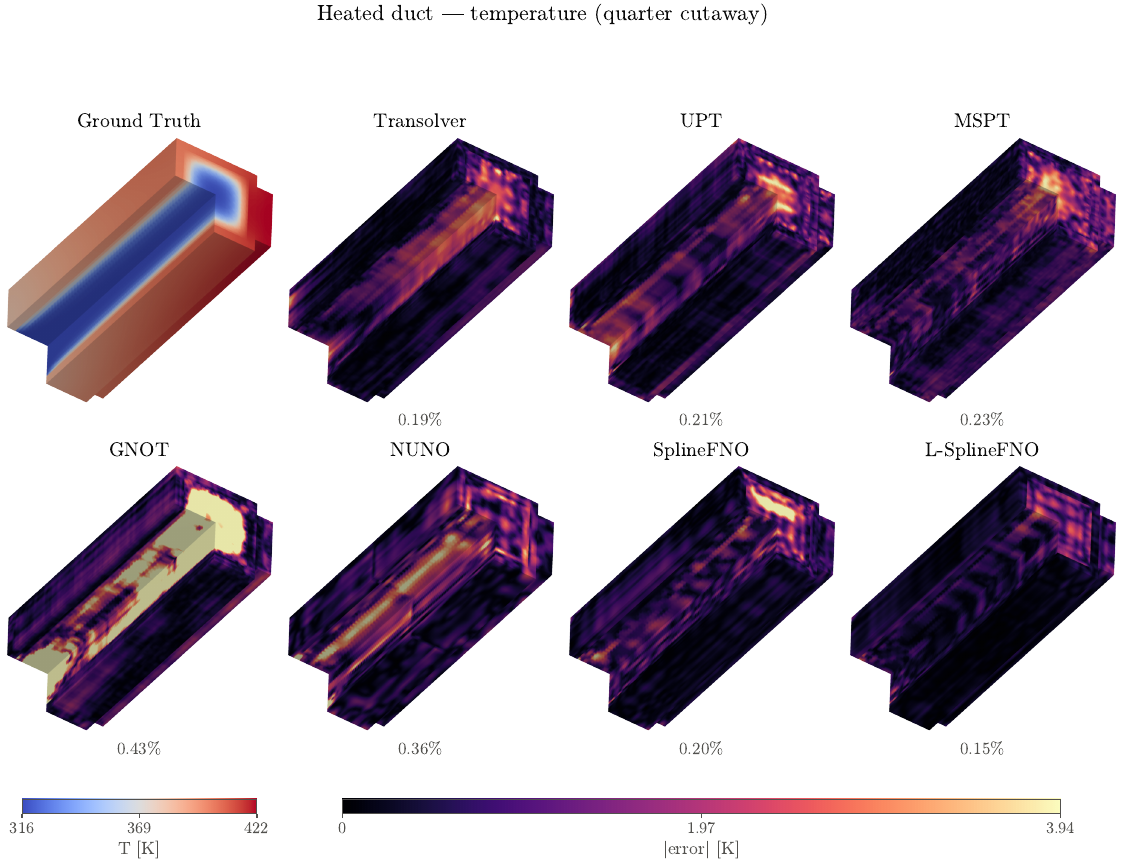}
\caption{Error maps produced by various architectures in modelling the temperature distribution from the conjugate heat transfer across the heated duct.}
\label{fig:heated_duct_error_maps}
\end{figure}

\subsection{HIVE monoblock}
\label{app:hive}

\paragraph{Physics.}
HIVE (Heating by Induction to Verify Extremes) is a UKAEA facility for testing
fusion divertor components under reactor-relevant heat loads
\citep{Hancock2018}. The simulation is a coupled multiphysics MOOSE model
\citep{MOOSE_Permann_2020,farscape_hive}: an induction coil drives eddy currents in a
divertor monoblock, and the resulting Joule heating is the source term
of a conduction problem \citep{hornejones2026virtualqualificationnuclearfusion},
\begin{equation}
\rho c_p \frac{\partial T}{\partial t}
= \nabla\!\cdot\!\left(\kappa \nabla T\right) + q_{\mathrm{EM}}(\mathbf{x}),
\qquad
-\kappa\frac{\partial T}{\partial n} = h\,(T - T_{\mathrm{cool}})
\ \ \text{on}\ \Gamma_{\mathrm{pipe}},
\label{eq:app_hive}
\end{equation}
run forward to convergence. The coolant-side film coefficient is set by a Nusselt
correlation of the standard form
\begin{equation}
\mathrm{Nu} = C\,\mathrm{Re}^{a}\,\mathrm{Pr}^{b},
\qquad h = \mathrm{Nu}\,\kappa_{\mathrm{f}} / D_{\mathrm{h}},
\label{eq:app_nusselt}
\end{equation}
whose three coefficients $(C, a, b)$ are the epistemically uncertain quantities
of the study, alongside the three components of the coil-to-target offset.

\paragraph{Parameter scan.}
Latin hypercube sampling over the six independent uniform priors of
\Cref{tab:app_hive_params}. The launched ensemble contains some failed runs;
we use the 186 that completed, listed in the release's completeness manifest. The
two groups act on the physics very differently --- the coil offsets move where
the induced heating is deposited, while the Nusselt coefficients rescale the
coolant-side film coefficient. 

\begin{table}[htbp]
\centering
\footnotesize
\setlength{\tabcolsep}{4pt}
\begin{tabular}{lllll}
\toprule
\textbf{Parameter} & \textbf{Group} & \textbf{Prior} & \textbf{Realised range} & \textbf{Role} \\
\midrule
\texttt{nusselt\_scale}         & HTC      & $\mathcal{U}(0.001, 0.031)$   & $[0.00112, 0.03090]$ & $C$ in \Cref{eq:app_nusselt} \\
\texttt{nusselt\_re\_power}     & HTC      & $\mathcal{U}(0.8, 0.9)$       & $[0.80043, 0.89950]$ & $a$ in \Cref{eq:app_nusselt} \\
\texttt{nusselt\_pr\_power}     & HTC      & $\mathcal{U}(0.35, 0.45)$     & $[0.35048, 0.44997]$ & $b$ in \Cref{eq:app_nusselt} \\
\texttt{coil\_target\_offset\_x}& geometry & $\mathcal{U}(-0.01, 0.01)$~m  & $[-0.00996, 0.00995]$ & coil position \\
\texttt{coil\_target\_offset\_y}& geometry & $\mathcal{U}(-0.01, 0.01)$~m  & $[-0.00994, 0.00996]$ & coil position \\
\texttt{coil\_target\_offset\_z}& geometry & $\mathcal{U}(0.019, 0.039)$~m & $[0.01901, 0.03892]$  & coil standoff \\
\bottomrule
\end{tabular}
\caption{HIVE: the six uncertain parameters, their sampling priors and the ranges
realised over the 186 completed runs.}
\label{tab:app_hive_params}
\end{table}

\paragraph{Mesh and a node-ordering subtlety.}
The solid meshes (coil and target) are identical across runs, with the coil
displaced relative to a stationary monoblock. A separate tool then builds a
vacuum mesh around the two bodies, and that step permutes the point ordering on
the monoblock. The release resolves this by taking the first sample's mesh as a template and storing, per run, the permutation that matches it; we apply those maps, after which all 186 runs share a single 194{,}691-node ordering. The preprocessed archive retains only that
template's coordinates --- and the coil is in any case not part of the stored
mesh --- so the per-run coil displacement is \emph{not} visible in the geometry
presented to the operator. It reaches the model through $\boldsymbol{\mu}$ alone,
as does everything else that distinguishes one run from another. The
bounding box is $x \in [-11.5, 11.5]$, $y \in [-75, 75]$,
$z \in [-11.5, 19.5]$~mm. Unlike the other 3D datasets, no spatial subsampling is
applied: all 194{,}691 nodes are used.

\paragraph{Learning setup.}
$\mathbf{F} \in \mathbb{R}^{186\times 1\times 194691}$ carrying the converged
$T_{\mathrm{final}}$ alone, with $c_{\mathrm{in}} = 0$: inputs are
$\mathbb{R}^{N\times 0\times 194691}$, the network sees only the 99-channel
encoded coordinates, $\mathbb{R}^{B_{\mathrm{batch}}\times 194691\times 99}$, and targets are
$\mathbb{R}^{N\times 1\times 194691}$. $\boldsymbol{\mu} \in
\mathbb{R}^{186\times 6}$; split 80/20 (149 train, 37 test) at batch size 5. No
spatial subsampling is applied, so this is the largest per-sample node count in
the set.

\begin{figure}[H]
\centering
\includegraphics[width=0.78\linewidth]{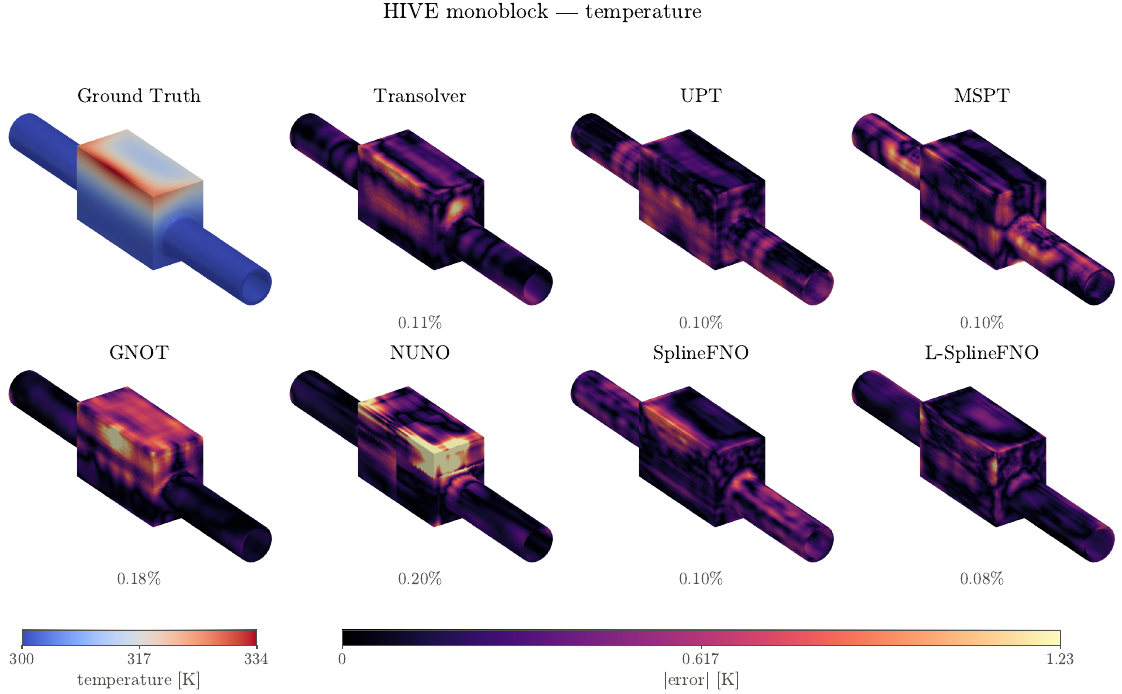}
\caption{Error maps produced by various architectures in modelling the temperature distribution from inductive heating in a monoblock in the HIVE case.}
\label{fig:heated_duct_error_maps_19}
\end{figure}

With one shared mesh and no field input, the model is fitting a
$6 \to \mathbb{R}^{194691}$ map from roughly 148 training points in a
six-dimensional design space. The conditioning module, not the backbone, is
expected to be the bottleneck --- which is precisely what makes this case a
measurement of the conditioning pathway rather than of the operator. The corpus
also ships the first stored timestep $T_{\mathrm{initial}}$; a companion
configuration feeds it in as a field channel at the same seed and split, so the
pair measures directly what an initial field is worth against the parameter
path.

\paragraph{Data statistics.}
$T_{\mathrm{final}} \in [300.00, 557.23]$~K with mean 341.43~K and standard
deviation 36.67~K ($\mathrm{Skew} = 1.30$, $\mathrm{Kurt} = 4.78$). Case-mean temperatures
span $[310.98, 414.67]$~K. As on the heat sink, the $\sim\!300$~K pedestal
flatters \Cref{eq:app_relL2}: a single global constant scores 9.3\% and a
per-case constant 6.5\%. A linear fit of $\boldsymbol{\mu}$ to the case mean
explains $R^2 = 0.89$ of the case-to-case variation and scores 6.9\%, so the six
parameters set the bulk thermal level well and what the operator must actually
supply is the spatial structure. For scale, the stored initial field
$T_{\mathrm{initial}}$ sits at $[294.06, 405.23]$~K and the rise to convergence
spans $[0.03, 162.58]$~K with mean 39.34~K --- the field the operator predicts is
far from the state the simulation started in.

\section{MAST-U: full-device structural reconstruction}
\label{app:mastu}

\paragraph{The device.}
MAST-U is the upgraded successor to MAST, the Mega-Ampere Spherical Tokamak at
UKAEA's Culham site \Citep{Milnes2015}. A spherical tokamak has a low
aspect ratio: the plasma wraps closely around a slender centre column, so
magnets, vessel, and in-vessel structures are packed into a compact volume.
MAST-U places many of its coils inside the main vacuum
vessel \Citep{Fiesta_Cunningham_2013}, and adds closed upper and lower
divertors that can run configurations from conventional to the long-leg
Super-X \Citep{Morris2018}. Its experimental programme targets the
integration of core, pedestal, and exhaust physics
\Citep{Harrison_2024}, and it informs the design of the Spherical Tokamak
for Energy Production (STEP) \Citep{Chapman2024}. Structurally, the
machine is a $6.8$~m tall, $6.2$~m wide assembly of roughly $110$~tonnes
\Citep{Ellis_2025}, held on four support legs.

\paragraph{Why a whole-device model.}
Fusion components are usually analysed one at a time, with loads from the
rest of the machine prescribed as boundary conditions. For a tightly coupled
assembly those boundary conditions are not known in advance: temperature,
stress, and magnetic loads vary with position and are related through the
operating state of the whole device, so a component-scale view makes the
mechanical equilibrium problem difficult even to pose. \Citet{Ellis_2025} demonstrates a 
single mechanical model of the complete MAST-U structure from $3{,}938$ CAD
volumes, using a conformal tetrahedral mesh and the solid-mechanics module of
MOOSE \citep{MOOSE_Permann_2020}, as a first step towards a structural digital
twin. The paper describes approximately $127$ million elements, whereas the
archived mesh used here contains $134{,}609{,}743$ four-node tetrahedra and
$29{,}183{,}754$ unique nodes. Each static solve takes about $500$ CPU-hours
on $600$--$800$ cores \Citep{Ellis_2025}. A digital twin must be queried far more
often than that cost allows, across load cases, operating scenarios, material
uncertainty, and design changes, which is the role a trained surrogate would
fill \Citep{neural_parareal_pamela2025}. We take the first step of that
programme: a surrogate conditioned on the two parameters of a thermal-load
parameter study.

\paragraph{Governing equations.}
For small-strain linear elasticity, displacement
$\mathbf u:\Omega\to\mathbb R^3$ satisfies
\begin{equation}
\nabla\!\cdot\boldsymbol\sigma+\rho\mathbf g=\mathbf 0,
\qquad
\boldsymbol\sigma=\mathbb C(\omega):
\bigl(\boldsymbol\varepsilon(\mathbf u)-\boldsymbol\varepsilon^{\mathrm{th}}\bigr),
\qquad
\boldsymbol\varepsilon^{\mathrm{th}}=\alpha\,\bigl(T(\mathbf x)-T_{\mathrm{sf}}\bigr)\mathbf I,
\label{eq:mastu-equilibrium}
\end{equation}
with $\boldsymbol\varepsilon(\mathbf u)=\tfrac12\bigl(\nabla\mathbf u+(\nabla\mathbf u)^\top\bigr)$,
$\mathbb C$ the piecewise isotropic stiffness tensor, $\alpha$ the thermal
expansion coefficient and $T_{\mathrm{sf}}$ the stress-free temperature.
Displacement is fixed at the four support-leg footprints. Stainless steel,
copper and graphite have Young's moduli $(210,125,11)$~GPa, densities
$(7800,8960,2260)$~kg\,m$^{-3}$ and a common Poisson ratio of $0.3$
\Citep{Ellis_2025}. Merged material
interfaces share nodes, enforcing displacement continuity while strain jumps
across the stiffness discontinuity. We predict nodal $(u_x,u_y,u_z)$ in metres;
element-wise stresses are not targets.

\paragraph{Parameter scan.}
The corpus contains a stochastic parameter sampling study giving $70$ simulations on the same device.
Each solve uses $1{,}120$ MPI ranks and samples the two parameters of
\Cref{tab:app_mastu_params} over a rectangle. Each sub-application writes its own partitioned Nemesis/Exodus
output. The sampled inputs are read from the corresponding study's parameter
table, with the study identifier and sub-application index cross-checked
against the row; the output files themselves carry no inputs. The product
$\alpha T_w$, which sets the thermal-loading scale, spans a factor of $3.36$
($5.32\times10^{-3}$ to $1.79\times10^{-2}$).

\begin{table}[htbp]
\centering
\footnotesize
\setlength{\tabcolsep}{4pt}
\begin{tabular}{llllll}
\toprule
\textbf{Parameter} & \textbf{Symbol} & \textbf{Unit} & \textbf{Realised range} & \textbf{Mean} & \textbf{Role} \\
\midrule
Thermal expansion coefficient & $\alpha$ & $10^{-5}$~K$^{-1}$ & $1.015$--$1.982$ & $1.50$ & eigenstrain scale \\
Wall temperature & $T_w$ & K & $508.6$--$996.2$ & $750.1$ & thermal boundary condition \\
\bottomrule
\end{tabular}
\caption{MAST-U: the two conditioning parameters over the pooled $70$ solves.
Both are min--max normalised per variable using the pooled corpus ranges before
entering the conditioning module.}
\label{tab:app_mastu_params}
\end{table}

\paragraph{Mesh.}
All $70$ solves share one mesh with $29{,}183{,}754$ nodes and
$134{,}609{,}743$ four-node tetrahedra; cache construction checks each new
solve's coordinates against the reference mesh node for node. Comparison with
the single archived solve of \Citet{Ellis_2025} confirms exactly matching
coordinates, per-node material membership and lumped control volumes. The
$1{,}120$-rank decomposition does, however, renumber nodes relative to that
archived solve, so the two must be matched by coordinate rather than global ID.
The mesh is strongly multiscale: merging thousands of parts leaves sliver
elements, with edge lengths from $1.4\times10^{-6}$ to $4.7\times10^{-2}$~m
around a $3$~mm median. Node count is therefore a poor proxy for volume: the
lumped nodal volume spans eleven orders of magnitude. Steel, copper and
graphite occupy $64.4\%$, $22.3\%$ and $13.3\%$ of the $13.11$~m$^3$ of solid
and meet at $1{,}290{,}161$ shared interface nodes ($1{,}164{,}457$
steel/copper, $125{,}704$ steel/graphite, none copper/graphite).

\paragraph{Split.}
The split contains $60$ training solves and $10$ named test solves.  All ten lie inside the training parameter ranges, so the task measures interpolation in $\boldsymbol\mu$. 

\paragraph{Learning setup.}
The full corpus is
$\mathbf F\in\mathbb R^{70\times3\times29{,}183{,}754}$, carrying
$(u_x,u_y,u_z)$ in metres, with $c_{\mathrm{in}}=0$ and
$\boldsymbol\mu\in\mathbb R^{70\times2}$. Training and the ten-case evaluation uses the entire 29.2M nodes with material specific routing for the mesh transfer. The shared mesh allows positional encodings and
spline stencils to be computed once and reused. Native-mesh inference for \Cref{fig:mastu-displacement} uses all $29{,}183{,}754$ nodes as both sources and queries.

\paragraph{Data statistics.}
Over all $70$ solves and all native nodes, $u_x\in[-3.70,3.77]$~cm
and $u_y\in[-3.71,3.73]$~cm are approximately centred (means
$-0.004$ and $0.003$~cm, standard deviations $0.92$ and
$0.98$~cm), while $u_z\in[-0.18,5.43]$~cm has mean
$2.10$~cm and standard deviation $0.92$~cm: heating expands the
machine upward and outward from its supports. With equal weight per node,
$u_z$ carries $74.3\%$ of $\|\mathbf u\|_2^2$, compared with
$12.1\%$ and $13.7\%$ for $u_x$ and $u_y$, so the joint
relative $L_2$ is weighted towards the vertical. Displacement magnitude peaks
at $5.89$~cm and the case-mean magnitude spans $[1.52,3.54]$~cm,
a factor of $2.3$. Displacement is pinned to zero at the supports;
predicting zero scores exactly $100\%$ under \Cref{eq:app_relL2}.

\paragraph{Model and optimisation.}
The model used for \Cref{fig:mastu-displacement} is FiLM-conditioned
L-SplineFNO with a $120\times120\times173$ latent grid, four dense FNO layers
of width $32$, $12$ Fourier modes, spline scatter scales $(1,2,4)$ and
material-specific lifting, projection and spline kernels around a shared
spectral backbone; $28{,}359{,}627$ trainable parameters. A local bypass carries
the conditioned lifted node features directly to the decoder, where they are
concatenated with the decoded grid features and passed through a
material-specific GELU readout with hidden width $64$. Coordinates are
min--max normalised per axis and encoded with integrated positional encoding
at $16$ frequencies; displacement components are standardised separately.
We train for $500$ epochs with Adam, an initial learning rate of $10^{-3}$,
cosine decay to zero, gradient clipping at $1.0$, the relative $L_2$ loss on
standardised targets and deterministic kernels. The training takes $18.7$~h on one Nvidia H100 with $96$~GB.

\paragraph{Native-mesh reconstruction.}
\Cref{fig:mastu-displacement} shows displacement magnitude and absolute vector
error for the held-out solve with
$(\alpha,T_w)=(1.43283\times10^{-5}~\mathrm{K}^{-1},863.979~\mathrm{K})$,
on the undeformed device with the front quarter removed. All native source
nodes contribute to a single global encoding and FNO evaluation, and all
$29{,}183{,}754$ query nodes are decoded in chunks. This full-mesh evaluation
reaches a relative vector $L_2$ error of $3.40\%$ on this case, compared with
$1.60\%$ for its saved prediction on the sampled nodes. The ten-case mean above
therefore describes the sampled-node evaluation; the native-mesh result
additionally changes the source set used to construct the latent field.
The figure uses millimetres and caps the error colour scale at the $99$th
percentile, $2.87$~mm; the maximum nodal vector error is $13.02$~mm.

\newpage
\section{Models}
\label{app:models}

This appendix describes the six baselines against which the spline-gridded
operators are compared. They fall into three families: \emph{point-set
transformers} that never leave the mesh (Transolver, GNOT), \emph{latent-token
transformers} that compress the mesh into a fixed-size bottleneck (UPT, MSPT),
and \emph{gridded neural operators} that transfer the mesh onto a uniform
latent grid so that an FNO can be applied (NUNO, GINO). The last family is the
one our method belongs to, and the transfer operators used by NUNO and GINO ---
nearest-cell scatter and a radius-graph kernel integral, respectively --- are
exactly the components the spline encode/decode pair replaces.

All six consume the same per-node tensor produced by the pipeline,
\begin{equation}
\mathbf{x}_i \;=\; \big[\,\omega_i \;\big\|\; u_0(\omega_i)\,\big] \in \mathbb{R}^{d + c_{\mathrm{in}}},
\qquad i = 1,\dots,N ,
\label{eq:app-model-input}
\end{equation}
where $\omega_i \in \Omega$ is the coordinate of mesh node $i$ and
$u_0(\omega_i)$ the input field features at that node, and all six emit
$\mathcal{Y} \in \mathbb{R}^{N \times c_{\mathrm{out}}}$ at the same nodes.
In all six benchmarks there is no input field, and the tensor is the
positional encoding of \Cref{eq:pipeline-lift},
$\mathbf{x}_i=[\bar\omega_i\,\|\,\varphi(\bar\omega_i)]\in\mathbb{R}^{3+96}$.
Conditioning on $\boldsymbol{\mu} \in \mathcal{P}$ enters through the shared conditioning
module (FiLM, concatenation, or cross-attention) applied to the encoded
features $\varphi(\bar\omega_i)$ before each model's input embedding, with the
leading $d$ coordinate channels passed through unmodulated, except where a model has its own native mechanism --- GINO's adaptive
instance normalisation, noted below. Throughout, $c$ denotes the model width,
$\mathrm{LN}$ layer normalisation, and
$\mathrm{MHA}(Q,K,V) = \mathrm{Softmax}\!\big(QK^{\top}/\sqrt{c_h}\big)V$
standard softmax attention with head dimension $c_h$; $\mathrm{MHSA}(X)$ is the
self-attention special case $\mathrm{MHA}(X,X,X)$.

\begin{table}[htbp]
\centering
\footnotesize
\setlength{\tabcolsep}{3pt}
\begin{tabular}{lllll}
\toprule
\textbf{Model} & \textbf{Encoder} & \textbf{Processor} & \textbf{Decoder} & \textbf{Attention Cost} \\
\midrule
Transolver & feature slicing & softmax MHSA & deslicing & $O(N n_{s} c + n_{s}^{2} c)$ \\
GNOT       & trunk/branch MLPs     & linear cross + self-attn. & MLP & $O(Nc^2)$ \\
UPT        & Perceiver pooling & softmax MHSA & Perceiver query & $O(N n_{s} c + n_{s}^{2} c)$ \\
MSPT       & patch pooling & chunked MHSA + supernodes & MLP & $O(N^2/n_{p} + N n_{p}n_{q})c$ \\
NUNO       & K-D tree + scatter & FNO on subdomain grid & trig. interpolation & --- \\
GINO       & radius-graph GNO      & FNO on latent grid & radius-graph GNO & --- \\
\bottomrule
\end{tabular}
\caption{The six baselines by encode--process--decode role. $N$ is the number
of mesh nodes, $n_{s}$ the number of slices or latent tokens, $n_{p}$ the number of
patches and $n_{q}$ the supernodes per patch, $c$ the model width. The two gridded
operators have no attention term; their cost is dominated by the FFT on the
latent grid, $O(G \log G)$ for $G$ grid points.}
\label{tab:app-model-roles}
\end{table}

\subsection{Transolver}
\label{app:model-transolver}

Transolver \citep{wu2024transolverfasttransformersolver} is a transformer solver for PDEs on general
geometries. Applying a standard transformer to a discretised 3D mesh is
quadratic in the node count and therefore infeasible at the $10^4$--$10^6$ nodes
these benchmarks carry. Transolver's answer is to stop attending between nodes
altogether: it learns a soft partition of the nodes into a small number $n_{s} \ll N$
of \emph{slices}, attends only among the $n_{s}$ slice tokens, and broadcasts the
result back. Crucially the partition is a function of the node \emph{features},
not of spatial proximity, so nodes in physically similar states --- a boundary
layer distributed across a wing, say --- are grouped into the same token
regardless of where they sit in $\Omega$. This is the \emph{Physics-Attention}
mechanism, and it is what lets the model be geometry-general without a graph.
Transolver++ \citep{luo2025transolveraccurateneuralsolver} scales the same idea to million-node
meshes.

The mesh is treated as an unordered point set. There is no adjacency structure,
no grid, and no requirement that nodes be in correspondence across samples, so
the model is permutation-equivariant in the node index and consumes per-sample
geometries without modification.

\paragraph{Encoding.}
Node features are lifted, then assigned to slices by a softmax with a
per-node adaptive temperature $s_i$, which controls how sharply a node
commits to a single slice:
\begin{equation}
h_i^{(0)} = \mathrm{MLP}_{\mathrm{lift}}(x_i) \in \mathbb{R}^{c},
\qquad
s_i = \mathrm{Softplus}\big(\mathrm{MLP}_{s}(h_i)\big) + \varepsilon ,
\end{equation}
\begin{equation}
w_{ij}
=
\frac{\exp\!\big((W_s h_i)_j / s_i\big)}
 {\sum_{j'=1}^{n_{s}} \exp\!\big((W_s h_i)_{j'} / s_i\big)} ,
\qquad
W_s \in \mathbb{R}^{M \times c},
\qquad
\sum_{j=1}^{n_{s}} w_{ij} = 1 .
\label{eq:transolver-slice}
\end{equation}
The $n_{s}$ slice tokens are the weighted means of their member nodes,
\begin{equation}
z_j
=
\frac{\sum_{i=1}^{N} w_{ij}\, h_i}{\sum_{i=1}^{N} w_{ij} + \varepsilon}
\in \mathbb{R}^{c} .
\label{eq:transolver-slicetoken}
\end{equation}

\paragraph{Processing.}
Full softmax self-attention is applied among the $n_{s}$ tokens only, which is
where the $N^2$ cost is avoided:
\begin{equation}
Z' = Z + \mathrm{MHSA}\big(\mathrm{LN}(Z)\big),
\qquad
Z = [z_1,\dots,z_M]^{\top} \in \mathbb{R}^{M \times c} .
\end{equation}

\paragraph{Decoding.}
Tokens are scattered back to nodes with the same assignment weights
(``deslicing''), followed by a pointwise feed-forward block; $L$ such blocks are
stacked and a final projection produces the output field:
\begin{equation}
h_i \leftarrow h_i + \sum_{j=1}^{n_{s}} w_{ij}\, z'_j ,
\qquad
h_i \leftarrow h_i + \mathrm{FFN}\big(\mathrm{LN}(h_i)\big) ,
\qquad
\mathcal{Y}_i = \mathrm{MLP}_{\mathrm{proj}}\big(h_i^{(L)}\big) .
\label{eq:transolver-deslice}
\end{equation}
Each block costs $O(N n_{s} c + n_{s}^{2} c)$, linear in $N$. The LinearNO variant used in
some of our sweeps replaces the token-level softmax attention with separate
key and query slicing weights and a spectrally-normalised value projection,
$z_j = \sum_i w^{k}_{ij} (W_v h_i) / \sum_i w^{k}_{ij}$ and
$\hat h_i = \sum_j w^{q}_{ij} \mathrm{LN}(z_j)$, removing the $n_{s}^{2}$ term.

\subsection{General Neural Operator Transformer (GNOT)}
\label{app:model-gnot}

GNOT \citep{hao2023gnotgeneralneuraloperator} is a transformer built for the case where the operator
consumes \emph{heterogeneous} inputs --- a geometry, one or more input
functions, and a set of scalar parameters --- which is precisely the conditional
setting of this paper. It keeps a \emph{trunk} stream of query points and one
\emph{branch} stream per input function, with separate key/value projections for
each branch, so streams of different physical meaning are never forced into a
shared embedding. Two design choices make it usable at mesh scale: all
attention is normalised \emph{linear} attention, $O(Nc^2)$ rather than
$O(N^2c)$, and the feed-forward blocks are mixtures of experts gated by the
point's coordinates, which lets different regions of the domain (boundary layer
against far field) be handled by different experts --- a soft, learned analogue
of domain decomposition.

Inputs are point clouds. Query points are arbitrary and need not coincide with
the points at which the input functions are sampled, so the model imposes no
mesh correspondence across samples.

\paragraph{Encoding.}
The trunk and each of the $n_{b}$ branches are embedded by residual MLPs,
\begin{equation}
h_i^{(0)} = \mathrm{MLP}_{\mathrm{trunk}}(x_i) \in \mathbb{R}^{c},
\qquad
y^{(s)}_{t} = \mathrm{MLP}^{(s)}_{\mathrm{branch}}\big(b^{(s)}_t\big) \in \mathbb{R}^{c},
\quad s = 1,\dots,S .
\end{equation}
In our single-input-function configuration $n_{b} = 1$ and the branch is fed the
same node tensor as the trunk.

\paragraph{Processing.}
Write $\varsigma$ for the softmax taken over the \emph{feature} dimension. The
normalised linear attention kernel is
\begin{equation}
\mathrm{LinAttn}(Q,K,V)_i
=
\frac{\varsigma(q_i)^{\top} \Big(\sum_{t} \varsigma(k_t)\, v_t^{\top}\Big)}
 {\varsigma(q_i)^{\top} \sum_{t} \varsigma(k_t)} ,
\label{eq:gnot-linattn}
\end{equation}
which never forms the $N \times N$ attention matrix: the context
$\sum_t \varsigma(k_t) v_t^{\top} \in \mathbb{R}^{c_h \times c_h}$ is accumulated
first, giving $O(Nc^2)$ cost. Cross-attention sums the per-branch
contributions, and the geometry gate selects among $n_{e}$ experts,
\begin{equation}
\mathrm{CrossAttn}(h, \{y^{(s)}\})_i
=
\sum_{s=1}^{n_{b}} \mathrm{LinAttn}\big(W_q h,\, W^{(s)}_k y^{(s)},\, W^{(s)}_v y^{(s)}\big)_i ,
\end{equation}
\begin{equation}
g_i = \mathrm{Softmax}\big(\mathrm{MLP}_{\mathrm{gate}}(\omega_i)\big) \in \Delta^{n_{e}-1},
\qquad
\mathrm{MoE}(h_i) = \sum_{e=1}^{n_{e}} g_{ie}\, \mathrm{FFN}_e(h_i) .
\label{eq:gnot-moe}
\end{equation}
The gate sees the coordinates alone, so its specialisation is spatial while the
attention remains feature-driven. One GNOT layer applies both attention types,
each followed by a gated mixture:
\begin{equation}
\begin{aligned}
h &\leftarrow h + \mathrm{CrossAttn}\big(\mathrm{LN}_1(h), \{\mathrm{LN}^{(s)}(y^{(s)})\}\big) ,
&\qquad
h &\leftarrow h + \mathrm{LN}_3\big(\mathrm{MoE}_1(h)\big) , \\
h &\leftarrow h + \mathrm{LinSelfAttn}\big(\mathrm{LN}_4(h)\big) ,
&\qquad
h &\leftarrow h + \mathrm{LN}_5\big(\mathrm{MoE}_2(h)\big) .
\end{aligned}
\end{equation}

\paragraph{Decoding.}
A residual MLP maps the final trunk tokens to the output channels,
$\mathcal{Y}_i = \mathrm{MLP}_{\mathrm{out}}\big(h_i^{(L)}\big)$. Because the
trunk is evaluated pointwise, the operator can be queried at coordinates that
appear in no training mesh.

\subsection{Universal Physics Transformer (UPT)}
\label{app:model-upt}

The Universal Physics Transformer \citep{alkin2025universalphysicstransformersframework} is an
encode--process--decode architecture whose defining property is that the
processor never sees the discretisation. A Perceiver-style pooling stage
\citep{jaegle2022perceiveriogeneralarchitecture} compresses a variable-length point cloud into a
\emph{fixed} number $n_{s}$ of latent tokens; the physics is modelled entirely
among those tokens; and a decoder cross-attends from arbitrary query positions
back into the latent set. Cost in the processor is therefore $O(n_{s}^{2} c)$ and
independent of $N$ --- the model scales with the complexity of the solution,
not the density of the mesh --- and the encoder and decoder discretisations are
decoupled, so the field can be read out on a mesh other than the one it was
encoded from.

Meshes are handled as raw point clouds: no padding, no graph construction, no
correspondence between samples. Our steady-state configuration drops the latent
time-propagation stage of the original, which advances the latent tokens
between time steps; the encoder here pools directly over all nodes rather than
over the message-passed supernodes of the reference implementation.

\paragraph{Encoding.}
Coordinates are embedded with geometrically-spaced sinusoids,
$\mathrm{PE}(\omega) = \big[\sin(\omega^{(\ell)}\!f_r),\, \cos(\omega^{(\ell)}\!f_r)\big]_{\ell, r}$
with $f_r = 10000^{-r/F}$, and added to the lifted features:
\begin{equation}
h_i = \mathrm{MLP}_{\mathrm{lift}}(x_i) + W_p\,\mathrm{PE}(\omega_i) \in \mathbb{R}^{c} .
\end{equation}
$n_{s}$ learnable query tokens $Q \in \mathbb{R}^{n_{s} \times c}$ pool the mesh:
\begin{equation}
Z^{(0)}
=
Q + \mathrm{MHA}\big(\mathrm{LN}(Q),\, \mathrm{LN}(H),\, \mathrm{LN}(H)\big)
\in \mathbb{R}^{M \times c},
\qquad H = [h_1,\dots,h_N]^{\top} .
\label{eq:upt-encode}
\end{equation}

\paragraph{Processing.}
$L$ pre-norm transformer blocks act on the latent set alone:
\begin{equation}
Z \leftarrow Z + \mathrm{MHSA}\big(\mathrm{LN}(Z)\big),
\qquad
Z \leftarrow Z + \mathrm{FFN}\big(\mathrm{LN}(Z)\big) .
\end{equation}

\paragraph{Decoding.}
Any query position $\omega_q$ forms a query token from the same lifting and
positional embedding, and cross-attends into the processed latents:
\begin{equation}
\mathcal{Y}(\omega_q)
=
\mathrm{MLP}_{\mathrm{proj}}\Big(
h_q + \mathrm{MHA}\big(\mathrm{LN}(h_q),\, \mathrm{LN}(Z^{(L)}),\, \mathrm{LN}(Z^{(L)})\big)
\Big) .
\label{eq:upt-decode}
\end{equation}

\subsection{Multi-Scale Patch Transformer (MSPT)}
\label{app:model-mspt}

MSPT \citep{curvo2026msptefficientlargescalephysical} attacks the same quadratic-attention problem from the
opposite direction to UPT: instead of compressing the whole mesh into a global
bottleneck, it partitions the nodes into $n_{p}$ patches, runs full attention
\emph{inside} each patch, and couples the patches through a small set of
\emph{supernodes} --- $n_{q}$ pooled summaries per patch --- that every patch also
attends to. Local detail is therefore resolved at full resolution while global
communication is carried by $n_{p}n_{q}$ tokens, giving the two scales the name refers
to. Supernodes persist across blocks by residual accumulation, so the global
state is refined depth-wise rather than recomputed.

Partitioning is either sequential over the node ordering or, for spatially
coherent patches, a ball-tree partition of the coordinates; the latter is what
makes patches correspond to compact regions of $\Omega$ on an unstructured
mesh. Rotary positional embeddings may be applied within a patch sequence.
Node counts that are not a multiple of $n_{p}$ are zero-padded and the padding is
stripped on output.

\paragraph{Encoding.}
Tokens are formed from coordinates (optionally replaced by a
\emph{unified position} embedding: distances to a fixed $\mathrm{ref}^d$
reference lattice, which gives every sample a common frame) concatenated with
field features, and permuted into patches $\{C_v\}_{v=1}^{n_{p}}$ by the partition
$\pi$:
\begin{equation}
h_i = \mathrm{MLP}_{\mathrm{pre}}\big([\,\phi(\omega_i)\,\|\,u_0(\omega_i)\,]\big) + p ,
\qquad
\{C_v\}_{v=1}^{n_{p}} = \pi\big(\{\omega_i\}\big),
\qquad |C_v| = N/V .
\end{equation}

\paragraph{Processing.}
Each block pools $n_{q}$ supernodes per patch and adds the previous block's,
\begin{equation}
s^{(\ell)}_{v,q}
=
\mathrm{pool}_q\big(\{h^{(\ell)}_i\}_{i \in C_v}\big) + s^{(\ell-1)}_{v,q} ,
\qquad
S^{(\ell)} = \big[s^{(\ell)}_{v,q}\big] \in \mathbb{R}^{VQ \times c} ,
\label{eq:mspt-supernode}
\end{equation}
with $\mathrm{pool}$ a mean, top-$n_{q}$ max, or learned linear pooling. Attention
then runs on each patch \emph{concatenated with the full supernode set}, so
every node sees a summary of every other patch:
\begin{equation}
\tilde H_v = \big[\,H_{C_v} \;\big\|\; S^{(\ell)}\,\big] \in \mathbb{R}^{(N/V + VQ) \times c},
\qquad
\tilde H_v \leftarrow \tilde H_v + \mathrm{MHSA}\big(\mathrm{LN}(\tilde H_v)\big),
\end{equation}
\begin{equation}
\tilde H_v \leftarrow \tilde H_v + \mathrm{FFN}\big(\mathrm{LN}(\tilde H_v)\big),
\qquad
H_{C_v} = \tilde H_v[1{:}N/V],
\qquad
S^{(\ell)} \leftarrow \frac{1}{V}\sum_{v=1}^{V} \tilde H_v[-VQ{:}] .
\end{equation}
The cost is $O\big(n_{p}(N/n_{p} + n_{p}n_{q})^2 c\big) = O\big((N^2/n_{p} + N n_{p} n_{q})c\big)$, minimised
at $n_{p} \sim \sqrt{N/n_{q}}$.

\paragraph{Decoding.}
The final block projects to the output channels and the patch permutation is
inverted, $\mathcal{Y} = \pi^{-1}\big(W_{\mathrm{out}}\,\mathrm{LN}(H^{(L)})\big)$.

\subsection{Non-Uniform Neural Operator (NUNO)}
\label{app:model-nuno}

NUNO \citep{liu2023nunogeneralframeworklearningnuno} extends grid-based operator learning to non-uniform
data. Fast neural operators such as the FNO \citep{li2021fourierneuraloperatorparametric} rest on the
FFT, which requires a uniform grid; applying one to a tetrahedral mesh means
interpolating globally onto a single grid, and the resulting error is set by the
worst-resolved region of the domain. NUNO's remedy is decomposition: a K-D tree
recursively splits $\Omega$ into $n_{c}$ subdomains of roughly equal node count, so
that dense regions receive small cells and sparse regions large ones, and the
local interpolation error is bounded within each subdomain rather than globally.
The subdomain grids are then processed by an FNO, retaining FFT speed.

\paragraph{Encoding.}
The tree splits each node set $P$ at the median along its widest axis,
\begin{equation}
\begin{aligned}
\ell^{*} &= \operatorname*{arg\,max}_{\ell \in \{1,\dots,d\}}
\Big(\max_{i \in P} \omega^{(\ell)}_i - \min_{i \in P} \omega^{(\ell)}_i\Big), \\
P &\mapsto
\big\{ i \in P : \omega^{(\ell^{*})}_i \le m_{\ell^{*}} \big\}
\;\cup\;
\big\{ i \in P : \omega^{(\ell^{*})}_i > m_{\ell^{*}} \big\},
\end{aligned}
\label{eq:nuno-kdtree}
\end{equation}
recursed until $n_{c}$ leaves remain. Lifted features are transferred to the
uniform grid $\mathcal{G}$ of shape $S_1 \times \cdots \times S_d$ by
nearest-cell scatter,
\begin{equation}
\mathcal{X}_{\mathcal{G}}[\mathbf{g}]
=
\sum_{i \,:\, \gamma(\omega_i) = g} \mathcal{X}_i ,
\qquad
\gamma(\omega_i)^{(\ell)}
=
\Big\lfloor
\frac{\omega^{(\ell)}_i - \min_j \omega^{(\ell)}_j}
 {\max_j \omega^{(\ell)}_j - \min_j \omega^{(\ell)}_j}\,(S_\ell - 1)
\Big\rceil ,
\label{eq:nuno-scatter}
\end{equation}
after which the normalised grid coordinates are concatenated as channels and
lifted. \Cref{eq:nuno-scatter} is the piecewise-constant limit of the
transfer operator our spline encoder generalises: it is order-zero, so it
assigns a node's whole contribution to one cell and provides no gradient with
respect to $\omega_i$.

\paragraph{Processing.}
Standard Fourier layers act on the grid, truncated to $|k_\ell| \le m_\ell$
modes per axis:
\begin{equation}
v^{(\ell+1)}
=
\sigma\Big(
W v^{(\ell)}
+
\mathcal{F}^{-1}\big(R^{(\ell)} \cdot \mathcal{F}(v^{(\ell)})\big)
\Big) .
\label{eq:nuno-fno}
\end{equation}

\paragraph{Decoding.}
A pointwise MLP expands the grid features into $n_{c}$ per-subdomain copies, and
each subdomain's nodes are read out by evaluating the truncated Fourier series
directly at their true coordinates, rather than by interpolating off the grid a
second time:
\begin{equation}
\mathcal{Y}_i
=
\mathrm{MLP}_{\mathrm{out}}\!\left(
\sum_{\|k\|_\infty \le m} \hat v_k \, e^{\,2\pi \mathrm{i} \langle k,\, \omega_i \rangle}
\right),
\qquad
\hat v_k = \sum_{\mathbf{g} \in \mathcal{G}} v[\mathbf{g}]\, e^{-2\pi \mathrm{i} \langle k,\, \xi_{\mathbf{g}} \rangle} .
\label{eq:nuno-decode}
\end{equation}
This is the same non-uniform forward/inverse transform pair used by Geo-FNO
\citep{GeoFNO_Li2024}. In three dimensions our implementation falls back to
trilinear \texttt{grid\_sample} for the read-out, since the explicit
$O(N \prod_\ell 2m_\ell)$ basis becomes the dominant memory cost.

\subsection{Geometry-Informed Neural Operator (GINO)}
\label{app:model-gino}

GINO \citep{li2023geometryinformedneuraloperatorlargescale} is the other
gridded baseline, and the closest architectural comparison to our method: it is
an FNO on a uniform latent grid, bracketed by learned transfer operators that
move features between the mesh and that grid. Where we use a tensor-product
spline scatter and its analytic adjoint, GINO uses a Graph Neural Operator on
both sides --- a radius-graph kernel integral that aggregates, for each latent
grid point, the mesh nodes within a ball of radius $r$, and then the reverse.
The encoder and decoder are therefore Monte-Carlo/Riemann approximations of a
continuous kernel integral, learned as a neural network of the relative
displacement.

The radius graph is built between the mesh and the latent grid, so the mesh
needs no fixed connectivity and per-sample geometries are admitted. Two
practical points follow from the construction. First, $r$ is measured in the
normalised $[0,1]^d$ coordinates of the latent grid and must be at least the
largest grid spacing, or latent points collect no neighbours and their features
are identically zero. Second, because the neighbourhood structure differs per
sample, the reference implementation batches only when the geometry is shared,
and our wrapper therefore loops over samples.

\paragraph{Encoding.}
For each latent grid point $\xi_{\mathbf{g}}$, $\mathbf{g} \in \mathcal{G}$, with neighbourhood
$\mathcal{N}_r(\xi_{\mathbf{g}}) = \{ i : \|\omega_i - \xi_{\mathbf{g}}\|_2 \le r \}$,
\begin{equation}
\mathcal{X}_{\mathcal{G}}[\mathbf{g}]
=
\frac{1}{|\mathcal{N}_r(\xi_{\mathbf{g}})|}
\sum_{i \in \mathcal{N}_r(\xi_{\mathbf{g}})}
\kappa_{\mathrm{G}}\big(\xi_{\mathbf{g}},\, \omega_i,\, \mathcal{X}_i\big)\, \mathcal{X}_i
\;\approx\;
\int_{B_r(\xi_{\mathbf{g}})} \kappa_{\mathrm{G}}(\xi_{\mathbf{g}}, \omega)\, u(\omega)\, \mathrm{d}\omega ,
\label{eq:gino-encode}
\end{equation}
with $\kappa_{\mathrm{G}}$ an MLP of the sinusoidally-embedded relative displacement
$\xi_{\mathbf{g}} - \omega_i$.

\paragraph{Processing.}
An FNO acts on the latent grid as in \Cref{eq:nuno-fno}. Conditioning is
native here: when $k > 0$ the normalisation layers are adaptive instance norms
driven by $\boldsymbol{\mu}$,
\begin{equation}
\mathrm{AdaIN}(v;\boldsymbol{\mu})
=
\gamma_{\mathrm{G}}\big(e(\boldsymbol{\mu})\big)\,
\frac{v - \mathbb{E}[v]}{\sqrt{\mathrm{Var}[v] + \varepsilon}}
+
\beta_{\mathrm{G}}\big(e(\boldsymbol{\mu})\big) ,
\qquad
e(\boldsymbol{\mu}) = \big[\sin(f_r \boldsymbol{\mu}),\, \cos(f_r \boldsymbol{\mu})\big]_{r=1}^{F} ,
\label{eq:gino-adain}
\end{equation}
where $\gamma_{\mathrm{G}}, \beta_{\mathrm{G}}$ are MLPs. Their width scales with $k$, so the parameter
count of GINO grows with the conditioning dimension --- at $k = 40$ the norm
MLPs alone add roughly $1.5$M parameters, which is why $F$
(\texttt{ada\_in\_frequencies}) is kept small in our configurations.

\paragraph{Decoding.}
A second GNO integrates back from the grid to the query nodes,
\begin{equation}
\mathcal{Y}_i
=
\frac{1}{|\mathcal{N}_r(\omega_i)|}
\sum_{g \in \mathcal{N}_r(\omega_i)}
\kappa'_{\mathrm{G}}\big(\omega_i,\, \xi_{\mathbf{g}}\big)\, \mathcal{Y}_{\mathcal{G}}[\mathbf{g}] ,
\qquad
\mathcal{N}_r(\omega_i) = \{ g : \|\xi_{\mathbf{g}} - \omega_i\|_2 \le r \} .
\label{eq:gino-decode}
\end{equation}
Like the spline decoder of \Cref{eq:pipeline-output}, \Cref{eq:gino-decode} is
linear in the grid values: once trained, the prediction at node $i$ is a
combination of the $|\mathcal{N}_r(\omega_i)|$ grid values inside the ball, with
weights that depend only on the positions of the node and the grid points, never on
the field itself. The ball $\mathcal{N}_r$ plays the role our spline stencil plays,
with two differences: its size is set by the radius $r$ rather than by the spline
degree, and its weights $\kappa'_{\mathrm{G}}$ are learned without any constraint,
whereas the analytic spline weights form a partition of unity.

\newpage
\section{Conditioning Methods}
\label{app:conditioning_methods}

All conditioning methods receive the physical parameter vector $\boldsymbol{\mu} \in \mathbb{R}^k$ and act at the same place in the pipeline: on the per-node encoded coordinate features $\mathbf{f}=\varphi(\bar\omega)\in\mathbb{R}^{B_{\mathrm{batch}}\times N\times c_f}$, $c_f=96$, before the backbone's input lift (\Cref{eq:pipeline-lift}). The three normalised coordinates $\bar\omega$ are excluded from conditioning for every method and every backbone; the conditioned features $\tilde{\mathbf f}$ are concatenated with $\bar\omega$ and lifted, $\mathbf h_i=P_{\mathrm{in}}[\tilde{\mathbf f}_i;\bar\omega_i]$. No method acts on the width-$c$ lifted features or on intermediate backbone states in any reported run.

\subsection{FiLM}

Feature-wise Linear Modulation (FiLM) \citep{perez2018film} is a lightweight conditioning strategy that affinely transforms each channel of the field representation based on the conditioning parameters. An MLP $H_\mu : \mathbb{R}^k \to \mathbb{R}^{2c_f}$, composed of the embedding $E_\mu$ and the two heads $H_\gamma$, $H_\beta$ of \Cref{sec:lsfno-pipeline}, maps $\boldsymbol{\mu}$ to per-channel scale and shift parameters:

\begin{equation}
(\boldsymbol{\gamma}_\mu, \boldsymbol{\beta}_\mu) = H_\mu(\boldsymbol{\mu}), \qquad \boldsymbol{\gamma}_\mu, \boldsymbol{\beta}_\mu \in \mathbb{R}^{c_f}
\end{equation}

The conditioned field is then given by an element-wise affine transformation applied uniformly across all spatial locations:

\begin{equation}
\tilde{\mathbf f} = \boldsymbol{\gamma}_\mu \odot \mathbf f + \boldsymbol{\beta}_\mu
\end{equation}

where $\odot$ denotes element-wise multiplication broadcast over the spatial dimensions. The FiLM layer preserves the channel count $c_f$ and does not alter the spatial structure of the representation. The scale and shift heads have zero-initialised weights and biases $\mathbf{1}$ and $\mathbf{0}$, so $\boldsymbol{\gamma}_\mu = \mathbf{1}$ and $\boldsymbol{\beta}_\mu = \mathbf{0}$ exactly at initialisation and the unconditioned backbone behaviour is recovered.

\subsection{Concatenation}

The simplest conditioning strategy directly augments the input field by appending the conditioning parameters as additional spatial channels. Each scalar in $\boldsymbol{\mu}$ is broadcast uniformly across all mesh nodes or grid locations and concatenated to the field:

\begin{equation}
\tilde{\mathbf f} = \bigl[\mathbf f \;\|\; \mathbf{1}_\Omega \otimes \boldsymbol{\mu}\bigr] \in \mathbb{R}^{c_f + k}
\end{equation}

where $\mathbf{1}_\Omega \otimes \boldsymbol{\mu}$ denotes the spatial broadcast of $\boldsymbol{\mu}$ to every node or grid point. This increases the number of feature channels from $c_f$ to $c_f + k$, so the lift becomes $P_{\mathrm{in}}:\mathbb{R}^{c_f+k+3}\to\mathbb{R}^{c}$. Concatenation conditioning places no constraint on how the model uses the parameter information, leaving the entire burden of interpreting $\boldsymbol{\mu}$ to the backbone architecture.

\subsection{Cross-Attention}

Cross-attention conditioning enables the model to selectively query physical parameter information at each spatial location through an attention mechanism. The conditioning parameters $\boldsymbol{\mu}$ are first embedded into a set of $n_\tau$ learnable parameter tokens:

\begin{equation}
\mathcal{T} = \operatorname{reshape}\bigl(W_\tau \boldsymbol{\mu}\bigr) + E_\tau \in \mathbb{R}^{n_\tau \times c_\tau}
\end{equation}

where $W_\tau \in \mathbb{R}^{(n_\tau \cdot c_\tau) \times k}$ is a learnable projection and $E_\tau \in \mathbb{R}^{n_\tau \times c_\tau}$ are learnable positional embeddings. The field features at each spatial location serve as queries, attending to the parameter tokens as keys and values:

\begin{align}
Q &= W_Q \, \mathbf f \in \mathbb{R}^{N \times c_\tau}, \\
K = V &= \operatorname{LayerNorm}(\mathcal{T}) \in \mathbb{R}^{n_\tau \times c_\tau}, \\
A &= \operatorname{softmax}\!\left(\frac{Q K^\top}{\sqrt{c_\tau}}\right) V \in \mathbb{R}^{N \times c_\tau}
\end{align}

where $N$ is the number of spatial locations and multi-head attention is used in practice. The output is projected back to the $c_f$ feature channels and added as a residual:

\begin{equation}
\tilde{\mathbf f} = \mathbf f + W_{\mathrm{out}} A
\end{equation}

The output projection $W_{\mathrm{out}}$ is initialised to zero so that the module acts as an identity at the start of training. This preserves the original backbone dynamics and allows the conditioning signal to be introduced gradually.

This section presents tables showcasing full scale ablations across all of the experiments across all models and conditioning methods. We find that the learnable Spline-FNO, on average offers the best performance and compute-efficiency. Similarly FiLM conditioning method across the spatial features of the input performs remarkably well compared to concatenation and cross-attention. 

\begin{table}[htbp]
\centering
\label{tab:heat_sink_metrics}
\resizebox{\linewidth}{!}{%
\begin{tabular}{llrrrrrr}
\hline
\textbf{Model} & \textbf{Cond. Method} & \textbf{\# Parameters} & \makecell{\textbf{Train Time} \\ (hrs:mins)} & \makecell{\textbf{Rel. $L_2$} \% ($\downarrow$)} & \makecell{\textbf{Rel. $L_1$} \%  ($\downarrow$)} & \textbf{$L_2/L_1$} & \makecell{\textbf{$R^2$} ($\uparrow$)} \\ \hline
Transolver & None & 2,433,413 & 0:41 & 0.0589 & 0.0460 & 1.2804 & 0.9989 \\
Transolver & Concatenation & 2,434,437 & 0:41 & 0.0458 & 0.0341 & 1.3449 & 0.9993 \\
Transolver & Cross-attention & 2,464,422 & 0:44 & 0.0438 & 0.0330 & 1.3281 & 0.9994 \\
Transolver & FiLM & 2,450,503 & 0:41 & \textbf{0.0385} & \textbf{0.0280} & 1.3742 & \textbf{0.9995} \\
\hline
UPT & None & 2,252,097 & 0:33 & 0.0907 & 0.0741 & 1.2230 & 0.9972 \\
UPT & Concatenation & 2,252,993 & 0:32 & 0.0423 & 0.0308 & 1.3749 & 0.9994 \\
UPT & Cross-attention & 2,283,106 & 0:37 & 0.0410 & 0.0295 & 1.3884 & \textbf{0.9995} \\
UPT & FiLM & 2,269,187 & 0:34 & \textbf{0.0398} & \textbf{0.0283} & 1.4070 & \textbf{0.9995} \\
\hline
MSPT & None & 2,190,529 & 1:26 & 0.0975 & 0.0819 & 1.1897 & 0.9963 \\
MSPT & Concatenation & 2,192,065 & 1:26 & 0.0399 & 0.0286 & 1.3935 & 0.9995 \\
MSPT & Cross-attention & 2,221,409 & 1:29 & 0.0397 & 0.0287 & 1.3816 & 0.9995 \\
MSPT & FiLM & 2,207,489 & 1:26 & \textbf{0.0368} & \textbf{0.0258} & 1.4274 & \textbf{0.9996} \\
\hline
GNOT & None & 2,192,841 & 1:33 & 0.0674 & 0.0546 & 1.2359 & 0.9985 \\
GNOT & Concatenation & 2,193,609 & 1:33 & 0.0448 & 0.0330 & 1.3549 & 0.9994 \\
GNOT & Cross-attention & 2,223,721 & 1:37 & \textbf{0.0416} & 0.0303 & 1.3723 & 0.9994 \\
GNOT & FiLM & 2,209,801 & 1:34 & 0.0420 & \textbf{0.0301} & 1.3941 & \textbf{0.9995} \\
\hline
NUNO & None & 2,392,455 & 0:13 & 0.0793 & 0.0606 & 1.3096 & 0.9975 \\
NUNO & Concatenation & 2,392,523 & 0:12 & 0.0720 & 0.0534 & 1.3491 & 0.9978 \\
NUNO & Cross-attention & 2,423,335 & 0:16 & 0.0798 & 0.0591 & 1.3497 & 0.9970 \\
NUNO & FiLM & 2,409,415 & 0:13 & \textbf{0.0688} & \textbf{0.0502} & 1.3700 & \textbf{0.9979} \\
\hline
GINO & None & 2,326,267 & 1:06 & 0.0671 & 0.0505 & 1.3267 & 0.9984 \\
GINO & Concatenation & 2,326,775 & 1:06 & 0.0522 & 0.0373 & 1.3996 & \textbf{0.9991} \\
GINO & Cross-attention & 2,357,147 & 1:10 & 0.0553 & 0.0397 & 1.3945 & 0.9989 \\
GINO & FiLM & 2,343,227 & 1:08 & \textbf{0.0518} & \textbf{0.0367} & 1.4090 & \textbf{0.9991} \\
\hline
SplineFNO & None & 2,350,813 & 1:16 & 0.0875 & 0.0753 & 1.1620 & 0.9973 \\
SplineFNO & Concatenation & 2,350,941 & 1:16 & 0.0483 & 0.0378 & 1.2779 & 0.9992 \\
SplineFNO & Cross-attention & 2,381,693 & 1:20 & 0.0386 & 0.0296 & 1.3027 & 0.9995 \\
SplineFNO & FiLM & 2,367,773 & 1:16 & \textbf{0.0343} & \textbf{0.0264} & 1.3027 & \textbf{0.9996} \\
\hline
L-SplineFNO & None & 2,378,815 & 0:42 & 0.0849 & 0.0722 & 1.1773 & 0.9973 \\
L-SplineFNO & Concatenation & 2,378,911 & 0:42 & 0.0340 & 0.0268 & 1.2656 & 0.9996 \\
L-SplineFNO & Cross-attention & 2,409,695 & 0:45 & 0.0335 & 0.0259 & 1.2928 & 0.9996 \\
L-SplineFNO & FiLM & 2,395,775 & 0:42 & \textbf{0.0271} & \textbf{0.0213} & 1.2687 & \textbf{0.9998} \\
\hline
\end{tabular}}%
\caption{Performance metrics of various models and conditioning methods for the Heat Sink (3D, steady-state, thermal) case. Best value per metric within each architecture in bold.}
\end{table}

\begin{table}[htbp]
\centering
\label{tab:bracket_metrics}
\resizebox{\linewidth}{!}{%
\begin{tabular}{llrrrrrr}
\hline
\textbf{Model} & \textbf{Cond. Method} & \textbf{\# Parameters} & \makecell{\textbf{Train Time} \\ (hrs:mins)} & \makecell{\textbf{Rel. $L_2$} \% ($\downarrow$)} & \makecell{\textbf{Rel. $L_1$} \%  ($\downarrow$)} & \textbf{$L_2/L_1$} & \makecell{\textbf{$R^2$} ($\uparrow$)} \\ \hline
Transolver & None & 2,433,413 & 1:47 & 23.6733 & 22.9590 & 1.0311 & 0.6963 \\
Transolver & Concatenation & 2,434,437 & 1:47 & 0.9006 & 0.6526 & 1.3799 & 0.9994 \\
Transolver & Cross-attention & 2,464,422 & 1:57 & 0.8573 & 0.6495 & 1.3198 & 0.9995 \\
Transolver & FiLM & 2,450,503 & 1:49 & \textbf{0.5698} & \textbf{0.4059} & 1.4039 & \textbf{0.9998} \\
\hline
UPT & None & 2,252,097 & 1:28 & 27.0258 & 25.6378 & 1.0541 & 0.6351 \\
UPT & Concatenation & 2,252,993 & 1:28 & 1.9136 & 1.3180 & 1.4519 & 0.9975 \\
UPT & Cross-attention & 2,283,106 & 1:40 & 0.8241 & 0.5924 & 1.3912 & 0.9995 \\
UPT & FiLM & 2,269,187 & 1:32 & \textbf{0.6263} & \textbf{0.4434} & 1.4127 & \textbf{0.9996} \\
\hline
MSPT & None & 2,190,529 & 3:28 & 24.0645 & 23.3410 & 1.0310 & 0.6848 \\
MSPT & Concatenation & 2,192,065 & 3:27 & 0.6546 & 0.4612 & 1.4193 & 0.9997 \\
MSPT & Cross-attention & 2,221,409 & 3:38 & 0.6231 & 0.4599 & 1.3549 & 0.9997 \\
MSPT & FiLM & 2,207,489 & 3:30 & \textbf{0.5117} & \textbf{0.3479} & 1.4708 & \textbf{0.9998} \\
\hline
GNOT & None & 2,192,841 & 4:06 & 22.4876 & 21.7846 & 1.0323 & 0.7350 \\
GNOT & Concatenation & 2,193,609 & 4:06 & 0.6393 & 0.4682 & 1.3656 & 0.9997 \\
GNOT & Cross-attention & 2,223,721 & 4:17 & 0.7305 & 0.5406 & 1.3513 & 0.9996 \\
GNOT & FiLM & 2,209,801 & 4:09 & \textbf{0.4998} & \textbf{0.3529} & 1.4164 & \textbf{0.9998} \\
\hline
NUNO & None & 2,392,455 & 0:40 & 22.5074 & 21.4936 & 1.0472 & 0.7652 \\
NUNO & Concatenation & 2,392,523 & 0:40 & 2.5484 & 1.8541 & 1.3744 & 0.9970 \\
NUNO & Cross-attention & 2,423,335 & 0:50 & \textbf{1.8124} & \textbf{1.2775} & 1.4187 & \textbf{0.9985} \\
NUNO & FiLM & 2,409,415 & 0:42 & 2.0016 & 1.4065 & 1.4231 & 0.9981 \\
\hline
SplineFNO & None & 2,370,430 & 3:16 & 21.5520 & 20.4320 & 1.0548 & 0.7872 \\
SplineFNO & Concatenation & 2,370,498 & 3:21 & 5.6180 & 3.4894 & 1.6100 & 0.9885 \\
SplineFNO & Cross-attention & 2,401,310 & 3:30 & \textbf{5.4254} & \textbf{3.3500} & 1.6195 & \textbf{0.9895} \\
SplineFNO & FiLM & 2,387,390 & 3:22 & 5.4867 & 3.3884 & 1.6192 & 0.9893 \\
\hline
L-SplineFNO & None & 2,370,771 & 0:52 & 23.3930 & 22.7500 & 1.0283 & 0.7366 \\
L-SplineFNO & Concatenation & 2,370,839 & 0:54 & 2.2050 & 1.2575 & 1.7535 & 0.9978 \\
L-SplineFNO & Cross-attention & 2,401,651 & 1:03 & \textbf{1.9297} & \textbf{1.0596} & 1.8212 & \textbf{0.9984} \\
L-SplineFNO & FiLM & 2,387,731 & 0:52 & 1.9710 & 1.0817 & 1.8222 & 0.9983 \\
\hline
\end{tabular}}%
\caption{Performance metrics of various models and conditioning methods for the Bracket (3D, steady-state, elasto-plastic) case. Best value per metric within each architecture in bold.}
\end{table}

\begin{table}[htbp]
\centering
\label{tab:aircraft_metrics}
\resizebox{\linewidth}{!}{%
\begin{tabular}{llrrrrrr}
\hline
\textbf{Model} & \textbf{Cond. Method} & \textbf{\# Parameters} & \makecell{\textbf{Train Time} \\ (hrs:mins)} & \makecell{\textbf{Rel. $L_2$} \% ($\downarrow$)} & \makecell{\textbf{Rel. $L_1$} \%  ($\downarrow$)} & \textbf{$L_2/L_1$} & \makecell{\textbf{$R^2$} ($\uparrow$)} \\ \hline
Transolver & None & 2,434,441 & 1:26 & 64.9073 & 54.6929 & 1.1868 & 0.3622 \\
Transolver & Concatenation & 2,435,209 & 1:27 & \textbf{7.6032} & \textbf{5.7802} & 1.3154 & \textbf{0.9514} \\
Transolver & Cross-attention & 2,464,874 & 1:36 & 14.8217 & 11.5540 & 1.2828 & 0.8201 \\
Transolver & FiLM & 2,451,467 & 1:26 & 9.2876 & 6.9728 & 1.3320 & 0.9273 \\
\hline
UPT & None & 2,252,997 & 0:57 & 62.9624 & 52.7489 & 1.1936 & 0.3524 \\
UPT & Concatenation & 2,253,669 & 0:56 & 14.9982 & 11.6138 & 1.2914 & 0.8207 \\
UPT & Cross-attention & 2,283,430 & 1:07 & 14.7274 & 11.4321 & 1.2882 & 0.8215 \\
UPT & FiLM & 2,270,023 & 0:59 & \textbf{13.9923} & \textbf{10.7691} & 1.2993 & \textbf{0.8446} \\
\hline
MSPT & None & 2,191,301 & 6:13 & 63.2662 & 52.9631 & 1.1945 & 0.4050 \\
MSPT & Concatenation & 2,192,453 & 6:14 & 7.6296 & 5.7169 & 1.3346 & 0.9424 \\
MSPT & Cross-attention & 2,221,605 & 6:22 & 8.5414 & 6.1388 & 1.3914 & 0.9317 \\
MSPT & FiLM & 2,208,197 & 6:15 & \textbf{7.5028} & \textbf{5.7109} & 1.3138 & \textbf{0.9487} \\
\hline
GNOT & None & 2,193,229 & 3:23 & 63.7600 & 53.2095 & 1.1983 & 0.4010 \\
GNOT & Concatenation & 2,193,805 & 3:24 & \textbf{6.8812} & \textbf{5.1839} & 1.3274 & \textbf{0.9578} \\
GNOT & Cross-attention & 2,223,533 & 3:33 & 7.6322 & 5.5376 & 1.3782 & 0.9453 \\
GNOT & FiLM & 2,210,125 & 3:26 & 7.2853 & 5.2563 & 1.3860 & 0.9496 \\
\hline
NUNO & None & 2,407,909 & 0:14 & 67.6140 & 57.0382 & 1.1854 & 0.3014 \\
NUNO & Concatenation & 2,408,005 & 0:14 & \textbf{13.8962} & 10.4103 & 1.3348 & 0.8482 \\
NUNO & Cross-attention & 2,438,213 & 0:24 & 13.9679 & \textbf{10.3816} & 1.3454 & \textbf{0.8508} \\
NUNO & FiLM & 2,424,805 & 0:14 & 14.0354 & 10.4892 & 1.3381 & 0.8466 \\
\hline
SplineFNO & None & 2,341,799 & 0:43 & 70.2940 & 59.4350 & 1.1827 & 0.2824 \\
SplineFNO & Concatenation & 2,341,877 & 0:43 & 21.6660 & 14.1320 & 1.5332 & 0.8196 \\
SplineFNO & Cross-attention & 2,372,103 & 0:52 & 19.5110 & 13.5550 & 1.4395 & 0.8223 \\
SplineFNO & FiLM & 2,358,695 & 0:42 & \textbf{19.0370} & \textbf{13.5230} & 1.4077 & \textbf{0.8290} \\
\hline
L-SplineFNO & None & 2,352,051 & 1:04 & 62.2450 & 51.3970 & 1.2111 & 0.4520 \\
L-SplineFNO & Concatenation & 2,352,147 & 1:04 & 10.4780 & 6.7238 & 1.5583 & 0.9412 \\
L-SplineFNO & Cross-attention & 2,382,355 & 1:13 & \textbf{7.8806} & \textbf{5.5703} & 1.4147 & \textbf{0.9589} \\
L-SplineFNO & FiLM & 2,368,947 & 1:05 & 9.3923 & 6.4468 & 1.4569 & 0.9484 \\
\hline
\end{tabular}}%
\caption{Performance metrics of various models and conditioning methods for the Aircraft (3D, steady-state, compressible flow, 5 fields) case. Best value per metric within each architecture in bold. $R^2$ is the unweighted mean of the per-variable field $R^2$ over the five outputs.}
\end{table}

\newcommand{\tbd}{\textcolor{red}{TBD}}   %

\begin{table}[htbp]
\centering
\label{tab:fds_fire_metrics}
\resizebox{\linewidth}{!}{%
\begin{tabular}{llrrrrrr}
\hline
\textbf{Model} & \textbf{Cond. Method} & \textbf{\# Parameters} & \makecell{\textbf{Train Time} \\ (hrs:mins)} & \makecell{\textbf{Rel. $L_2$} \% ($\downarrow$)} & \makecell{\textbf{Rel. $L_1$} \%  ($\downarrow$)} & \textbf{$L_2/L_1$} & \makecell{\textbf{$R^2$} ($\uparrow$)} \\ \hline
Transolver & None & 2,433,927 & 1:11 & 14.6908 & 11.7212 & 1.2534 & 0.4330 \\
Transolver & Concatenation & 2,435,719 & 1:11 & 7.7780 & 5.6677 & 1.3723 & 0.9355 \\
Transolver & Cross-attention & 2,467,432 & 1:17 & 8.4583 & 6.2634 & 1.3504 & 0.9195 \\
Transolver & FiLM & 2,451,209 & 1:12 & \textbf{7.4685} & \textbf{5.5415} & 1.3477 & \textbf{0.9378} \\
\hline
UPT & None & 2,252,547 & 0:54 & 16.7382 & 13.1177 & 1.2760 & 0.4039 \\
UPT & Concatenation & 2,254,115 & 0:54 & 8.6675 & 6.2352 & 1.3901 & 0.9217 \\
UPT & Cross-attention & 2,286,052 & 1:01 & 8.4415 & 6.1447 & 1.3738 & 0.9226 \\
UPT & FiLM & 2,269,829 & 0:56 & \textbf{7.3297} & \textbf{5.4953} & 1.3338 & \textbf{0.9382} \\
\hline
MSPT & None & 2,190,915 & 6:18 & 15.4142 & 12.4451 & 1.2386 & 0.3535 \\
MSPT & Concatenation & 2,193,603 & 6:16 & 7.5068 & 5.5820 & 1.3448 & \textbf{0.9381} \\
MSPT & Cross-attention & 2,224,291 & 6:22 & \textbf{7.4161} & \textbf{5.4697} & 1.3558 & 0.9372 \\
MSPT & FiLM & 2,208,067 & 6:17 & 7.4516 & 5.5379 & 1.3456 & 0.9378 \\
\hline
GNOT & None & 2,193,035 & 3:01 & 14.9833 & 11.9531 & 1.2535 & 0.4415 \\
GNOT & Concatenation & 2,194,379 & 3:02 & 7.9032 & \textbf{5.8251} & 1.3567 & 0.9311 \\
GNOT & Cross-attention & 2,226,411 & 3:08 & 8.5310 & 6.2841 & 1.3576 & 0.9238 \\
GNOT & FiLM & 2,210,187 & 3:03 & \textbf{7.8389} & 5.8529 & 1.3393 & \textbf{0.9317} \\
\hline
NUNO & None & 2,392,713 & 0:32 & 15.5528 & 12.9050 & 1.2052 & 0.1954 \\
NUNO & Concatenation & 2,392,832 & 0:32 & 6.9904 & 5.1730 & 1.3513 & 0.9446 \\
NUNO & Cross-attention & 2,426,089 & 0:37 & 6.9994 & 5.1271 & 1.3652 & \textbf{0.9453} \\
NUNO & FiLM & 2,409,865 & 0:33 & \textbf{6.8665} & \textbf{5.1208} & 1.3409 & 0.9428 \\
\hline
GINO & None & 2,326,353 & 3:00 & 16.4890 & 13.6609 & 1.2070 & 0.1204 \\
GINO & Concatenation & 2,327,242 & 3:01 & 7.2132 & 5.2772 & 1.3669 & \textbf{0.9439} \\
GINO & Cross-attention & 2,359,729 & 3:08 & \textbf{7.1349} & \textbf{5.2345} & 1.3630 & 0.9433 \\
GINO & FiLM & 2,343,505 & 3:03 & 7.3412 & 5.4224 & 1.3539 & 0.9414 \\
\hline
SplineFNO & None & 2,357,477 & 1:54 & 15.6269 & 12.4913 & 1.2510 & 0.3607 \\
SplineFNO & Concatenation & 2,358,149 & 1:54 & 7.6161 & 5.6945 & 1.3375 & 0.9284 \\
SplineFNO & Cross-attention & 2,390,853 & 2:00 & \textbf{7.2728} & \textbf{5.4664} & 1.3304 & 0.9342 \\
SplineFNO & FiLM & 2,374,629 & 1:59 & 7.6387 & 5.6773 & 1.3455 & \textbf{0.9345} \\
\hline
L-SplineFNO & None & 2,374,351 & 0:37 & 14.3254 & 11.6969 & 1.2247 & 0.3515 \\
L-SplineFNO & Concatenation & 2,374,708 & 0:37 & 7.2337 & 5.2447 & 1.3792 & 0.9403 \\
L-SplineFNO & Cross-attention & 2,407,727 & 0:43 & \textbf{6.4678} & \textbf{4.6610} & 1.3876 & \textbf{0.9522} \\
L-SplineFNO & FiLM & 2,391,503 & 0:38 & 6.5106 & 4.7972 & 1.3572 & 0.9519 \\
\hline
\end{tabular}}%
\caption{Performance metrics of various models and conditioning methods for the FDS Building Fire (3D, time-averaged LES, three output fields) case. Single run per entry (seed 42); training times on a single NVIDIA H100; $N=92{,}837$ nodes. Best value per metric within each architecture in bold.}
\end{table}

\begin{table}[htbp]
\centering
\label{tab:heated_duct_metrics}
\resizebox{\linewidth}{!}{%
\begin{tabular}{llrrrrrr}
\hline
\textbf{Model} & \textbf{Cond. Method} & \textbf{\# Parameters} & \makecell{\textbf{Train Time} \\ (hrs:mins)} & \makecell{\textbf{Rel. $L_2$} \% ($\downarrow$)} & \makecell{\textbf{Rel. $L_1$} \%  ($\downarrow$)} & \textbf{$L_2/L_1$} & \makecell{\textbf{$R^2$} ($\uparrow$)} \\ \hline
Transolver & None & 2,433,413 & 1:14 & 18.8793 & 18.2311 & 1.0356 & 0.0207 \\
Transolver & Concatenation & 2,434,693 & 1:13 & 0.7063 & 0.5926 & 1.1918 & 0.9982 \\
Transolver & Cross-attention & 2,465,126 & 1:22 & 0.9252 & 0.8388 & 1.1029 & 0.9954 \\
Transolver & FiLM & 2,450,567 & 1:14 & \textbf{0.4481} & \textbf{0.3704} & 1.2097 & \textbf{0.9993} \\
\hline
UPT & None & 2,252,097 & 0:48 & 18.8869 & 18.2296 & 1.0361 & 0.0181 \\
UPT & Concatenation & 2,253,217 & 0:48 & 1.1621 & 0.9795 & 1.1864 & 0.9959 \\
UPT & Cross-attention & 2,283,810 & 0:58 & 0.9316 & 0.8183 & 1.1385 & 0.9952 \\
UPT & FiLM & 2,269,251 & 0:49 & \textbf{0.4518} & \textbf{0.3523} & 1.2825 & \textbf{0.9993} \\
\hline
MSPT & None & 2,190,529 & 5:24 & 18.9094 & 18.2530 & 1.0360 & 0.0255 \\
MSPT & Concatenation & 2,192,449 & 5:24 & 1.0336 & 0.8848 & 1.1681 & 0.9959 \\
MSPT & Cross-attention & 2,222,113 & 5:39 & 0.7046 & 0.6227 & 1.1315 & 0.9961 \\
MSPT & FiLM & 2,207,553 & 5:25 & \textbf{0.4253} & \textbf{0.3454} & 1.2312 & \textbf{0.9994} \\
\hline
GNOT & None & 2,192,841 & 2:51 & 18.9037 & 18.2508 & 1.0358 & 0.0238 \\
GNOT & Concatenation & 2,193,801 & 2:52 & 1.4763 & 1.3633 & 1.0828 & 0.9886 \\
GNOT & Cross-attention & 2,224,425 & 2:59 & 0.6300 & 0.5438 & 1.1586 & 0.9980 \\
GNOT & FiLM & 2,209,865 & 2:52 & \textbf{0.6042} & \textbf{0.4920} & 1.2280 & \textbf{0.9989} \\
\hline
NUNO & None & 2,392,455 & 0:04 & 18.8982 & 18.2430 & 1.0359 & 0.0234 \\
NUNO & Concatenation & 2,392,540 & 0:04 & \textbf{0.8991} & \textbf{0.6720} & 1.3381 & \textbf{0.9972} \\
NUNO & Cross-attention & 2,424,039 & 0:13 & 1.2873 & 1.1101 & 1.1596 & 0.9886 \\
NUNO & FiLM & 2,409,479 & 0:05 & 1.0169 & 0.8279 & 1.2282 & 0.9956 \\
\hline
GINO & None & 2,326,267 & 1:16 & 18.8870 & 18.2430 & 1.0353 & 0.0158 \\
GINO & Concatenation & 2,326,902 & 1:16 & \textbf{1.0203} & \textbf{0.7004} & 1.4568 & \textbf{0.9958} \\
GINO & Cross-attention & 2,357,851 & 1:26 & 1.3255 & 0.9886 & 1.3408 & 0.9916 \\
GINO & FiLM & 2,343,291 & 1:17 & 1.0631 & 0.7514 & 1.4148 & 0.9955 \\
\hline

SplineFNO & None & 2,357,279 & 0:17 & 18.9045 & 18.2496 & 1.0359 & 0.0245 \\
SplineFNO & Concatenation & 2,357,759 & 0:16 & 0.8617 & 0.7338 & 1.1744 & 0.9957 \\
SplineFNO & Cross-attention & 2,388,863 & 0:25 & 0.7999 & 0.7095 & 1.1275 & 0.9972 \\
SplineFNO & FiLM & 2,374,303 & 0:17 & \textbf{0.5080} & \textbf{0.4185} & 1.2138 & \textbf{0.9991} \\
\hline
L-SplineFNO & None & 2,371,727 & 0:26 & 18.8990 & 18.2470 & 1.0357 & 0.0239 \\
L-SplineFNO & Concatenation & 2,371,812 & 0:24 & 1.2306 & 1.0546 & 1.1669 & 0.9958 \\
L-SplineFNO & Cross-attention & 2,403,311 & 0:32 & 0.6924 & 0.6282 & 1.1021 & 0.9963 \\
L-SplineFNO & FiLM & 2,388,751 & 0:25 & \textbf{0.4357} & \textbf{0.3487} & 1.2494 & \textbf{0.9991} \\
\hline
\end{tabular}}%
\caption{Performance metrics of various models and conditioning methods for the Heated Duct (3D, steady-state, multi-region conjugate heat transfer) case. Best value per metric within each architecture in bold.}
\end{table}

\begin{table}[htbp]
\centering
\label{tab:hive_monoblock_metrics}
\resizebox{\linewidth}{!}{%
\begin{tabular}{llrrrrrr}
\hline
\textbf{Model} & \textbf{Cond. Method} & \textbf{\# Parameters} & \makecell{\textbf{Train Time} \\ (hrs:mins)} & \makecell{\textbf{Rel. $L_2$} \% ($\downarrow$)} & \makecell{\textbf{Rel. $L_1$} \%  ($\downarrow$)} & \textbf{$L_2/L_1$} & \makecell{\textbf{$R^2$} ($\uparrow$)} \\ \hline
Transolver & None & 2,119,957 & 2:23 & 6.3799 & 5.5841 & 1.1425 & 0.2876 \\
Transolver & Concatenation & 2,121,397 & 2:24 & 0.5016 & 0.4167 & 1.2038 & 0.9953 \\
Transolver & Cross-attention & 2,219,766 & 2:42 & 0.3052 & \textbf{0.2462} & 1.2396 & 0.9983 \\
Transolver & FiLM & 2,178,903 & 2:26 & \textbf{0.2884} & 0.2463 & 1.1708 & \textbf{0.9984} \\
\hline
UPT & None & 2,382,641 & 1:15 & 6.3255 & 5.5387 & 1.1420 & 0.2766 \\
UPT & Concatenation & 2,384,033 & 1:16 & 0.9408 & 0.8203 & 1.1470 & 0.9844 \\
UPT & Cross-attention & 2,482,450 & 1:37 & 0.3034 & 0.2574 & 1.1791 & 0.9985 \\
UPT & FiLM & 2,441,587 & 1:20 & \textbf{0.1983} & \textbf{0.1661} & 1.1942 & \textbf{0.9994} \\
\hline
MSPT & None & 2,190,529 & 7:39 & 6.3307 & 5.5168 & 1.1475 & 0.2818 \\
MSPT & Concatenation & 2,192,833 & 7:39 & 0.4219 & 0.3525 & 1.1970 & 0.9969 \\
MSPT & Cross-attention & 2,290,081 & 7:58 & 0.2981 & 0.2561 & 1.1637 & 0.9987 \\
MSPT & FiLM & 2,249,217 & 7:40 & \textbf{0.2811} & \textbf{0.2381} & 1.1804 & \textbf{0.9991} \\
\hline
GNOT & None & 2,377,409 & 8:48 & 6.3687 & 5.5638 & 1.1447 & 0.2882 \\
GNOT & Concatenation & 2,378,609 & 8:46 & 0.6576 & 0.5646 & 1.1648 & 0.9924 \\
GNOT & Cross-attention & 2,476,961 & 9:09 & 0.3251 & 0.2636 & 1.2336 & 0.9986 \\
GNOT & FiLM & 2,436,097 & 8:50 & \textbf{0.2129} & \textbf{0.1729} & 1.2319 & \textbf{0.9994} \\
\hline
NUNO & None & 2,374,939 & 0:08 & 6.3130 & 5.5067 & 1.1464 & 0.2771 \\
NUNO & Concatenation & 2,375,095 & 0:08 & 0.4393 & 0.3493 & 1.2575 & 0.9953 \\
NUNO & Cross-attention & 2,474,491 & 0:25 & 0.4614 & 0.3822 & 1.2074 & 0.9935 \\
NUNO & FiLM & 2,433,627 & 0:07 & \textbf{0.3053} & \textbf{0.2470} & 1.2362 & \textbf{0.9980} \\
\hline
SplineFNO & None & 2,350,813 & 0:19 & 6.3637 & 5.5509 & 1.1464 & 0.2875 \\
SplineFNO & Concatenation & 2,351,005 & 0:18 & 0.7942 & 0.6758 & 1.1751 & 0.9848 \\
SplineFNO & Cross-attention & 2,450,365 & 0:37 & 0.2720 & 0.2284 & 1.1910 & 0.9979 \\
SplineFNO & FiLM & 2,409,501 & 0:19 & \textbf{0.2179} & \textbf{0.1817} & 1.1989 & \textbf{0.9992} \\
\hline
L-SplineFNO & None & 2,351,919 & 1:05 & 6.3322 & 5.5151 & 1.1482 & 0.2815 \\
L-SplineFNO & Concatenation & 2,352,111 & 1:06 & 0.7221 & 0.6081 & 1.1876 & 0.9872 \\
L-SplineFNO & Cross-attention & 2,451,471 & 1:23 & 0.4227 & 0.3647 & 1.1592 & 0.9978 \\
L-SplineFNO & FiLM & 2,410,607 & 1:06 & \textbf{0.2115} & \textbf{0.1699} & 1.2450 & \textbf{0.9991} \\
\hline
\end{tabular}}%
\caption{Performance metrics of various models and conditioning methods for the HIVE Monoblock (3D, steady-state, thermal) case. Best value per metric within each architecture in bold.}
\end{table}

\newpage
\section{Scalability}
\label{app:scalability}
We extend the comparison in \Cref{sec:scalability} using synthetic
three-dimensional point clouds with increasing node count, scalar FiLM
conditioning, and one output channel. All runs use a single A100 80GB GPU,
FP32 with TF32 disabled, and batch size one. We retain the HIVE architecture
settings and 99 coordinate features, holding each model's latent resolution
fixed as the point cloud becomes denser. We measure inference without
an autograd graph, a forward pass with autograd enabled, a backward pass,
and a complete Adam training step. Backward timing excludes the prerequisite
forward pass; its memory includes the retained graph. Measurements use five
repetitions after warm-up for each of three seeds. Timing curves report
medians of per-seed timing medians, and memory curves report medians of
per-seed peak PyTorch allocations, including resident tensors but excluding
CUDA-context and non-PyTorch allocations. Model and data setup are outside
the timed regions.

\begin{figure}[!htbp]
    \centering
    \includegraphics[width=\linewidth]{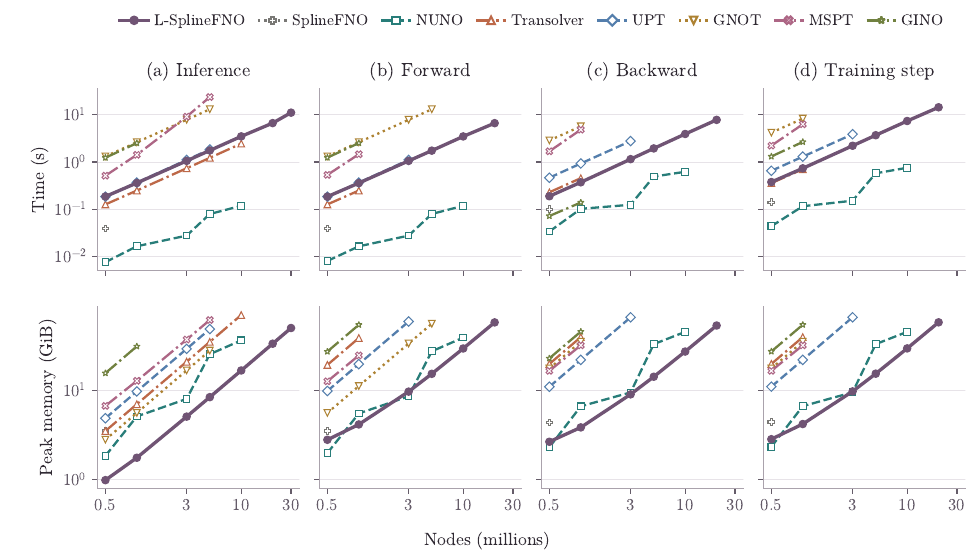}
    \caption{Runtime (top) and peak allocated GPU memory (bottom) as node
    count increases, separated by execution phase. Both axes are logarithmic.
    Points summarise successful seeds only, and lines connect measured sizes
    without extrapolation. NUNO's training result at ten million nodes
    includes two successful seeds; the remaining seed runs out of memory.
    Missing points indicate unsuccessful scaling rather than zero cost.}
    \label{fig:app-scalability-scaling}
\end{figure}

\begin{figure}[!htbp]
    \centering
    \includegraphics[width=\linewidth]{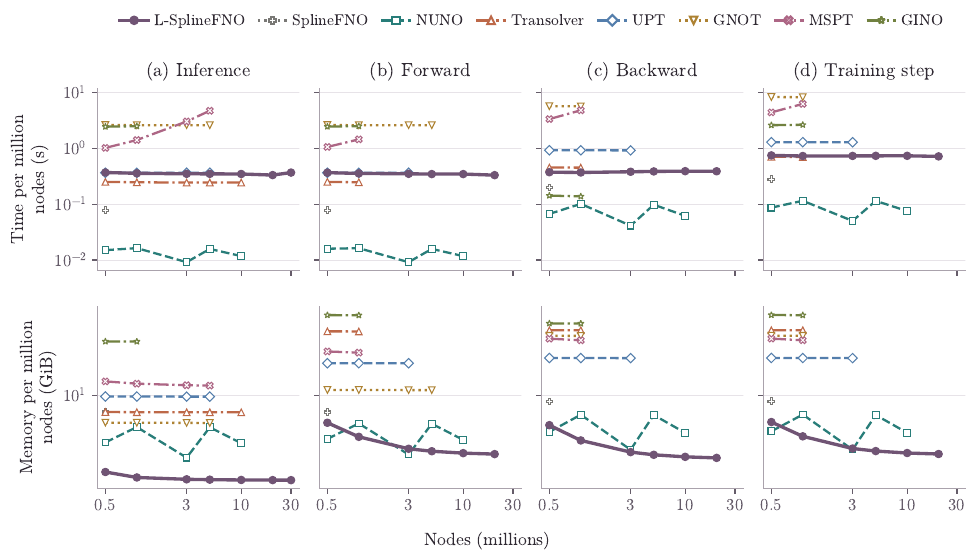}
    \caption{Runtime and peak allocated memory per million nodes, using the
    same measurements and aggregation as
    \Cref{fig:app-scalability-scaling}. Both axes are logarithmic. A nearly
    horizontal curve indicates approximately linear growth of the total cost
    over the measured range; a decreasing curve can reflect amortisation of
    fixed overhead. L-SplineFNO maintains a broadly stable cost per node as
    the point cloud grows.}
    \label{fig:app-scalability-per-node}
\end{figure}

\Cref{fig:app-scalability-scaling} shows that L-SplineFNO's runtime and memory
grow approximately linearly over the observed range, while its lower memory
cost allows execution on substantially larger point clouds. Several baselines
also exhibit near-linear trends over their feasible ranges; NUNO is faster
in the successful steady-state trials, so runtime and capacity capture
different aspects of scalability.
The approximately level L-SplineFNO curves in
\Cref{fig:app-scalability-per-node} make its stable cost per node more
apparent, with fixed overhead contributing more at smaller sizes.
These results along with \cref{sec:scalability} describe the tested implementations at fixed latent resolution; they do not establish accuracy scaling or a unique asymptotic advantage.

\newpage
\section{Ablations}
\label{app:ablations}

\subsection{Zero-shot super-resolution}
\label{sec:super-resolution}

\begin{wraptable}{r}{0.58\textwidth}
\centering
\small
\caption{Zero-shot super-resolution with FiLM-conditioned L-SplineFNO.}
\label{tab:lsplinefno_super_resolution}
\begin{tabular}{lrrr}
\toprule
\textbf{Experiment} &
\textbf{Nodes} &
\textbf{Rel. $L_2$ (\%)} &
\textbf{$R^2$} \\
\midrule
Heat Sink & 31{,}313           & 0.0239 & 0.9996 \\
Bracket   & 14{,}703--15{,}210 & 3.5142 & 0.9958 \\
Aircraft  & 331{,}971          & 8.3237 & 0.9899 \\
\bottomrule
\end{tabular}
\end{wraptable}

We evaluate L-SplineFNO without fine-tuning on held-out native meshes containing at least $95\%$ of the maximum available nodes, which is $1.91\times$, $1.79$--$1.86\times$, and $2.53\times$ the training resolution for Heat Sink, Bracket, and Aircraft, respectively. We chose these tasks to demonstrate super-resolution because of the inherent mesh variability across their datasets. As shown in \Cref{tab:lsplinefno_super_resolution}, the model retains high predictive agreement at each resolution. This zero-shot transfer is enabled by the coordinate-dependent spline scatter and decoder, which operate independently of the number and ordering of mesh nodes: L-SplineFNO can therefore be deployed on substantially finer meshes without retraining while largely preserving its predictive accuracy.

\subsection{Multiscale depth and spline degree}

\begin{figure*}[t]
\centering
\begin{subfigure}[t]{0.75\textwidth}
    \centering
    \includegraphics[width=\linewidth]{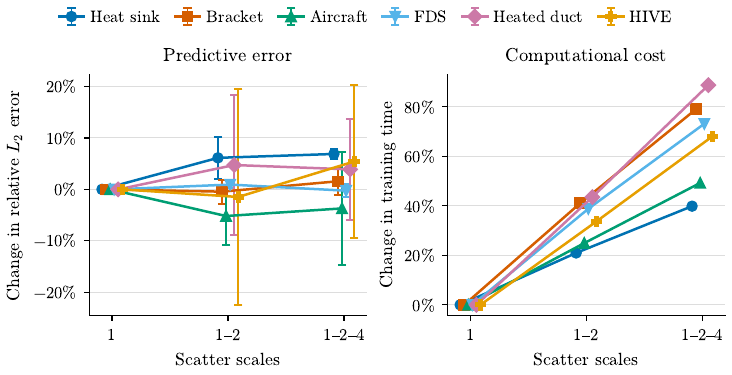}
    \caption{Effect of scatter-scale depth on predictive error and training cost.}
    \label{fig:multiscale_lsfno_tradeoff}
\end{subfigure}
\hfill
\begin{subfigure}[t]{0.75\textwidth}
    \centering
    \includegraphics[width=\linewidth]{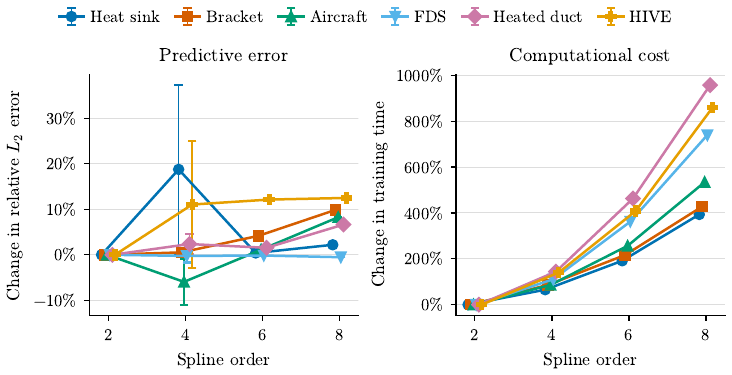}
    \caption{Effect of spline degree on predictive error and training cost.}
    \label{fig:spline_order_tradeoff}
\end{subfigure}

\caption{Accuracy--cost trade-offs for multiscale scattering and spline degree. Values are percentage changes relative to the single-scale and
cubic configurations, respectively, computed against the baseline trained at the same seed. Points are means over three seeds and error bars one standard deviation; spline orders $6$ and $8$ were run at a single seed and carry no bar. Spline orders $2$, $4$, $6$, and $8$ correspond to degrees $p=3$, $5$, $7$, and $9$, i.e.\ a $C^{p-1}$ basis. By increasing spline order and the scatter length, the model only makes marginal gains against a linear climb of computational costs. Absolute results are reported in
\Cref{tab:multiscale_lsfno_metrics,tab:spline_order_metrics}.}
\label{fig:spline_ablation_tradeoffs}
\end{figure*}

We isolate two properties of the spline transfer while holding the backbone, FiLM conditioning pathway, optimiser, and training budget fixed: the range of geometric information supplied to the FNO, and the polynomial degree, smoothness, and compact support of the spline basis. Every configuration is trained on all six benchmark cases at three seeds; a seed redraws the train/test split and the initialisation while the sampled point cloud stays fixed, and the heated duct and building fire use region mode \texttt{basis}. Because error and runtime differ substantially between benchmarks, \Cref{fig:multiscale_lsfno_tradeoff,fig:spline_order_tradeoff} report changes relative to the single-scale cubic configuration used for the main results, paired by seed so that every configuration is compared with the baseline on the same test set; the corresponding absolute errors, runtimes, parameter counts, and $R^2$ values are given in \Cref{tab:multiscale_lsfno_metrics,tab:spline_order_metrics}.

\paragraph{Multiscale scatter.}
The single-scale L-SplineFNO scatters the unstructured nodal features onto one fine latent grid. A scatter scale $s>1$ adds a second branch over the same physical domain with approximately $s$ times fewer grid intervals along each axis, so its compact spline stencil covers a wider physical neighbourhood; the coarse representation is interpolated back to the fine grid and fused with the other branches by a learned pointwise projection before the FNO blocks. We compare a fine branch alone ($1$), fine and medium branches ($1$--$2$), and fine, medium, and coarse branches ($1$--$2$--$4$), with the analytic cubic decoder unchanged in all three configurations. The additional branches add at most $0.13\%$ to the parameter count.

Multiscale depth gives no systematic accuracy gain. Aircraft is the only case it helps: the $1$--$2$ hierarchy lowers its relative $L_2$ error in every seed, by $5.2 \pm 5.7\%$, while $1$--$2$--$4$ gives $3.7 \pm 11.0\%$ with no consistent sign. The heat sink is consistently harmed, with errors $6.1 \pm 4.1\%$ and $6.9 \pm 0.9\%$ higher under the two hierarchies. On the bracket and building fire the paired changes stay below $2\%$, and on the heated duct and HIVE they are smaller than the seed-to-seed spread. The cost, by contrast, is systematic: training time rises by $21$--$44\%$ for $1$--$2$ and by $40$--$89\%$ for $1$--$2$--$4$. The aircraft results suggest that its data form a thin surface inside a volumetric latent grid, where a fine stencil leaves most knots unreached and a coarser branch carries surface information further; on volumetric meshes the fine branch already covers the grid, and the wider branches mainly add smoothing. Additional scatter scales are therefore not a reliable source of accuracy at this budget, which is why the main results use a single scale.

\begin{table}[htbp]
\centering
\caption{Multiscale L-SplineFNO (FiLM conditioning) across the benchmark cases at increasing scatter-scale depth, all at spline degree 3. Heated-duct and FDS run under region mode \texttt{basis}. Errors, $L_2/L_1$ and $R^2$ are mean $\pm$ s.d. over seeds 42, 43 and 44 where a cell has several (train/test split and initialisation vary with the seed); training time is the seed mean. For the multi-variable cases (Aircraft, FDS) $R^2$ is the mean of the per-variable scores.}
\label{tab:multiscale_lsfno_metrics}
\resizebox{\linewidth}{!}{%
\begin{tabular}{llrrrrrr}
\hline
\textbf{Case} & \textbf{Scatter Scales} & \textbf{\# Parameters} & \makecell{\textbf{Train Time} \\ (hrs:mins)} & \makecell{\textbf{Rel. $L_2$} \% ($\downarrow$)} & \makecell{\textbf{Rel. $L_1$} \%  ($\downarrow$)} & \textbf{$L_2/L_1$} & \makecell{\textbf{$R^2$} ($\uparrow$)} \\ \hline

Heat-sink   & 1       & 2,395,775 & 0:43 & \textbf{0.02842 $\pm$ 0.00331} & \textbf{0.02201 $\pm$ 0.00269} & 1.291 & \textbf{0.99968} \\
            & 1--2    & 2,396,969 & 0:51 & 0.03012 $\pm$ 0.00311 & 0.02303 $\pm$ 0.00257 & 1.309 & 0.99966 \\
            & 1--2--4 & 2,397,563 & 0:59 & 0.03036 $\pm$ 0.00327 & 0.02310 $\pm$ 0.00243 & 1.314 & 0.99965 \\ \hline

Bracket     & 1       & 2,387,731 & 0:54 & 1.9825 $\pm$ 0.0207 & 1.1096 $\pm$ 0.0283 & 1.787 & 0.99832 \\
            & 1--2    & 2,388,344 & 1:16 & \textbf{1.9734 $\pm$ 0.0284} & \textbf{1.0996 $\pm$ 0.0316} & 1.795 & \textbf{0.99833} \\
            & 1--2--4 & 2,388,651 & 1:37 & 2.0134 $\pm$ 0.0328 & 1.1319 $\pm$ 0.0225 & 1.779 & 0.99826 \\ \hline

Aircraft    & 1       & 2,368,947 & 1:05 & 8.9469 $\pm$ 0.5229 & 6.2004 $\pm$ 0.1986 & 1.442 & 0.95759 \\
            & 1--2    & 2,371,045 & 1:21 & \textbf{8.4801 $\pm$ 0.6638} & 6.1162 $\pm$ 0.4741 & 1.386 & 0.95956 \\
            & 1--2--4 & 2,372,087 & 1:37 & 8.5837 $\pm$ 0.6641 & \textbf{5.8942 $\pm$ 0.1995} & 1.455 & \textbf{0.96172} \\ \hline

FDS         & 1       & 2,391,467 & 1:02 & 6.6007 $\pm$ 0.0976 & 4.8465 $\pm$ 0.0496 & 1.362 & \textbf{0.95193} \\
            & 1--2    & 2,392,080 & 1:27 & 6.6617 $\pm$ 0.0594 & 4.8770 $\pm$ 0.0303 & 1.366 & 0.95020 \\
            & 1--2--4 & 2,392,387 & 1:48 & \textbf{6.5833 $\pm$ 0.0639} & \textbf{4.8186 $\pm$ 0.0508} & 1.366 & 0.95111 \\ \hline

Heated-duct & 1       & 2,391,231 & 0:13 & \textbf{0.4291 $\pm$ 0.0411} & \textbf{0.3569 $\pm$ 0.0540} & 1.209 & 0.99916 \\
            & 1--2    & 2,391,844 & 0:19 & 0.4519 $\pm$ 0.0961 & 0.3725 $\pm$ 0.1009 & 1.225 & 0.99906 \\
            & 1--2--4 & 2,392,151 & 0:25 & 0.4479 $\pm$ 0.0825 & 0.3756 $\pm$ 0.0755 & 1.195 & \textbf{0.99921} \\ \hline

HIVE        & 1       & 2,410,607 & 1:06 & \textbf{0.2890 $\pm$ 0.0692} & 0.2437 $\pm$ 0.0695 & 1.198 & \textbf{0.99861} \\
            & 1--2    & 2,412,705 & 1:28 & 0.2943 $\pm$ 0.1341 & \textbf{0.2412 $\pm$ 0.1188} & 1.233 & 0.99852 \\
            & 1--2--4 & 2,413,747 & 1:50 & 0.3079 $\pm$ 0.1056 & 0.2587 $\pm$ 0.0964 & 1.198 & 0.99848 \\ \hline

\end{tabular}}%
\end{table}

\paragraph{Spline degree.}
Increasing the spline degree $p$ raises the continuity of the analytic basis from $C^2$ for the cubic basis to $C^4$, $C^6$, and $C^8$ for degrees five, seven, and nine, and widens the compact support from $p+1=4$ knots per coordinate axis to six, eight, and ten. We vary this degree while retaining a single scatter scale, so the ablation changes the scatter and decode bases but not the FNO backbone; the parameter count moves by at most nine, through the rank of the partition-of-unity subspace of the learnable kernel. Higher degrees do not improve accuracy. Degree five, measured at all three seeds, lowers the aircraft error and leaves the building fire unchanged, but raises the error on the bracket, heated duct, HIVE, and heat sink. At degrees seven and nine, run at a single seed, the error is never meaningfully below the cubic one and grows with degree on the bracket, aircraft, and heated duct. Aircraft is again the case that benefits from a moderately wider support, in line with the multiscale result, but the extra smoothing of degrees seven and nine outweighs it there too. This is consistent with a wider, smoother basis offering little for fields governed by local geometric detail and, for the heated duct, by sharp changes in gradient across material interfaces, where excess smoothness mixes information across these features.

The computational penalty is substantially larger and systematic. In three dimensions, a degree-$p$ tensor-product basis contains $(p+1)^3$ stencil entries per point --- $64$, $216$, $512$, and $1000$ for degrees three, five, seven, and nine --- so the scatter and decode work scales as $O(N(p+1)^3)$, where $N$ is the number of mesh nodes. Training time accordingly rises exponenially with the degree. Because the stencil is streamed in memory-bounded chunks, every degree completes on every mesh, and the cost appears as training time rather than memory. With the parameter count effectively fixed, this cost comes from basis construction alone, which leaves the cubic basis as the most reliable accuracy--efficiency compromise across the benchmark suite.

\begin{table}[htbp]
\centering
\caption{Spline degree on single-scale L-SplineFNO (FiLM conditioning) across the benchmark cases. Heated-duct and FDS run under region mode \texttt{basis}. Errors, $L_2/L_1$ and $R^2$ are mean $\pm$ s.d. over seeds 42, 43 and 44 where a cell has several (train/test split and initialisation vary with the seed); training time is the seed mean. For the multi-variable cases (Aircraft, FDS) $R^2$ is the mean of the per-variable scores. Degrees 7 and 9 are a single seed (42), so their absolute errors are not comparable with the seed means above them; the bracketed figure is the change against degree 3 at the same seed.}
\label{tab:spline_order_metrics}
\resizebox{\linewidth}{!}{%
\begin{tabular}{llrrrrrr}
\hline
\textbf{Case} & \textbf{Spline Degree} & \textbf{\# Parameters} & \makecell{\textbf{Train Time} \\ (hrs:mins)} & \makecell{\textbf{Rel. $L_2$} \% ($\downarrow$)} & \makecell{\textbf{Rel. $L_1$} \%  ($\downarrow$)} & \textbf{$L_2/L_1$} & \makecell{\textbf{$R^2$} ($\uparrow$)} \\ \hline

Heat-sink   & 3       & 2,395,775 & 0:42 & \textbf{0.02867 $\pm$ 0.00184} & \textbf{0.02215 $\pm$ 0.00157} & 1.295 & \textbf{0.99968} \\
            & 5       & 2,395,769 & 1:10 & 0.03425 $\pm$ 0.00750 & 0.02627 $\pm$ 0.00551 & 1.302 & 0.99957 \\
            & 7       & 2,395,778 & 2:00 & 0.02852 (+0.4\%) & 0.02198 & 1.298 & 0.99973 \\
            & 9       & 2,395,769 & 3:24 & 0.02903 (+2.2\%) & 0.02265 & 1.282 & 0.99972 \\ \hline

Bracket     & 3       & 2,387,731 & 0:54 & \textbf{1.9702 $\pm$ 0.0315} & \textbf{1.0861 $\pm$ 0.0352} & 1.815 & \textbf{0.99833} \\
            & 5       & 2,387,725 & 1:40 & 1.9806 $\pm$ 0.0328 & 1.1317 $\pm$ 0.0272 & 1.750 & 0.99830 \\
            & 7       & 2,387,734 & 2:50 & 2.0421 (+4.2\%) & 1.1991 & 1.703 & 0.99812 \\
            & 9       & 2,387,725 & 4:44 & 2.1541 (+9.9\%) & 1.2939 & 1.665 & 0.99790 \\ \hline

Aircraft    & 3       & 2,368,947 & 1:05 & 9.0509 $\pm$ 0.2748 & 6.2016 $\pm$ 0.1803 & 1.459 & 0.95720 \\
            & 5       & 2,368,941 & 2:01 & \textbf{8.5180 $\pm$ 0.6485} & \textbf{6.1067 $\pm$ 0.1776} & 1.394 & \textbf{0.95968} \\
            & 7       & 2,368,950 & 3:53 & 9.4539 (+1.1\%) & 6.3742 & 1.483 & 0.94792 \\
            & 9       & 2,368,941 & 6:55 & 10.121 (+8.2\%) & 6.7832 & 1.492 & 0.94308 \\ \hline

FDS         & 3       & 2,391,467 & 1:02 & 6.6110 $\pm$ 0.0925 & 4.8578 $\pm$ 0.0438 & 1.361 & 0.95170 \\
            & 5       & 2,391,461 & 2:11 & \textbf{6.5946 $\pm$ 0.1550} & \textbf{4.8214 $\pm$ 0.1068} & 1.368 & \textbf{0.95187} \\
            & 7       & 2,391,470 & 4:41 & 6.7042 ($-$0.2\%) & 4.9129 & 1.365 & 0.94892 \\
            & 9       & 2,391,461 & 8:30 & 6.6797 ($-$0.5\%) & 4.8714 & 1.371 & 0.94975 \\ \hline

Heated-duct & 3       & 2,391,231 & 0:13 & \textbf{0.4401 $\pm$ 0.0460} & \textbf{0.3693 $\pm$ 0.0574} & 1.198 & \textbf{0.99911} \\
            & 5       & 2,391,225 & 0:32 & 0.4510 $\pm$ 0.0556 & 0.3771 $\pm$ 0.0661 & 1.203 & 0.99907 \\
            & 7       & 2,391,234 & 1:11 & 0.4383 (+1.5\%) & 0.3631 & 1.207 & 0.99918 \\
            & 9       & 2,391,225 & 2:14 & 0.4604 (+6.6\%) & 0.3723 & 1.237 & 0.99914 \\ \hline

HIVE        & 3       & 2,410,607 & 1:06 & \textbf{0.2919 $\pm$ 0.1259} & \textbf{0.2484 $\pm$ 0.1181} & 1.190 & \textbf{0.99855} \\
            & 5       & 2,410,601 & 2:38 & 0.3134 $\pm$ 0.0953 & 0.2682 $\pm$ 0.0900 & 1.176 & 0.99839 \\
            & 7       & 2,410,610 & 5:37 & 0.2339 (+12.1\%) & 0.1878 & 1.246 & 0.99901 \\
            & 9       & 2,410,601 & 10:36 & 0.2347 (+12.5\%) & 0.1958 & 1.199 & 0.99902 \\ \hline

\end{tabular}}%
\end{table}

\subsection{Analytic Gradient}

\begin{figure}[H]
    \centering
    \includegraphics[width=\textwidth]{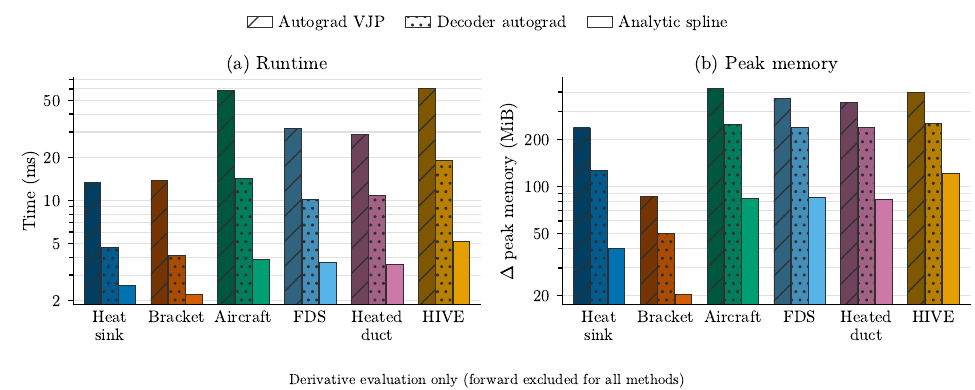}
    \caption{Gradient comparisons across autograd vector Jacobian products across the entire model, Decoder queries and the analytic spline across the six benchmark cases.}
    \label{fig:gradient_per_case}
\end{figure}

With the encoded latent grid held fixed, the L-SplineFNO decoder defines a
continuous field over the query coordinates. For a selected scalar output,
its affine readout can be applied to the latent coefficients before
interpolation, giving
\begin{equation}
    \widehat{u}(q)
    = \sum_{g\in\mathcal{G}} c_g\,\Phi_g(q) + b,
    \qquad
    \nabla_q\widehat{u}(q)
    = \sum_{g\in\mathcal{G}} c_g\,\nabla_q\Phi_g(q),
    \qquad c_g = w^{\mathsf T}h_g,
    \label{eq:appendix-spline-gradient}
\end{equation}
where $h_g$ is a latent-grid coefficient and $\Phi_g$ is the normalized
cubic tensor-product basis. The readout bias contributes no derivative.
Boundary normalization must also be differentiated: writing
$\Phi_g=B_g/S$ with $S=\sum_h B_h>0$ on the covered query domain gives
$\nabla\Phi_g=(S\nabla B_g-B_g\nabla S)/S^2$.
Thus the calculation differentiates the actual normalized decoder,
including queries near grid boundaries.

Each query touches at most $4^d$ grid points in $d$ dimensions. After
projecting the latent coefficients onto the requested output channel, the
implementation gathers each local coefficient once and reuses it for all
first-derivative components. Axis-basis evaluation and material-specific
readouts are batched to reduce repeated operations and temporary arrays.
This gives $O(N4^d)$ query-evaluation work for a fixed number of output
channels, in addition to the grid projection. For material-specific
readouts, region membership is held fixed, so derivatives are evaluated
within each material. Coordinate and output normalization scales convert
the results to physical units. These expressions remain differentiable
with respect to the latent coefficients and readout weights, allowing
first-derivative losses to train the model.

The two pointwise methods in Table~\ref{tab:analytic-gradient-cost} evaluate
this same fixed-grid field. Full-model autograd VJP instead differentiates
the sum of the selected output over all nodes with respect to the input
coordinates. Moving an input node can change the encoded grid and hence
predictions elsewhere, so this coordinate sensitivity is a different
quantity from the pointwise field gradient. Decoder autograd starts from
a prepared forward graph, whereas the analytic method constructs its
basis derivatives and projects the coefficients during the timed call;
these operations are not cached outside the measurement.

Across the six cases, the analytic evaluator is $1.84$--$3.67\times$ faster
than decoder autograd and uses $52$--$68\%$ less additional peak memory,
using the unrounded measurements. The largest relative $L_2$ disagreement
between their physical gradients is $3.04\times10^{-6}$.
Figure~\ref{fig:gradient_per_case} shows the corresponding per-case costs.
These results concern first-derivative evaluation after forward preparation;
they do not measure end-to-end inference or the memory needed to retain
that prepared state.

\begin{table}[H]
\centering
\setlength{\tabcolsep}{4pt}
\begin{tabular}{llr rr rr rr}
\toprule
& & & \multicolumn{2}{c}{\textbf{Autograd VJP}}
    & \multicolumn{2}{c}{\textbf{Decoder autograd}}
    & \multicolumn{2}{c}{\textbf{Analytic spline}} \\
& & & \multicolumn{2}{c}{(coord.\ sensitivity)}
    & \multicolumn{4}{c}{(pointwise $\nabla\widehat{u}$, latent fixed)} \\
\cmidrule(lr){4-5}\cmidrule(lr){6-7}\cmidrule(lr){8-9}
\textbf{Geometry} & $u$ & $N$ & ms & MiB & ms & MiB & ms & MiB \\
\midrule
Heat sink   & $T$        & 16\,384  & 13.3 & 235.8 & 4.7  & 125.8 & 2.5 & 39.9  \\
Bracket     & $\sigma$   & 8\,192   & 13.7 & 86.3  & 4.1  & 49.8  & 2.2 & 20.5  \\
Aircraft    & $P$        & 131\,072 & 58.9 & 423.1 & 14.2 & 248.4 & 3.9 & 83.4  \\
FDS         & $\Delta T$ & 98\,496  & 31.7 & 362.6 & 10.2 & 237.4 & 3.7 & 84.3  \\
Heated duct & $T$        & 124\,800 & 28.9 & 343.9 & 10.8 & 239.0 & 3.5 & 81.8  \\
HIVE        & $T$        & 194\,691 & 60.6 & 395.3 & 19.0 & 251.5 & 5.2 & 120.4 \\
\bottomrule
\end{tabular}
\caption{L-SplineFNO spatial-derivative cost: time and additional peak memory.
Decoder autograd and the analytic spline compute the same pointwise
gradient of field $u$ with the encoded latent grid fixed. Autograd VJP
measures full-model coordinate sensitivity. Timings cover only derivative
evaluation and unit conversion, excluding encoding and prediction, with a
fresh forward state per call. Values are medians of 20 runs after 3 warm-ups
on an A100 80GB (batch 1, FP32, TF32 off). Memory is the extra peak CUDA
allocation above the prepared state, including any saved autograd graph,
not total GPU memory.}
\label{tab:analytic-gradient-cost}
\end{table}

\end{document}